\documentclass[10pt,twocolumn,letterpaper]{article}

\usepackage[pagenumbers]{cvpr} %
\usepackage{cuted}
\usepackage{currfile}
\usepackage{caption}
\usepackage{titletoc}

\newcommand{\TODO}[1]{\textbf{\color{red}[TODO: #1]}}

\definecolor{revblue}{HTML}{1F4FBF}
\newif\ifshowrev\showrevfalse
\newif\ifpreprint %
\def\projecturl{https://real-lab-gist.github.io/valop}
\def\projecturlshort{real-lab-gist.github.io/valop} %
\newcommand{\projectlink}{\ifx\projecturl\empty\TODO{project page URL}\else\href{\projecturl}{\mbox{\small\ttfamily\projecturlshort}}\fi}
\newif\ifqualteaser\qualteasertrue %
\makeatletter\long\def\rev#1{\ifshowrev{\color{revblue}#1}\else#1\fi}\makeatother
\makeatletter\long\def\old#1{#1}\makeatother

\usepackage{amsthm,mathtools,bm}
\newcounter{proposition}
\newenvironment{proposition}[1][]{\par\addvspace{6pt}\refstepcounter{proposition}\noindent
  {\bfseries Proposition~\theproposition\if\relax\detokenize{#1}\relax\else\ (#1)\fi.}\ \itshape\ignorespaces}
  {\par\addvspace{6pt}}
\newcounter{lemma}
\newenvironment{lemma}[1][]{\par\addvspace{6pt}\refstepcounter{lemma}\noindent
  {\bfseries Lemma~\thelemma\if\relax\detokenize{#1}\relax\else\ (#1)\fi.}\ \itshape\ignorespaces}
  {\par\addvspace{6pt}}
\newenvironment{restated}[1]{\par\addvspace{6pt}\noindent
  {\bfseries #1 (restated).}\ \itshape\ignorespaces}
  {\par\addvspace{6pt}}
\newcounter{corollary}

\newcounter{appres}[section]
\renewcommand{\theappres}{\thesection.\arabic{appres}}
\newenvironment{appresult}[2][]{\par\addvspace{8pt}\refstepcounter{appres}\noindent
  {\bfseries #2~\theappres\if\relax\detokenize{#1}\relax\else\ (#1)\fi.}\ \itshape\ignorespaces}
  {\par\addvspace{6pt}}
\newenvironment{remark}[1][]{\par\addvspace{6pt}\noindent
  {\bfseries Remark\if\relax\detokenize{#1}\relax\else\ (#1)\fi.}\ \ignorespaces}
  {\par\addvspace{6pt}}
\newcommand{\proofhead}{\par\addvspace{4pt}\noindent\emph{Proof.}\ \ignorespaces}
\newcommand{\pstep}[1]{\par\medskip\noindent\textbf{#1}\ \ignorespaces}
\newcommand{\qedmark}{\hfill$\square$}
\newcommand{\pfirst}[1]{\textbf{#1}\ \ignorespaces}
\AtBeginDocument{\crefname{equation}{Eq.}{Eqs.}\Crefname{equation}{Eq.}{Eqs.}\crefname{proposition}{Prop.}{Props.}\crefname{corollary}{Cor.}{Cors.}\crefname{lemma}{Lemma}{Lemmas}\Crefname{lemma}{Lemma}{Lemmas}\Crefname{corollary}{Corollary}{Corollaries}\Crefname{proposition}{Proposition}{Propositions}
  \crefname{algorithm}{Alg.}{Algs.}\crefname{appendix}{App.}{Apps.}\Crefname{appendix}{App.}{Apps.}\crefname{subappendix}{App.}{Apps.}\Crefname{subappendix}{App.}{Apps.}\Crefname{algorithm}{Alg.}{Algs.}\crefname{section}{Sec.}{Secs.}\Crefname{section}{Sec.}{Secs.}\crefname{table}{Tab.}{Tabs.}\Crefname{table}{Tab.}{Tabs.}
  \crefname{figure}{Fig.}{Figs.}\Crefname{figure}{Fig.}{Figs.}}
\newcommand{\E}{\mathbb{E}}
\newcommand{\Var}{\mathrm{Var}}
\newcommand{\Cov}{\mathrm{Cov}}

\newcommand{\sg}{\mathrm{sg}}
\newcommand{\vz}{{\bm z}}\newcommand{\vv}{{\bm v}}\newcommand{\vu}{{\bm u}}
\newcommand{\ve}{{\bm e}}\newcommand{\vm}{{\bm m}}\newcommand{\va}{{\bm a}}
\newcommand{\vb}{{\bm b}}\newcommand{\vg}{{\bm g}}

\newcommand{\vh}{{\bm h}}\newcommand{\vxi}{{\bm\xi}}\newcommand{\vy}{{\bm y}}\newcommand{\veps}{{\bm\epsilon}}

\newcommand{\method}{VaLoP\xspace}
\newcommand{\methodfast}{VaLoP-Fast\xspace}

\usepackage{multirow,array,tabularx}
\usepackage{placeins}
\usepackage{enumitem}
\usepackage{algorithm}
\usepackage[noend]{algpseudocode}
\usepackage{pifont,makecell,colortbl}
\definecolor{oursrow}{HTML}{E6F2E6}
\definecolor{ckgreen}{HTML}{2E8B57}
\definecolor{xred}{HTML}{C0392B}
\newcommand{\cmark}{{\color{ckgreen}\ding{51}}}
\newcommand{\xmark}{{\color{xred}\ding{55}}}

\definecolor{refgray}{HTML}{8a8a8a}
\definecolor{winec}{HTML}{8E2242}

\newcommand{\appcontents}{%
  \begin{minipage}{0.86\textwidth}\small
  {\large\bfseries Contents}\par\vspace{0.5em}\hrule\vspace{0.6em}
  \printcontents[app]{l}{1}{\setcounter{tocdepth}{2}}
  \vspace{0.4em}\hrule
  \end{minipage}}
\titlecontents{lsection}[0em]{\addvspace{0.45em}\bfseries}{\contentslabel[\thecontentslabel]{1.6em}}{}{\titlerule*[0.6pc]{.}\contentspage}
\titlecontents{lsubsection}[1.6em]{\addvspace{0.1em}}{\contentslabel[\thecontentslabel]{2.3em}}{}{\titlerule*[0.6pc]{.}\contentspage}

\showrevfalse %
\preprinttrue
\newif\ifshowpreprintmark\showpreprintmarkfalse %

\definecolor{cvprblue}{rgb}{0.21,0.49,0.74}
\usepackage[pagebackref,breaklinks,colorlinks,allcolors=cvprblue]{hyperref}

\def\paperID{*****}
\def\confName{CVPR}
\def\confYear{2027}

\title{\old{VaLoP: Value Gradient Lookahead and Pullback for Reward-Guided Diffusion}}

\author{Kwanyoung Kim\\
Department of AI, GIST\\
{\tt\small k0.kim@gist.ac.kr}
}

\begin{document}
\ifshowpreprintmark\AddToShipoutPicture*{\AtPageUpperLeft{\put(\LenToUnit{0.5\paperwidth},\LenToUnit{-0.6in}){\makebox[0pt][c]{\small\textit{Preprint.}}}}}\fi
\ifqualteaser
\twocolumn[{%
\renewcommand\twocolumn[1][]{#1}%
\maketitle
\vspace{-1.2em}
\begin{center}
\ifpreprint\includegraphics[width=0.99\linewidth]{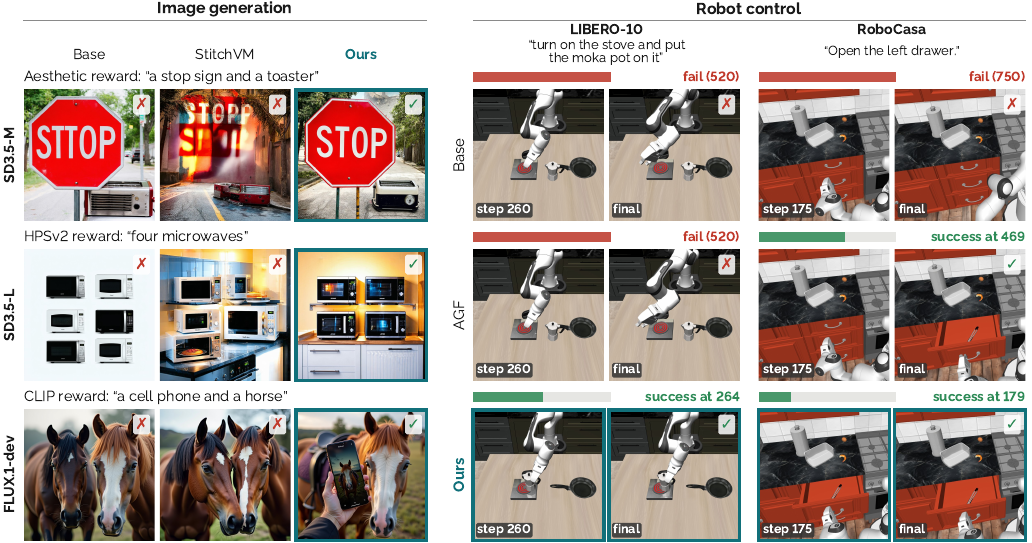}\else\includegraphics[width=\linewidth]{fig/teaser_final.pdf}\fi %
\ifpreprint\vspace{-0.2em}\else\vspace{-0.2em}\fi %
\captionof{figure}{\old{\textbf{Our proposed method guides both image generators and robot policies.}} \old{It succeeds where the unguided model and StitchVM~\cite{stitchvm} miss the GenEval prompt, and, from the same initial state, where the base policy and AGF~\cite{agf} fail the task or finish much later.} \old{\cmark/\xmark{}: GenEval correctness or task success.} \old{Bars: steps to success or the time limit.}}
\label{fig:teaser}
\end{center}
\vspace{-0.4em}
}]
\else
\twocolumn[{%
\renewcommand\twocolumn[1][]{#1}%
\maketitle
\vspace{-1.2em}
\begin{center}
\includegraphics[width=0.95\linewidth]{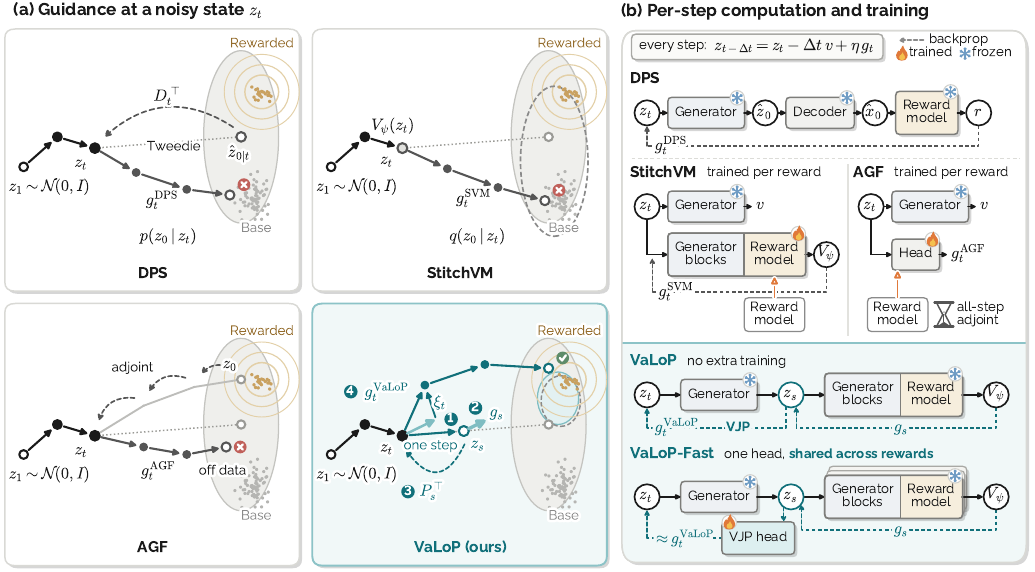}
\vspace{-0.5em}
\captionof{figure}{\old{\textbf{Comparison and pipeline overview.} (a) DPS~\cite{dps} differentiates the reward of the denoised estimate, StitchVM~\cite{stitchvm} a value model at $\vz_t$, and AGF~\cite{agf} predicts the gradient with a head trained by adjoint rollouts, while \method steps to $\vz_s$ (\ding{182}), takes its value gradient (\ding{183}), and pulls it back with a correction (\ding{184}) toward the rewarded mode (\ding{185}). (b) Per-step computation and training.} \old{\method is training-free, and \methodfast amortizes its correction with one head for all rewards, avoiding the generator backward pass and running faster than StitchVM.}}
\label{fig:overview}
\end{center}
\vspace{0.1em}
}]
\fi
\ifqualteaser
\begin{figure*}[t]
\centering
\includegraphics[width=0.92\linewidth]{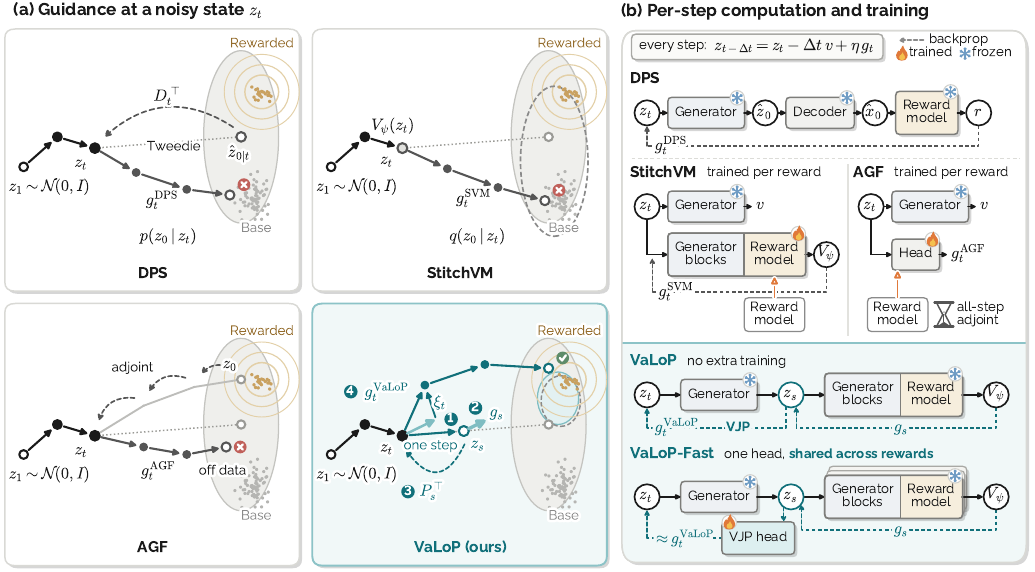}
\vspace{-0.5em}
\caption{\old{\textbf{Comparison and pipeline overview.}} \old{(a) Where the value is read: DPS~\cite{dps} at the denoised estimate ($s=0$), StitchVM~\cite{stitchvm} at $\vz_t$ ($s=t$), and \method in between, at a lookahead state $\vz_s$ (\ding{182}), whose gradient (\ding{183}) it pulls back \old{(\ding{184}) into the guidance $\vg^{\mathrm{VaLoP}}_t$} toward the rewarded mode (\ding{185}). AGF~\cite{agf} \old{learns the guidance} with a head trained by adjoint rollouts.} \old{(b) Per-step computation and training.} \old{\method{} \old{needs no extra training}, and \methodfast amortizes its correction with one head for all rewards, avoiding the generator backward pass\old{.}}}
\label{fig:overview}
\vspace{-0.5em} %
\end{figure*}
\fi

\begin{abstract}
Reward guidance steers a pretrained diffusion or flow model toward \old{images or actions} that a reward model prefers, at sampling time and without retraining the model for each reward. It follows the gradient of a value function, the expected reward given the noisy state. The two common estimates of this gradient, gradients through a one-step denoised estimate and learned value models, are least accurate at high noise, where the coarse structure of the sample is decided. Interestingly, we find that both errors are posterior covariances, which grow with the uncertainty about the clean sample, and hence at higher noise. To reduce them, we evaluate the value at a less noisy state, one sampler step ahead, where this uncertainty is smaller, and pull its gradient back to the current state. This lookahead places both estimators at the two ends of one family and reveals that an intermediate lookahead requires one additional term, a correction through the generator. We therefore propose \textbf{Va}lue Gradient \textbf{Lo}okahead and \textbf{P}ullback (\method), which computes this correction with one vector-Jacobian product, and \methodfast, which amortizes it with a \old{small} head that\old{, for text-to-image models,} is trained with a single reward and reused for the others. Across various models, rewards, and benchmark datasets, both robustly outperform existing value-gradient guidance on the target reward, \methodfast at about the cost of learned-value guidance. \old{Both extend to critic-guided robot policies, where \methodfast{} attains the highest success rates on LIBERO and RoboCasa.}%
\ifpreprint\par\vspace{0.1em}\noindent{\normalfont\old{\textbf{Project page:} \projectlink}}\fi
\end{abstract}

\section{Introduction}
\label{sec:intro}

Diffusion and flow models~\cite{scoresde,flowmatching,rectflow,sd3} generate images and robot actions, and they are increasingly steered toward outputs that a reward model prefers, such as images with higher aesthetic, preference, or text-alignment scores~\cite{aesthetic,hpsv2,imagereward,clip} and actions with higher critic value~\cite{agf}. One route fine-tunes the generator on the reward, by backpropagation through sampling~\cite{draft}, reinforcement learning~\cite{flowgrpo,diffusionnft}, or stochastic optimal control~\cite{adjointmatching}, but it must be repeated for every reward and changes the generator for all prompts. Inference-time methods instead keep the generator frozen and sample from the reward-tilted distribution \old{of}~\cite{uehara2024}. Search-based methods select or resample among many candidate samples~\cite{svdd,fksteering,ma2025scaling}, at a cost that grows with the number of candidates. Reward guidance steers a single sample by adding, at each step, the gradient of a value function, the expected reward given the current noisy state.

In this work, we focus on reward guidance, which avoids the many candidates of search. Specifically, \old{\cref{fig:overview}a contrasts} existing estimates of this gradient. Training-free guidance, such as diffusion posterior sampling (DPS)~\cite{dps}, differentiates the reward of the one-step denoised estimate of the clean sample (\cref{fig:overview}a). It backpropagates through the denoiser, the decoder, and the reward model at every step \old{(\cref{fig:overview}b)}, and the reward of this estimate, an average over plausible clean samples, differs from the expected reward. Other methods \old{build on} this estimate~\cite{freedom,universalguidance,tfg} or replace it with clean samples \old{from} a few-step solver~\cite{lidar} \old{or a distilled flow map~\cite{fmtt,fmrg,diamondmaps,mfm}}. In contrast, learned guidance evaluates a network trained on noisy states, so that the gradient avoids the denoised estimate and the decoder. The stitched value model (StitchVM)~\cite{stitchvm} predicts the reward of a clean sample from its noised version (\cref{fig:overview}b), and adjoint guidance flow (AGF)~\cite{agf} trains a lightweight guidance network that predicts adjoint gradients along the sampling paths of a robot policy (\cref{fig:overview}\old{b}).

However, neither estimate is accurate at high noise, early in sampling. Interestingly, we find that both errors stem from one source, a posterior covariance that grows with the uncertainty about the clean sample, and hence at higher noise. We verify this on a two-dimensional mixture with a known value and on a text-to-image (T2I) model, SD3.5~\cite{sd3,sd35} (\cref{fig:motivation}).

This suggests evaluating the value at a less noisy state, where this uncertainty is smaller. \method takes one deterministic sampler step from the current state to a less noisy lookahead state, evaluates the value model there, and pulls its gradient back to the current state. The lookahead ratio between the two noise levels places learned-value guidance and DPS at the two ends of one family. Beyond the value gradient at the lookahead state, an intermediate lookahead then requires a correction term, which is the vector-Jacobian product (VJP) of the generator's velocity with that gradient (\cref{fig:overview}\old{a}). In our T2I ablation, this correction brings most of the gain over learned-value guidance.

While this correction improves the guidance, it requires one VJP through the generator at every step. To address this, we propose \methodfast, which amortizes the correction with a \old{small} head, while the generator and the value model stay frozen \old{(\cref{fig:overview}b)}. Since the correction depends on the reward only through the value gradient it acts on, one head per T2I model serves all rewards\old{. In the same implementation,} \methodfast runs faster than StitchVM while reaching a higher target reward (\cref{tab:main,tab:cost}). \old{For robot policies, one head per task, trained by backpropagating through one step rather than all steps as in AGF, adds only 3 to 5\% to the inference time of the base policy.} \old{We review related work in \cref{app:related}, and} our contributions are as follows.
\begin{itemize}
\item We show that \old{the errors of} DPS and learned-value guidance \old{are} posterior covariances that grow with the uncertainty about the clean sample, and that the lookahead ratio places \old{both methods} at the two ends of one family.
\item To address \old{this} high-noise error, we propose \method, which pulls the value gradient at a lookahead state back with a correction, and \methodfast, which amortizes it with a head shared across T2I rewards \old{and trained per task for robot policies}.
\item Across \old{T2I} models, rewards, and benchmarks, \method and \methodfast{} robustly outperform StitchVM and DPS on the target reward. \old{Extended to critic-guided robot policies, \methodfast{} attains the highest success rates on LIBERO and RoboCasa.}
\end{itemize}

\section{Background}
\label{sec:background}

\subsection{Flow-Based Models}

\old{Flow-based generative models transport a noise sample $\veps\sim\mathcal N(0,I)$ to a data sample $\vz_0$ along the ODE $d\vz_t=\vv_\theta(\vz_t,t)\,dt$, where the velocity field $\vv_\theta$ is trained with the conditional flow matching objective~\cite{flowmatching,rectflow},}
\begin{equation}\label{eq:cfm}
\old{\min_\theta\;\E_{t,\vz_0,\veps}\big\|(\veps-\vz_0)-\vv_\theta(\vz_t,t)\big\|^2,}
\end{equation}
where $\vz_t=(1-t)\vz_0+t\veps$ is the linear interpolant, so that $t=0$ is data and $t=1$ is noise. We write $p$ for the distribution of generated samples and $p_t$ for the marginal of $\vz_t$ under $p$. The posterior mean $\E[\vz_0\mid \vz_t]$ is estimated by Tweedie's formula~\cite{tweedie,noise2score} as $\hat \vz_{0\mid t}=\vz_t-t\,\vv_\theta(\vz_t,t)$, which is exact for the optimal velocity of \cref{eq:cfm}, and we write $D_t=\partial\hat\vz_{0\mid t}/\partial\vz_t$ for its Jacobian. For T2I models with prompt $c$, classifier-free guidance (CFG)~\cite{cfg} at scale $\omega$ replaces $\vv_\theta$ by $\vv_\theta(\cdot,\varnothing)+\omega\big(\vv_\theta(\cdot,c)-\vv_\theta(\cdot,\varnothing)\big)$, and $\vv_\theta$ denotes this guided velocity from here on. \old{For a VLA policy, $\vz_0$ is an action chunk, and $\vv_\theta$ is also conditioned on the observation~\cite{smolvla,pi05}.} One deterministic sampler step from $t$ to $s<t$ is the map
\begin{equation}\label{eq:step}
\old{\Phi_{t\to s}(\vz)=\vz-(t-s)\,\vv_\theta(\vz,t)=\tfrac{s}{t}\,\vz+\big(1-\tfrac{s}{t}\big)\,\hat\vz_{0\mid t}(\vz),}
\end{equation}
which moves $\vz_t$ toward its posterior mean. It is a one-step approximation of the flow map of the ODE from $t$ to $s$ and returns $\hat\vz_{0\mid t}$ at $s=0$~\cite{ddim}. We call $\Phi_{t\to s}(\vz_t)$ the lookahead state at $s$, which shows where the sampler heads from $\vz_t$.

\subsection{Reward-Tilted Sampling}

Given a reward $r(\vz_0)$, such as a critic value or an aesthetic, preference, or text-alignment score~\cite{agf,aesthetic,hpsv2,clip}, reward-tilted sampling targets $p^\star(\vz_0)\propto p(\vz_0)\,e^{\lambda r(\vz_0)}$, which maximizes the expected reward minus $\lambda^{-1}\mathrm{KL}(p^\star\,\|\,p)$~\cite{uehara2024}. Its noisy marginals are $p^\star_t\propto p_t\,e^{\lambda\mathcal V_t}$, where $\mathcal V_t(\vz_t)=\lambda^{-1}\log\E[e^{\lambda r(\vz_0)}\mid\vz_t]$ is the soft value~\cite{uehara2024,svdd,adjointmatching}, and by Tweedie's formula~\cite{tweedie} the tilt shifts the posterior mean,
\begin{equation}\label{eq:shift}
\E_{p^\star}[\vz_0\mid \vz_t]=\E_p[\vz_0\mid \vz_t]+\frac{t^2}{1-t}\,\lambda\nabla_{\vz_t}\mathcal V_t(\vz_t).
\end{equation}
Guidance therefore replaces $\hat \vz_{0\mid t}$ by $\hat \vz_{0\mid t}+\lambda\frac{t^2}{1-t}\vg_t$ before each sampler step, where $\vg_t$ estimates $\nabla_{\vz_t}\mathcal V_t$ and $\lambda$ is tuned per reward in practice. To leading order in the reward scale, the soft value reduces to the value~\cite{stitchvm}, the expected reward under the generator's posterior,
\begin{equation}\label{eq:values}
V_t(\vz_t)=\E[r(\vz_0)\mid \vz_t].
\end{equation}
Like existing methods from DPS to critic guidance~\cite{dps,stitchvm,agf}, \method estimates $\nabla_{\vz_t}V_t$ (\cref{app:limits}).

\ifqualteaser\begin{figure*}[t]
\centering
\includegraphics[width=0.95\textwidth]{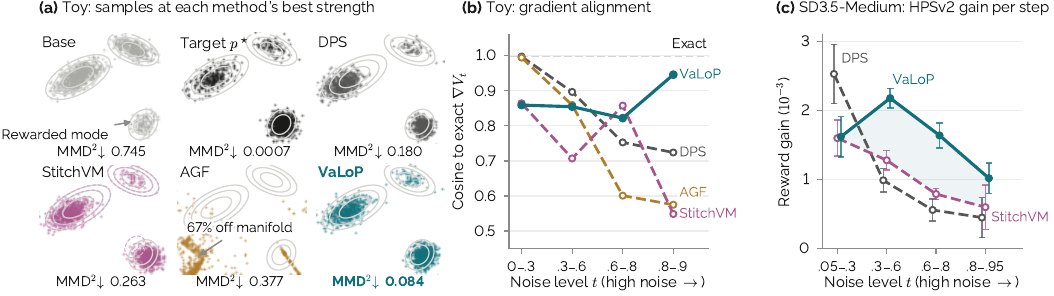}
\caption{\textbf{Value-gradient error at high noise.} \old{\textbf{(a, b)}} \old{Mixture with a misspecified value model and known exact value: samples at the best strength (a) and cosine to the exact gradient} (b\old{, \method: \old{$\nabla V_\psi$} at $\vz_s$}). \old{\textbf{(c)}} \old{HPSv2 gain per guided step} on SD3.5-Medium \old{(mean $\pm$ s.d.)}.}
\label{fig:motivation}
\vspace{-1em}
\end{figure*}
\fi
\subsection{Estimating the Value Gradient}
\label{sec:estimators}

\old{Reward guidance methods differ in the state at which they differentiate the reward or a value (\cref{fig:overview}\old{a}). DPS~\cite{dps} differentiates the reward of the posterior mean,}
\begin{equation}\label{eq:dps}
\old{\vg^{\mathrm{DPS}}_t=\nabla_{\vz_t}r(\hat\vz_{0\mid t})=D_t^{\!\top}\nabla r(\hat\vz_{0\mid t}),}
\end{equation}
\old{which backpropagates through the denoiser, the decoder, and the reward model at every step. Here $\nabla_{\vz_t}$ differentiates through every map applied to $\vz_t$, and $\nabla$ is the gradient in the first argument at the point shown. When the uncertainty about the clean sample is large, as at high noise, $\hat\vz_{0\mid t}$ averages over many clean samples, so $r(\hat\vz_{0\mid t})$ differs from the expected reward $\E[r(\vz_0)\mid\vz_t]$ by a Jensen gap~\cite{dps}.}

\old{StitchVM~\cite{stitchvm} builds a value model $V_\psi(\vz_t,t)$ by stitching the first blocks of the frozen denoiser to the reward model from one of its early blocks onward}. It is trained by regression on noised latents of dataset images $\vz_0\sim q$ from AVA~\cite{ava} and HPDv2~\cite{hpsv2} under an empty prompt,
\begin{equation}\label{eq:svmtrain}
\min_\psi\;\E_{t,\,\vz_0\sim q,\,\veps}\big\|V_\psi(\vz_t,t)-r(\vz_0)\big\|^2,
\end{equation}
in practice on the reward-model features of $\vz_0$, so that $V_\psi$ approximates the value of \cref{eq:values} under the data $q$ rather than under the generator $p$. It \old{guides with}
\begin{equation}\label{eq:svm}
\old{\vg^{\mathrm{SVM}}_t=\nabla_{\vz_t}V_\psi(\vz_t,t),}
\end{equation}
\old{which requires one backward pass through the value model.} AGF~\cite{agf} guides with a lightweight network $\vh_\phi(\vz_t,t)$ trained on its own guided paths to regress the costate,
\begin{equation}\label{eq:agf}
\vg^{\mathrm{AGF}}_t=\vh_\phi(\vz_t,t)\approx\Big(\frac{\partial\vz_0}{\partial\vz_t}\Big)^{\!\top}\nabla r(\vz_0),
\end{equation}
where the costate is the reward gradient at the endpoint $\vz_0$ of the path, pulled back to $\vz_t$ with the guidance held fixed. It needs no backward pass at inference. All three thus replace $\nabla_{\vz_t}V_t$ by a proxy that is accurate at low noise, and \cref{sec:motivation} examines how each departs from it.

\section{Motivation}
\label{sec:motivation}

\ifqualteaser\else\begin{figure*}[t]
\centering
\includegraphics[width=0.95\textwidth]{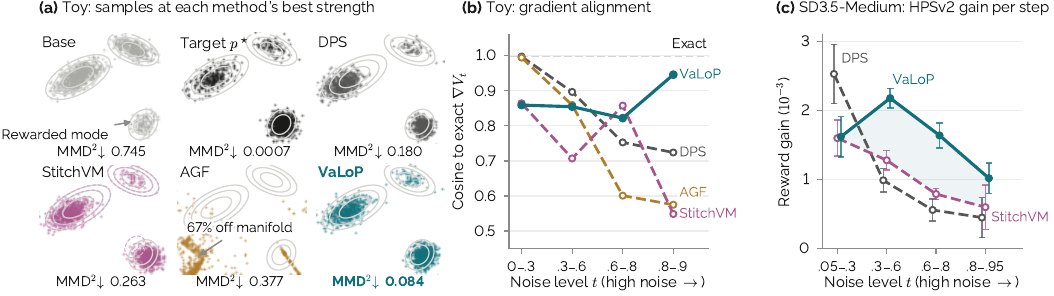}
\caption{\textbf{Value-gradient error at high noise.} \old{\textbf{(a, b)}} \old{Mixture with a misspecified value model and known exact value: samples at the best strength (a) and cosine to the exact gradient} (b\old{, \method: \old{$\nabla V_\psi$} at $\vz_s$}). \old{\textbf{(c)}} \old{HPSv2 gain per guided step} on SD3.5-Medium \old{(mean $\pm$ s.d.)}.}
\label{fig:motivation}
\vspace{-1em}
\end{figure*}
\fi

In this section, we show that the errors of all three estimators of \cref{sec:estimators} grow with the uncertainty about the clean sample given $\vz_t$, and hence at higher noise, and verify this on a two-dimensional mixture and on SD3.5 (\cref{fig:motivation}).

\noindent\textbf{\old{Value gradient as a posterior covariance.}} \old{\Cref{lem:cov} writes the target $\nabla V_t$ of all three estimators and the Jacobian $D_t$ of the posterior mean as covariances under the posterior over clean samples (proof in \old{\cref{app:proof-lem}}).}

\begin{lemma}\label{lem:cov}
\old{Let $\vz_t=(1-t)\vz_0+t\veps$ with $t\in(0,1)$, $\veps\sim\mathcal N(0,I)$ independent of $\vz_0$, and $\vz_0$ drawn from any distribution with finite second moment, and let $r$ have finite variance. Then the value $V_t(\vz)=\E[r(\vz_0)\mid\vz_t=\vz]$ and the Jacobian $D_t$ of the exact posterior mean satisfy}
\begin{align}
\old{\nabla V_t(\vz)}&\old{=\tfrac{1-t}{t^2}\,\Cov\big[\vz_0,\,r(\vz_0)\mid\vz_t=\vz\big],}\label{eq:cov}\\
\old{D_t(\vz)}&\old{=\tfrac{1-t}{t^2}\,\Cov\big[\vz_0\mid\vz_t=\vz\big].}\label{eq:dcov}
\end{align}
\end{lemma}
\old{\Cref{eq:dcov}, which applies \cref{eq:cov} to each coordinate $r(\vz_0)=z_{0,i}$, is the second-order Tweedie identity~\cite{meng2021highorder,tfg}.}

\noindent\textbf{One-step estimate (DPS).} \old{For the exact denoiser of $p$, \old{\cref{eq:dps,eq:dcov} show} that DPS is the covariance of \cref{eq:cov} with $r$ replaced by its }linearization\old{ $\bar r$ at $\hat\vz_{0\mid t}$, so that}
\begin{equation}\label{eq:dpserr}
\old{\vg^{\mathrm{DPS}}_t-\nabla V_t=-\tfrac{1-t}{t^2}\,\Cov\big[\vz_0,\,r(\vz_0)-\bar r(\vz_0)\mid\vz_t\big].}
\end{equation}
\old{DPS thus misses the part of the reward that is nonlinear over the posterior. For a smooth reward, this part vanishes as the posterior concentrates at low noise, and it can be large when the posterior spans several modes.}

\noindent\textbf{Learned value (StitchVM).} A value model is regressed onto the reward of clean samples from a data distribution $q$ given their noised versions, such as dataset images under an empty prompt in StitchVM or labeled data for a noisy classifier~\cite{dhariwal2021}. It converges to $V^q_t(\vz)=\E_q[r(\vz_0)\mid\vz_t=\vz]$, so that $\vg^{\mathrm{SVM}}_t=\nabla V^q_t$, and applying \cref{eq:cov} under $q$ and under the generator distribution $p$ gives
\begin{equation}\label{eq:svmerr}
\old{\vg^{\mathrm{SVM}}_t-\nabla V_t=\tfrac{1-t}{t^2}\big(\Cov_q-\Cov_p\big)\big[\vz_0,\,r(\vz_0)\mid\vz_t\big].}
\end{equation}
At low noise, both posteriors concentrate around $\vz_t$, where $q$ has positive density, and the difference vanishes. At high noise, the noisy state is weakly informative about the clean sample, so each posterior approaches its prior, $q$ or $p$, and the gradient follows the reward averaged over the data rather than over the generator's samples. Training on the generator's samples, as in SVDD~\cite{svdd}, removes this difference but leaves a regression target with variance $\Var[r(\vz_0)\mid\vz_t]$, which grows with the noise level on average.

\noindent\textbf{Learned gradient (AGF).} A guidance network can instead regress the gradient itself on the generator's own guided paths, as in AGF. Its target in \cref{eq:agf} is the reward gradient at the single endpoint $\vz_0$ of the path from $\vz_t$, rather than its average over the posterior. For a linear reward $r(\vz_0)=\bm w^{\!\top}\vz_0$, for which DPS is exact, a head that matches this target at zero guidance gives
\begin{equation}\label{eq:agfgap}
\vg^{\mathrm{AGF}}_t-\nabla V_t=\Big(\frac{\partial\vz_0}{\partial\vz_t}-D_t\Big)^{\!\top}\bm w,
\end{equation}
where $\partial\vz_0/\partial\vz_t$ compounds the Jacobians $D_\tau$, scaled posterior covariances, along this path, while $D_t$ is that at $\vz_t$. Both approach $I$ at low noise, and for Gaussian data \old{$\|\vg^{\mathrm{AGF}}_t\|/\|\nabla V_t\|\to\infty$ at high noise} (\cref{app:errors}). All three errors vanish at low noise and grow as the posterior widens.

\vspace{0.5em}\noindent\textbf{\old{Verification with the exact value.}} To verify these predictions where the exact value is known, we use a two-dimensional mixture (\old{\cref{fig:motivation}a, b and }\cref{app:details-toy}). \old{DPS loses alignment with the exact value gradient as the noise grows and contracts the rewarded mode toward its center.} \old{For StitchVM, we train a value model} on data in which the rewarded mode is rarer and displaced. By \cref{eq:svmerr}, this mismatch matters mainly at high noise, where the mode a sample joins is decided\old{, and the value model is indeed} least aligned at the highest noise and leaves less mass on the rewarded mode than the target. \old{AGF loses alignment faster than DPS and places many samples off the data manifold.} At high noise, the same value model is far more accurate at the less noisy lookahead state, and our method, \method, which pulls its gradient back from there (\cref{sec:method}), comes closest to the target.

\noindent\textbf{\old{Verification at scale.}} Since the exact value is unknown at scale, we instead measure the reward gain of one guided step at each noise level on SD3.5 (\cref{app:details-t2i}). Here\old{,} $q$ and $p$ differ without any construction, since StitchVM is trained on dataset images under an empty prompt while the generator samples prompt-conditioned images. In line with \old{\cref{eq:dpserr,eq:svmerr}, DPS and StitchVM steps raise} HPSv2 less than half as much at high noise as at low noise (\cref{fig:motivation}c). \old{\method} matches \old{StitchVM} at low noise, where the posterior is narrow, and gains most at intermediate noise.

Both checks thus support the analysis: the value-gradient error grows with the noise, and a value model evaluated at a less noisy state and pulled back reduces it. \Cref{sec:method} develops this lookahead and its trade-off with the bias of the step.

\section{Method: \method}
\label{sec:method}

Following \cref{sec:motivation}, \method evaluates a value model $V_\psi$, \old{such as StitchVM's}, at a less noisy lookahead state and pulls its gradient back (\cref{fig:overview}\old{a}). \old{For robot policies, a critic read at the denoised action plays the role of $V_\psi$.} \Cref{sec:pullback} defines this guidance, \old{with} learned-value guidance and DPS as its two ends, \old{\cref{sec:correction} shows that the lookahead reduces the posterior mismatch, \cref{sec:amortized} amortizes the guidance} with a \old{small} head, and \cref{sec:practical} \old{describes the implementation for T2I models and robot policies}.

\subsection{A Unified Family of Value Gradients}
\label{sec:pullback}

For $s\in[0,t]$, let $\vz_s=\Phi_{t\to s}(\vz_t)$ be the lookahead state, reached from $\vz_t$ by one deterministic step of size $t-s$ with the velocity at $\vz_t$ (\cref{eq:step}), and call $s/t$ the lookahead ratio. We propose \method, which guides at $\vz_t$ with the gradient of the value model at the lookahead state,
\begin{equation}\label{eq:valop}
\begin{aligned}
\vg^{\mathrm{VaLoP}}_t&=\nabla_{\vz_t}V_\psi\big(\Phi_{t\to s}(\vz_t),s\big)\\
&=\vg_s\underbrace{-(t-s)\,J_\vv^{\!\top}\vg_s}_{\vxi_t},
\end{aligned}
\end{equation}
where $\vg_s=\nabla V_\psi(\vz_s,s)$ and $J_\vv=\partial_\vz\vv_\theta(\vz_t,t)$. The correction $\vxi_t$, a VJP through the generator, carries $\vg_s$ back to $\vz_t$, the state that the guidance updates.

The step moves $\vz_t$ toward $\hat\vz_{0\mid t}$, so the lookahead state lies between $\vz_t$, where learned-value guidance evaluates the value, and $\hat\vz_{0\mid t}$, where DPS evaluates the reward. \Cref{prop:pullback} shows that the guidance moves between the two estimators in the same way (proofs in \old{\cref{app:proof-pullback,app:proof-mismatch}}).

\begin{proposition}\label{prop:pullback}
Let $V_\psi$ and $\vv_\theta$ be differentiable. Then
\begin{equation}\label{eq:decomp}
\vg^{\mathrm{VaLoP}}_t=\underbrace{\tfrac{s}{t}\,\vg_s}_{\text{learned-value form}}+\underbrace{\big(1-\tfrac{s}{t}\big)\,D_t^{\!\top}\vg_s}_{\text{DPS form}}.
\end{equation}
In particular, $\vg^{\mathrm{VaLoP}}_t=\vg^{\mathrm{SVM}}_t$ at $s=t$, and if $V_\psi(\cdot,0)=r$, then $\vg^{\mathrm{VaLoP}}_t=\vg^{\mathrm{DPS}}_t$ at $s=0$.
\end{proposition}

The first term of \cref{eq:decomp} is the learned-value guidance of \cref{eq:svm}, evaluated at the lookahead state, and the second is the DPS guidance of \cref{eq:dps}, with the reward gradient at the posterior mean replaced by the value gradient at $\vz_s$. The lookahead ratio thus sets where the guidance lies between the two estimators, and \cref{sec:correction} shows how this choice trades the errors of the two ends.

Finally, \method guides each step with this gradient,
\begin{equation}\label{eq:update}
\vz_{t-\Delta t}=\vz_t-\Delta t\,\vv_\theta(\vz_t,t)+\eta\,\vg^{\mathrm{VaLoP}}_t,
\end{equation}
with the strength $\eta$ (\cref{alg:valop}).

\subsection{Error Analysis of \method}
\label{sec:correction}

\leavevmode With the Jacobian $P_s=I-(t-s)J_\vv$ of the step, $\vg^{\mathrm{VaLoP}}_t=P_s^{\!\top}\vg_s$, and its error relative to $\nabla V_t$ (\cref{eq:values}) splits into a value-model error and the bias of the step,
\begin{equation}\label{eq:split}
\begin{aligned}
\vg^{\mathrm{VaLoP}}_t-\nabla V_t(\vz_t)={}&P_s^{\!\top}\big[\nabla V_\psi(\vz_s,s)-\nabla V_s(\vz_s)\big]\\
&+\big[P_s^{\!\top}\nabla V_s(\vz_s)-\nabla V_t(\vz_t)\big],
\end{aligned}
\end{equation}
where the second term vanishes at $s=t$ and equals the DPS error at $s=0$ (\cref{eq:dpserr}, \cref{app:split}). For a value model trained as in StitchVM, $V_\psi(\cdot,s)=V^q_s$, so the first term is the pulled-back gradient of the value error $V^q_s-V_s$ (\cref{eq:svmerr} at level $s$), which comes from the mismatch between the posteriors under $q$ and $p$ that the next proposition bounds.

\begin{proposition}\label{prop:mismatch}
Let $\mathrm{KL}(p\,\|\,q)<\infty$, $|r|\le B$, and $q_s$ be the marginal of $\vz_s$ under $q$, and let $\Delta_s=\E_{\vz_s\sim p_s}\mathrm{KL}\big(p(\vz_0\mid\vz_s)\,\|\,q(\vz_0\mid\vz_s)\big)$ be the posterior mismatch at level $s$. Then
\begin{equation}\label{eq:mismatch}
\Delta_s=\mathrm{KL}(p\,\|\,q)-\mathrm{KL}(p_s\,\|\,q_s),
\end{equation}
which is nondecreasing in $s$ with $\Delta_0=0$ and $\Delta_1=\mathrm{KL}(p\,\|\,q)$, and $\E_{\vz_s\sim p_s}\big|V^q_s(\vz_s)-V_s(\vz_s)\big|\le B\sqrt{2\Delta_s}$.
\end{proposition}

Together, \cref{eq:split} and \cref{prop:mismatch} show that lowering $s$ from $t$ tightens the bound behind the first term but moves the second term from zero toward the DPS error. \old{Each end is exact only in its own ideal case, $s=t$ for an exact value model and $s=0$ for a linear reward with $V_\psi(\cdot,0)=r$ and the exact denoiser, so neither is optimal in general.} In T2I generation, HPSv2 and PickScore \old{peak at $s=0.8t$ to $0.6t$ and decrease toward both ends}, and we set $s=0.6t$ (\cref{fig:depth}), a ratio that balances the two terms.

\old{Guiding} with $\vg_s$ alone, which avoids the VJP, ignores how \old{$\vz_s$} moves with $\vz_t$ \old{and is inexact even for a linear reward, for which DPS is exact} (\cref{app:split}). In T2I generation, the correction brings most of the HPSv2 gain over StitchVM (\cref{tab:ablation}).

\subsection{\texorpdfstring{\methodfast: Amortized Correction}{VaLoP-Fast: Amortized Correction}}
\label{sec:amortized}

\method improves the value gradient without \old{further} training, but its correction needs a VJP through the generator at every step, which adds to the cost of inference. To address this, we propose \methodfast, which amortizes \rev{the guidance $\vg^{\mathrm{VaLoP}}_t$} with a \old{small} head trained \rev{on states visited by the guided sampler, with the generator and the value model frozen. \Cref{alg:valop} summarizes both variants in both settings.}

\rev{\noindent\textbf{Image generation.} Since T2I latents are high-dimensional and one head should serve every reward, the head $\vh_\phi(\vz_t,t,\vg_s)$ takes the exact $\vg_s$ as input and regresses only the correction},
\begin{equation}\label{eq:amort}
\min_\phi\;\E_{\vz_t}\big\|\vh_\phi(\vz_t,t,\vg_s)-\sg[\vxi_t]\big\|^2,
\end{equation}
and at inference replaces the VJP with one forward pass of the head, without backpropagation through the generator (\old{\cref{app:head-t2i}}). Because the head learns only the correction, $\vg_s$ remains exact and carries the reward signal. The target $\vxi_t$ is the generator's VJP applied to $\vg_s$, so the reward enters only through this input, while the map from $\vg_s$ to $\vxi_t$ belongs to the generator. A head trained with one reward therefore serves the others, and we train one head per generator with the HPSv2 value and use it for all rewards (\cref{app:abl-t2i}).

\rev{\noindent\textbf{Robot policies.} Since the action chunk is low-dimensional, one head per task, $\vh_\phi(\vz_t,t)$, regresses the whole guidance instead of its correction,
\begin{equation}\label{eq:amort-robot}
\min_\phi\;\E_{\vz_t}\big\|\vh_\phi(\vz_t,t)-\sg[\vg^{\mathrm{VaLoP}}_t]\big\|^2,
\end{equation}
so that inference needs neither the critic nor a policy backward pass. In practice, the head also predicts the matched norm of \cref{sec:practical}, which needs the critic (\cref{app:head-robot}).}

\begin{algorithm}[t]
\caption{Guided sampling with \method{} \rev{and \methodfast. The head for a policy skips lines 4 and 5.}}
\label{alg:valop}
\begin{algorithmic}[1]
\Require $\vv_\theta$, \old{value $V$ ($V_\psi$, or $Q$ at the denoised action)}, ratio $\rho=s/t$, strength $\eta$, Jacobian scale $\gamma$, head $\vh_\phi$ (\methodfast)
\State $\vz_1\sim\mathcal N(0,I)$
\For{$t=1,\,1-\Delta t,\dots,\Delta t$}
  \State $\vv\gets\vv_\theta(\vz_t,t)$;\; $s\gets\rho t$
  \State $\vz_s\gets\vz_t-(t-s)\vv$ \Comment{one-step lookahead}
  \State $\vg_s\gets\nabla_{\vz_s}V(\vz_s,s)$ \Comment{value backward}
  \If{\method}
    \State \rev{$\vy_t\gets\vg_s-(t-s)J_\gamma^{\!\top}\vg_s$} \Comment{one VJP}
  \ElsIf{\rev{T2I}}
    \State \rev{$\vy_t\gets\vg_s+\vh_\phi(\vz_t,t,\vg_s)$} \Comment{head\rev{: T2I}}
  \Else
    \State \rev{$\vy_t\gets\vh_\phi(\vz_t,t)$} \Comment{\rev{head: robot policy}}
  \EndIf
  \State $\vz_{t-\Delta t}\gets\vz_t-\Delta t\,\vv+\eta\,\rev{\vy_t}$ \Comment{\cref{eq:update}}
\EndFor
\State \Return $\vz_0$
\end{algorithmic}
\end{algorithm}

\subsection{Practical Implementation}
\label{sec:practical}

\noindent\textbf{Jacobian scale.} \old{With CFG \rev{in T2I generation, the VJP in \cref{alg:valop} uses} $J_\gamma=J_u+\gamma(J_c-J_u)$, where $J_u$ and $J_c$ are the Jacobians of the unconditional and conditional velocities, rather than \rev{the} sampling scale $\omega=4.5$, which is tuned for image quality. We use $\gamma=2$ for $t\geq0.8$ and $\gamma=1$\old{, which needs only the conditional backward pass, for $t<0.8$. This schedule} performs as well as $\gamma=2$ at every step (\rev{\cref{app:abl-t2i}}).}

\noindent\textbf{Magnitude matching.} \old{In T2I generation, both variants replace $\eta\,\vg^{\mathrm{VaLoP}}_t$ in \cref{eq:update} by $\eta\,m_t\,\vy_t/\|\vy_t\|$, with $\vy_t=\vg_s+\vxi_t$ and the norm $m_t=\|\nabla V_\psi(\vz_t,t)\|$ of the StitchVM gradient, so each step moves as far as a StitchVM step. \methodfast fits $\hat m_t=e^{a(t)}\|\vg_s\|^{1+b(t)}$ per reward (\old{\cref{app:head-t2i}}) and thus needs one value-model backward pass, as StitchVM does, and no generator backward pass \rev{(\cref{alg:t2isample})}.}

\noindent\textbf{\old{Robot policies.}} \old{A VLA policy is guided by a critic $Q$ that scores a clean action chunk given the observation, trained per task as in AGF~\cite{agf}. We read it at the denoised lookahead action, $Q(\vz_s-s\,\vv_\theta(\vz_s,s))$, \rev{so that the first term of \cref{eq:split} pulls back the DPS error of \cref{eq:dpserr} at level $s$ rather than $t$, and match each update to the norm $m_t$ of the QGF~\cite{qgf} gradient at $\vz_t$ (\cref{alg:robotsample})}.}

\begin{table*}[t]
\centering\footnotesize
\caption{\old{\textbf{Reward guidance on DrawBench and GenEval}} \old{($1024^2$\old{,} strengths in \cref{app:details-t2i}). Rows are grouped by the guided reward\old{, whose column is marked by} \cmark{}, and the other columns are held-out metrics.} \old{DrawBench metrics average three seeds and GenE., the GenEval overall score, averages four (three for DPS\old{,} s.d. in \old{\cref{app:stats}}).} \old{Bold: best guided result per target and model.}}
\label{tab:main}
\vspace{-0.5em} %
\setlength{\tabcolsep}{1.38pt}
\old{\begin{tabular}{@{}l cccccc c cccccc c cccccc@{}}
\toprule
 & \multicolumn{6}{c}{\textbf{SD3.5-Medium}} && \multicolumn{6}{c}{\textbf{SD3.5-Large}} && \multicolumn{6}{c}{\textbf{FLUX.1-dev}}\\
\cmidrule(lr){2-7}\cmidrule(lr){9-14}\cmidrule(lr){16-21}
Method & HPSv2\old{$\scriptstyle\uparrow$} & Aes.\old{$\scriptstyle\uparrow$} & CLIP\old{$\scriptstyle\uparrow$} & Pick\old{$\scriptstyle\uparrow$} & IR\old{$\scriptstyle\uparrow$} & GenE.\old{$\scriptstyle\uparrow$} && HPSv2\old{$\scriptstyle\uparrow$} & Aes.\old{$\scriptstyle\uparrow$} & CLIP\old{$\scriptstyle\uparrow$} & Pick\old{$\scriptstyle\uparrow$} & IR\old{$\scriptstyle\uparrow$} & GenE.\old{$\scriptstyle\uparrow$} && HPSv2\old{$\scriptstyle\uparrow$} & Aes.\old{$\scriptstyle\uparrow$} & CLIP\old{$\scriptstyle\uparrow$} & Pick\old{$\scriptstyle\uparrow$} & IR\old{$\scriptstyle\uparrow$} & GenE.\old{$\scriptstyle\uparrow$}\\
\cmidrule(r){1-1}\cmidrule(lr){2-7}\cmidrule(lr){9-14}\cmidrule(lr){16-21}
Unguided & 0.286 & 5.347 & 28.80 & 22.65 & 0.987 & 0.694 && 0.294 & 5.461 & 28.61 & 22.87 & 1.065 & 0.721 && 0.298 & 5.729 & 27.71 & 22.79 & 0.995 & 0.660\\
\midrule
\textit{HPSv2 reward} & \cmark &  &  &  &  &  && \cmark &  &  &  &  &  && \cmark &  &  &  &  & \\
DPS & 0.361 & 5.426 & 28.99 & 23.54 & 1.130 & 0.711 && 0.349 & 5.519 & 28.82 & 23.55 & 1.111 & 0.715 && 0.320 & 5.440 & 27.09 & 22.78 & 0.922 & 0.633\\
DPS + StitchVM & 0.366 & 5.541 & 28.77 & 23.28 & 1.302 & 0.647 && 0.360 & 5.544 & 28.65 & 23.20 & 1.248 & 0.662 && 0.314 & 5.690 & 27.82 & 22.79 & 1.080 & 0.652\\
\rowcolor{oursrow}\textbf{\method} (Ours) & 0.382 & 5.654 & \textbf{29.13} & \textbf{23.77} & 1.332 & \textbf{0.726} && \textbf{0.371} & 5.633 & 29.24 & \textbf{23.78} & 1.290 & \textbf{0.755} && 0.328 & 5.590 & \textbf{28.36} & \textbf{23.12} & 1.190 & \textbf{0.701}\\
\rowcolor{oursrow}\textbf{\methodfast} (Ours) & \textbf{0.387} & \textbf{5.671} & 28.92 & 23.72 & \textbf{1.333} & 0.702 && 0.368 & \textbf{5.657} & \textbf{29.40} & 23.66 & \textbf{1.323} & 0.722 && \textbf{0.333} & \textbf{5.782} & 28.28 & 23.11 & \textbf{1.198} & 0.686\\
\midrule
\textit{Aesthetic reward} &  & \cmark &  &  &  &  &&  & \cmark &  &  &  &  &&  & \cmark &  &  &  & \\
DPS & 0.285 & 5.877 & \textbf{28.30} & 22.63 & 0.944 & 0.662 && 0.295 & 5.835 & \textbf{28.51} & 22.87 & \textbf{1.067} & 0.708 && 0.297 & 5.815 & \textbf{27.61} & 22.78 & 0.991 & 0.641\\
DPS + StitchVM & 0.279 & 5.968 & 26.95 & 22.29 & 0.801 & 0.644 && 0.287 & 5.807 & 28.05 & 22.64 & 0.992 & 0.691 && 0.297 & 5.893 & 27.37 & 22.74 & 0.967 & 0.651\\
\rowcolor{oursrow}\textbf{\method} (Ours) & 0.288 & \textbf{6.174} & 27.22 & 22.56 & 0.905 & \textbf{0.672} && \textbf{0.297} & 5.989 & 28.15 & \textbf{22.88} & 1.058 & \textbf{0.712} && 0.300 & \textbf{5.996} & 27.29 & 22.84 & 0.968 & \textbf{0.666}\\
\rowcolor{oursrow}\textbf{\methodfast} (Ours) & \textbf{0.293} & 6.035 & 27.62 & \textbf{22.69} & \textbf{0.974} & \textbf{0.672} && 0.296 & \textbf{6.144} & 28.06 & 22.80 & 1.066 & 0.706 && \textbf{0.301} & 5.929 & 27.46 & \textbf{22.85} & \textbf{1.007} & 0.659\\
\midrule
\textit{CLIP reward} &  &  & \cmark &  &  &  &&  &  & \cmark &  &  &  &&  &  & \cmark &  &  & \\
DPS & 0.254 & 4.955 & 34.27 & 21.70 & 0.622 & 0.633 && \textbf{0.286} & \textbf{5.321} & 32.13 & \textbf{22.60} & 0.982 & 0.713 && 0.279 & 5.410 & 29.28 & 22.25 & 0.819 & 0.637\\
DPS + StitchVM & 0.248 & 4.877 & 34.42 & 21.66 & 0.688 & 0.625 && 0.275 & 5.183 & 32.43 & 22.40 & 0.983 & 0.724 && 0.286 & 5.499 & 29.55 & 22.46 & 0.936 & 0.648\\
\rowcolor{oursrow}\textbf{\method} (Ours) & 0.266 & 4.999 & \textbf{35.98} & 22.08 & 0.942 & \textbf{0.725} && 0.284 & 5.281 & \textbf{33.69} & \textbf{22.60} & 1.087 & \textbf{0.759} && 0.291 & 5.514 & \textbf{30.79} & 22.65 & \textbf{1.097} & 0.691\\
\rowcolor{oursrow}\textbf{\methodfast} (Ours) & \textbf{0.275} & \textbf{5.085} & 35.03 & \textbf{22.26} & \textbf{1.022} & 0.722 && 0.282 & 5.222 & 33.65 & 22.44 & \textbf{1.095} & 0.735 && \textbf{0.295} & \textbf{5.545} & 30.37 & \textbf{22.67} & 1.082 & \textbf{0.695}\\
\bottomrule
\end{tabular}}
\end{table*}

\section{Experiments}
\label{sec:experiments}

\subsection{\old{Setup}}
\label{sec:setup}

\noindent\textbf{\old{Image generation.}} \old{We use SD3.5-Medium, SD3.5-Large~\cite{sd3,sd35}, and FLUX.1-dev~\cite{flux} at $1024^2$ on the 200 DrawBench prompts~\cite{imagen} with three shared seeds\old{ (sampling settings in \cref{app:details-t2i})}. We compare against the unguided model, pixel-space DPS~\cite{dps}, and DPS~+~StitchVM~\cite{stitchvm} with \old{the official StitchVM code}. DPS~+~StitchVM and our variants use the official StitchVM value models for HPSv2~\cite{hpsv2}, CLIP~\cite{clip}, and the CLIP+MLP aesthetic predictor~\cite{aesthetic}. We report HPSv2, the aesthetic (Aes.) and CLIP scores, PickScore (Pick)~\cite{pickscore}, ImageReward (IR)~\cite{imagereward}, and the GenEval~\cite{geneval} overall score, and call the guided reward the target reward and the others held-out metrics.}

\noindent\textbf{\rev{Robot policies.}} \old{Following AGF~\cite{agf}, we guide frozen VLA policies with a task-specific critic, SmolVLA~\cite{smolvla} on the Goal, Spatial, and LIBERO-10 suites of LIBERO~\cite{libero} and $\pi_{0.5}$~\cite{pi05} on the 18 atomic tasks of RoboCasa365~\cite{robocasa365}. We compare against critic-based best-of-$N$ selection (Q-BoN)~\cite{vgps}, QDPS\old{~\cite{qgf}}, which differentiates the critic at the denoised action through the policy, QGF~\cite{qgf}, which replaces \old{the policy Jacobian} with the identity, and AGF.} \old{\method evaluates the critic at the denoised lookahead action, and \methodfast amortizes this guidance with a head per task.}  \rev{All methods share the critic, protocol, and seeds (\cref{app:details-robot}), with success rates over 50 episodes per task.}

\subsection{\old{Main Results}}
\label{sec:main}

\noindent\textbf{Image generation.} To verify that the lookahead improves guidance across generators and rewards, we guide each model toward one reward at a time and measure the target reward together with the held-out metrics (\cref{tab:main}). Across all three rewards and models, \method raises the target reward above both StitchVM and pixel-space DPS\old{ by over one s.d.\ across seeds (\old{statistical analysis in \cref{app:stats}})}, and \methodfast performs comparably, exceeding \method in \old{three of the nine settings. \rev{To test robustness to the prompt set, we repeat this comparison on Pick-a-Pic prompts~\cite{pickscore}, where both variants again} exceed both baselines on the target reward in every setting (\cref{app:pap}).} \rev{StitchVM raises the target reward at the expense of the others, falling below the unguided model in all but one held-out entry under the aesthetic and CLIP rewards. \method instead exceeds StitchVM on every held-out metric except two entries on FLUX.1-dev, \methodfast{} does so in all nine settings, and \method exceeds both baselines on GenEval in all nine, which also shows in the qualitative results (\cref{app:qual}). Our variants thus raise the target reward more robustly than StitchVM.}

\begin{table}[t]
\centering\footnotesize
\caption{\old{\textbf{Critic guidance on robot policies.} Success rate (\%) of SmolVLA on LIBERO and $\pi_{0.5}$ on RoboCasa, averaged over the tasks of each suite. Bold: best. Rollouts in \old{\cref{app:qual}}.}}
\label{tab:robot}
\vspace{-0.5em} %
\setlength{\tabcolsep}{2.6pt}
\old{\begin{tabular}{@{}l cccc c c@{}}
\toprule
\old{Success (\%)$\uparrow$} & \multicolumn{4}{c}{LIBERO} && RoboCasa\\
\cmidrule(lr){2-5}\cmidrule(lr){7-7}
Method & Goal & Spatial & LIBERO-10 & Avg. && Atomic\\
\midrule
Base & 76.4 & 64.4 & 43.4 & 61.4 && 44.6\\
\midrule
Q-BoN & 78.2 & 69.6 & 41.2 & 63.0 && \old{46.7}\\
QGF & 79.0 & 69.4 & 48.6 & 65.7 && \old{47.1}\\
QDPS & 80.2 & 68.2 & 47.6 & 65.3 && \old{45.7}\\
AGF & 80.0 & 69.2 & 48.0 & 65.7 && 45.9\\
\midrule
\rowcolor{oursrow}\textbf{\method} (Ours) & 80.6 & 69.4 & 49.0 & 66.3 && \textbf{47.4}\\
\rowcolor{oursrow}\textbf{\methodfast} (Ours) & \textbf{81.8} & \textbf{69.8} & \textbf{49.4} & \textbf{67.0} && \textbf{47.4}\\
\bottomrule
\end{tabular}}
\end{table}

\noindent\textbf{\rev{Robot policies.}}
\label{sec:robot}
\old{To test whether the lookahead also improves critic guidance of robot policies, we guide SmolVLA and $\pi_{0.5}$ (\cref{tab:robot}). On LIBERO, every gradient-based baseline raises the success rate of the base policy, \rev{\method exceeds all baselines on average, and \methodfast{} is highest on every suite}}\old{, both significantly above the base policy (statistical analysis, \cref{app:stats})}\old{. On RoboCasa, both variants raise the success rate of the base policy more than any baseline.} \rev{In the qualitative results, \methodfast{} succeeds in every episode and is the first to succeed in most (\cref{fig:teaser}, \cref{app:qual}).} \old{The lookahead thus carries over to action generation.}

\subsection{Inference Cost}
\label{sec:cost}

\begin{table}[t]
\centering\footnotesize
\caption{\old{\textbf{Cost} on one H100: time (s) and peak memory (GB) per image at $1024^2$, and time (s) per robot action chunk. Bold: lowest.}}
\label{tab:cost}
\vspace{-0.5em} %
\setlength{\tabcolsep}{1.95pt}
\begin{tabular}{@{}lcccccc@{}}
\toprule
 & \multicolumn{2}{c}{SD3.5-Medium} & \multicolumn{2}{c}{SD3.5-Large} & \multicolumn{2}{c}{FLUX.1-dev}\\
\cmidrule(lr){2-3}\cmidrule(lr){4-5}\cmidrule(lr){6-7}
Method & Time$\downarrow$ & Mem.$\downarrow$ & Time$\downarrow$ & Mem.$\downarrow$ & Time$\downarrow$ & Mem.$\downarrow$\\
\midrule
DPS & 54.5 & 43.9 & 83.5 & 66.6 & 43.6 & 41.7\\
DPS + StitchVM & \old{17.8} & \old{18.9} & \old{36.8} & \old{34.9} & 9.9 & 32.2\\
\rowcolor{oursrow}\textbf{\method} (Ours) & 30.4 & 24.6 & 61.6 & 52.2 & 28.6 & 33.8\\
\rowcolor{oursrow}\textbf{\methodfast} (Ours) & \textbf{15.4} & \textbf{18.6} & \textbf{31.1} & \textbf{34.7} & \textbf{8.6} & \textbf{31.0}\\
\midrule
 & \multicolumn{6}{c}{\old{Time per robot action chunk (s)$\downarrow$}}\\
\cmidrule(lr){2-7}
\old{Policy} & \old{Q-BoN} & \old{QGF} & \old{QDPS} & \old{AGF} & \old{\method} & \old{\methodfast}\\
\midrule
\old{SmolVLA} & 0.463 & 0.587 & 1.045 & \textbf{0.460} & 2.048 & \textbf{0.460}\\
\old{$\pi_{0.5}$} & 0.250 & 0.303 & 0.633 & \textbf{0.240} & 1.139 & \textbf{0.240}\\
\bottomrule
\end{tabular}
\end{table}

\noindent\old{\textbf{Image generation.}} \old{\method adds one generator VJP and a second value-model backward pass per step, yet takes less time and memory than pixel-space DPS (\cref{tab:cost}). \methodfast replaces both with one head pass and the fitted magnitude, and it needs less time and memory than \old{StitchVM} on every model, \old{with both run in our optimized implementation} (\cref{app:details-t2i}).}

\noindent\textbf{\rev{Robot policies.}} \old{By amortizing the guidance, \methodfast adds only 3 to 5\% to the inference time of the base policy on SmolVLA and $\pi_{0.5}$, as AGF does, runs faster than QGF and QDPS \old{since it needs no critic}, and trains its head on SmolVLA in under half the time of AGF (\old{\cref{app:details-robot}}).}

\subsection{Ablation Study}
\label{sec:ablation}

\begin{table}[t]
\centering\footnotesize
\caption{\old{\textbf{\rev{Lookahead and correction}} (SD3.5-Medium, $1024^2$, DrawBench, HPSv2 reward, $\eta=300$\old{, one seed}).} \old{Bold: best.}}
\label{tab:ablation}
\vspace{-0.5em} %
\setlength{\tabcolsep}{3.6pt}
\old{\begin{tabular}{@{}lccccc@{}}
\toprule
 & HPSv2\old{$\uparrow$} & Aes.\old{$\uparrow$} & CLIP\old{$\uparrow$} & Pick\old{$\uparrow$} & IR\old{$\uparrow$}\\
\midrule
\rowcolor{oursrow}\textbf{\method} (Ours) & \textbf{0.382} & \rev{\textbf{5.665}} & \textbf{29.06} & \textbf{23.79} & \rev{\textbf{1.339}}\\
\midrule
\multicolumn{6}{@{}l}{\textit{Lookahead and correction}}\\
\old{No lookahead} ($s=t$) & 0.358 & 5.505 & 28.63 & 23.28 & 1.322\\
No correction ($\vg_s$ only) & 0.365 & 5.614 & 28.23 & 23.52 & 1.324\\
\midrule
\multicolumn{6}{@{}l}{\textit{Form of the correction}}\\
$J$ instead of $J^{\!\top}$ & 0.368 & 5.598 & 28.25 & 23.59 & 1.318\\
\bottomrule
\end{tabular}}
\end{table}

\begin{table}[t]
\centering\footnotesize
\caption{\old{\textbf{Lookahead and correction on SmolVLA} (success \%), all with the critic at the denoised action. Bold: best.}}
\label{tab:robot_abl}
\vspace{-0.5em}
\setlength{\tabcolsep}{4pt}
\old{\begin{tabular}{@{}lccc@{}}
\toprule
 & Goal & Spatial & LIBERO-10\\
\midrule
No lookahead ($s=t$) & 79.0 & 69.0 & 47.4\\
No correction ($\vg_s$ only) & 76.6 & 67.0 & 46.4\\
\rowcolor{oursrow}\textbf{\method} (Ours) & \textbf{80.6} & \textbf{69.4} & \textbf{49.0}\\
\bottomrule
\end{tabular}}
\end{table}

\begin{figure}[t]
\centering
\includegraphics[width=\columnwidth]{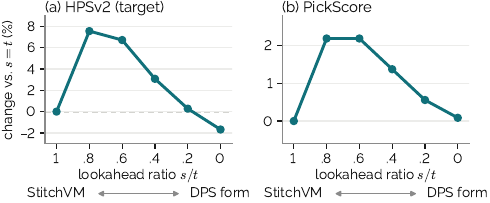}
\vspace{-1.5em}
\caption{\textbf{Lookahead ratio} (SD3.5-Medium, HPSv2 reward, DrawBench at $1024^2$), relative to learned-value guidance at $s=t$.}
\label{fig:depth}
\end{figure}

\noindent\textbf{Lookahead ratio.} To test the trade-off of \cref{sec:correction}, we sweep $s/t$ from $1$, learned-value guidance, to $0$, the DPS form. HPSv2 and PickScore peak at $s=0.8t$ to $0.6t$ and decrease toward both ends (\cref{fig:depth}), consistent with the dominance of the value-model error near $s=t$ and of the step bias near $s=0$. \old{All ratios but $0.8t$} are significantly worse than $0.6t$ in both metrics \old{(paired $t$-test, }\old{\cref{app:stats}}\old{)}, and $0.8t$ is slightly higher only in HPSv2. We select $0.6t$ on held-out prompts (\old{\cref{app:details-t2i}}). \old{On LIBERO-Goal, $s=0.6t$ \old{is also best on average over strengths} for SmolVLA (\old{\cref{app:abl-robot}}).}

\noindent\textbf{Lookahead and correction.} \old{Guided by $\vg_s$ alone, the lookahead raises} HPSv2 and PickScore above \old{$s=t$}, but the correction accounts for the larger part of the gain in both (\cref{tab:ablation})\old{, so the pullback carries the gain}. \old{\methodfast keeps most of }\old{it}\old{ (\old{\cref{app:abl-t2i}}).}

\rev{With the critic on SmolVLA, $\vg_s$ alone falls below $s=t$ on every suite, and only the correction lifts \method above both (\cref{tab:robot_abl}). The lookahead also improves a learned value: StitchVM, not in \cref{tab:robot} as it needs an extra value model on noisy actions for each task, rises on LIBERO-Goal from 77.4 to 79.6\% when this model, stitched to SmolVLA, is read with \method (\cref{app:abl-robot}).}

\noindent\textbf{Form of the correction.} \old{We} replace $J_\gamma^{\!\top}$ by $J_\gamma$, which forward differences compute without a backward pass. For the exact posterior mean, $D_t$ is symmetric by \cref{eq:dcov}, so both give the same correction, but with CFG the velocity Jacobian is strongly asymmetric on SD3.5 (\old{\cref{app:abl-t2i}}), and the replacement lowers every metric, HPSv2 nearly to the level of $\vg_s$ alone (\cref{tab:ablation}). \old{The pullback thus needs the transpose.}

\section{Conclusion}
\label{sec:conclusion}

\old{We proposed \method, which evaluates a value model at a less noisy lookahead state and pulls its gradient back, placing learned-value guidance and DPS at the two ends of one family and reducing the value-gradient error at high noise. Across three generators and three rewards, \method and \methodfast raise the target reward while better preserving the held-out metrics, \methodfast at a lower inference cost than learned-value guidance. In critic guidance of robot policies, \methodfast attains the highest success rates on LIBERO and RoboCasa. \old{As} the correction depends on the generator\old{, not} the reward, one head per} \old{image} \old{generator serves every reward, \old{which we believe opens} a path toward guidance modules \old{trained once} and reused for any reward.}

\paragraph{Acknowledgments.} We sincerely thank the authors of StitchVM~\cite{stitchvm} and AGF~\cite{agf} for sharing their code.
\paragraph{\old{LLM Usage Statement.}} \old{We used a commercial large language model (Claude) to help draft and edit the text of this paper, for wording, grammar, and consistency, and to write the plotting scripts for the figures. The authors directed this use, checked every statement against the results and the cited sources, and take full responsibility for the content.}

{
    \small
    \bibliographystyle{ieeenat_fullname}
    \bibliography{refs}

\begin{thebibliography}{76}
\providecommand{\natexlab}[1]{#1}
\providecommand{\url}[1]{\texttt{#1}}
\expandafter\ifx\csname urlstyle\endcsname\relax
  \providecommand{\doi}[1]{doi: #1}\else
  \providecommand{\doi}{doi: \begingroup \urlstyle{rm}\Url}\fi

\bibitem[Bansal et~al.(2024)Bansal, Chu, Schwarzschild, Sengupta, Goldblum,
  Geiping, and Goldstein]{universalguidance}
Arpit Bansal, Hong-Min Chu, Avi Schwarzschild, Soumyadip Sengupta, Micah
  Goldblum, Jonas Geiping, and Tom Goldstein.
\newblock {Universal Guidance for Diffusion Models}.
\newblock In \emph{International Conference on Learning Representations
  (ICLR)}, 2024.

\bibitem[Ben-Hamu et~al.(2024)Ben-Hamu, Puny, Gat, Karrer, Singer, and
  Lipman]{dflow}
Heli Ben-Hamu, Omri Puny, Itai Gat, Brian Karrer, Uriel Singer, and Yaron
  Lipman.
\newblock {D-Flow: Differentiating through Flows for Controlled Generation}.
\newblock In \emph{Proceedings of the 41st International Conference on Machine
  Learning (ICML)}, pages 3462--3483. PMLR, 2024.

\bibitem[Black et~al.(2024)Black, Janner, Du, Kostrikov, and Levine]{ddpo}
Kevin Black, Michael Janner, Yilun Du, Ilya Kostrikov, and Sergey Levine.
\newblock {Training Diffusion Models with Reinforcement Learning}.
\newblock In \emph{International Conference on Learning Representations
  (ICLR)}, 2024.

\bibitem[Black et~al.(2025)Black, Brown, Darpinian, Dhabalia, Driess, Esmail,
  Equi, Finn, Fusai, Galliker, Ghosh, Groom, Hausman, Ichter, Jakubczak, Jones,
  Ke, LeBlanc, Levine, Li-Bell, Mothukuri, Nair, Pertsch, Ren, Shi, Smith,
  Springenberg, Stachowicz, Tanner, Vuong, Walke, Walling, Wang, Yu, and
  Zhilinsky]{pi05}
Kevin Black, Noah Brown, James Darpinian, Karan Dhabalia, Danny Driess, Adnan
  Esmail, Michael~Robert Equi, Chelsea Finn, Niccolo Fusai, Manuel~Y. Galliker,
  Dibya Ghosh, Lachy Groom, Karol Hausman, Brian Ichter, Szymon Jakubczak, Tim
  Jones, Liyiming Ke, Devin LeBlanc, Sergey Levine, Adrian Li-Bell, Mohith
  Mothukuri, Suraj Nair, Karl Pertsch, Allen~Z. Ren, Lucy~Xiaoyang Shi, Laura
  Smith, Jost~Tobias Springenberg, Kyle Stachowicz, James Tanner, Quan Vuong,
  Homer Walke, Anna Walling, Haohuan Wang, Lili Yu, and Ury Zhilinsky.
\newblock {$\pi_{0.5}$: a Vision-Language-Action Model with Open-World
  Generalization}.
\newblock In \emph{Proceedings of The 9th Conference on Robot Learning (CoRL)},
  pages 17--40. PMLR, 2025.

\bibitem[{Black Forest Labs}(2024)]{flux}
{Black Forest Labs}.
\newblock {FLUX.1 [dev]}.
\newblock \url{https://github.com/black-forest-labs/flux}, 2024.

\bibitem[Chung et~al.(2023)Chung, Kim, McCann, Klasky, and Ye]{dps}
Hyungjin Chung, Jeongsol Kim, Michael~T. McCann, Marc~L. Klasky, and Jong~Chul
  Ye.
\newblock {Diffusion Posterior Sampling for General Noisy Inverse Problems}.
\newblock In \emph{International Conference on Learning Representations
  (ICLR)}, 2023.

\bibitem[Clark et~al.(2024)Clark, Vicol, Swersky, and Fleet]{draft}
Kevin Clark, Paul Vicol, Kevin Swersky, and David~J. Fleet.
\newblock {Directly Fine-Tuning Diffusion Models on Differentiable Rewards}.
\newblock In \emph{International Conference on Learning Representations
  (ICLR)}, 2024.

\bibitem[Dai et~al.(2025)Dai, Zhuang, Huang, Huang, Liao, Wang, and Yuan]{vard}
Fengyuan Dai, Zifeng Zhuang, Yufei Huang, Siteng Huang, Bangyan Liao, Donglin
  Wang, and Fajie Yuan.
\newblock {{VARD}: Efficient and Dense Fine-Tuning for Diffusion Models with
  Value-based {RL}}.
\newblock \emph{arXiv preprint arXiv:2505.15791}, 2025.

\bibitem[Dandapanthula and Boffi(2026)]{tilting}
Sanjit Dandapanthula and Nicholas~M. Boffi.
\newblock {Are We Really Tilting? The Mechanics of Reward Guidance in Flow and
  Diffusion Models}.
\newblock In \emph{Advances in Neural Information Processing Systems
  (NeurIPS)}, 2026.

\bibitem[Denker et~al.(2024)Denker, Vargas, Padhy, Didi, Mathis, Dutordoir,
  Barbano, Mathieu, Komorowska, and Li{\`o}]{deft}
Alexander Denker, Francisco Vargas, Shreyas Padhy, Kieran Didi, Simon Mathis,
  Vincent Dutordoir, Riccardo Barbano, Emile Mathieu, Urszula~Julia Komorowska,
  and Pietro Li{\`o}.
\newblock {DEFT: Efficient Fine-tuning of Diffusion Models by Learning the
  Generalised $h$-transform}.
\newblock In \emph{Advances in Neural Information Processing Systems
  (NeurIPS)}, pages 19636--19682, 2024.

\bibitem[Dhariwal and Nichol(2021)]{dhariwal2021}
Prafulla Dhariwal and Alexander Nichol.
\newblock {Diffusion Models Beat GANs on Image Synthesis}.
\newblock In \emph{Advances in Neural Information Processing Systems
  (NeurIPS)}, pages 8780--8794, 2021.

\bibitem[Domingo-Enrich et~al.(2025)Domingo-Enrich, Drozdzal, Karrer, and
  Chen]{adjointmatching}
Carles Domingo-Enrich, Michal Drozdzal, Brian Karrer, and Ricky T.~Q. Chen.
\newblock {Adjoint Matching: Fine-tuning Flow and Diffusion Generative Models
  with Memoryless Stochastic Optimal Control}.
\newblock In \emph{International Conference on Learning Representations
  (ICLR)}, 2025.

\bibitem[Efron(2011)]{tweedie}
Bradley Efron.
\newblock {{Tweedie's} Formula and Selection Bias}.
\newblock \emph{Journal of the American Statistical Association}, 106\penalty0
  (496):\penalty0 1602--1614, 2011.

\bibitem[Esser et~al.(2024)Esser, Kulal, Blattmann, Entezari, M{\"u}ller,
  Saini, Levi, Lorenz, Sauer, Boesel, Podell, Dockhorn, English, and
  Rombach]{sd3}
Patrick Esser, Sumith Kulal, Andreas Blattmann, Rahim Entezari, Jonas
  M{\"u}ller, Harry Saini, Yam Levi, Dominik Lorenz, Axel Sauer, Frederic
  Boesel, Dustin Podell, Tim Dockhorn, Zion English, and Robin Rombach.
\newblock {Scaling Rectified Flow Transformers for High-Resolution Image
  Synthesis}.
\newblock In \emph{Proceedings of the 41st International Conference on Machine
  Learning (ICML)}, pages 12606--12633. PMLR, 2024.

\bibitem[Eyring et~al.(2024)Eyring, Karthik, Roth, Dosovitskiy, and
  Akata]{reno}
Luca Eyring, Shyamgopal Karthik, Karsten Roth, Alexey Dosovitskiy, and Zeynep
  Akata.
\newblock {ReNO: Enhancing One-step Text-to-Image Models through Reward-based
  Noise Optimization}.
\newblock In \emph{Advances in Neural Information Processing Systems
  (NeurIPS)}, pages 125487--125519, 2024.

\bibitem[Fan et~al.(2023)Fan, Watkins, Du, Liu, Ryu, Boutilier, Abbeel,
  Ghavamzadeh, Lee, and Lee]{dpok}
Ying Fan, Olivia Watkins, Yuqing Du, Hao Liu, Moonkyung Ryu, Craig Boutilier,
  Pieter Abbeel, Mohammad Ghavamzadeh, Kangwook Lee, and Kimin Lee.
\newblock {DPOK: Reinforcement Learning for Fine-tuning Text-to-Image Diffusion
  Models}.
\newblock In \emph{Advances in Neural Information Processing Systems
  (NeurIPS)}, pages 79858--79885, 2023.

\bibitem[Gao et~al.(2025)Gao, Zha, Zhang, Xue, and Boning]{reg}
Zhengqi Gao, Kaiwen Zha, Tianyuan Zhang, Zihui Xue, and Duane~S. Boning.
\newblock {REG: Rectified Gradient Guidance for Conditional Diffusion Models}.
\newblock In \emph{Proceedings of the 42nd International Conference on Machine
  Learning (ICML)}, pages 18724--18743. PMLR, 2025.

\bibitem[Ghosh et~al.(2023)Ghosh, Hajishirzi, and Schmidt]{geneval}
Dhruba Ghosh, Hannaneh Hajishirzi, and Ludwig Schmidt.
\newblock {{GenEval}: An Object-Focused Framework for Evaluating Text-to-Image
  Alignment}.
\newblock In \emph{Advances in Neural Information Processing Systems
  (NeurIPS)}, pages 52132--52152, 2023.

\bibitem[Go et~al.(2026)Go, Chung, Truong, Bhat, Mi, An, Zhao, Narnhofer,
  Belongie, Tombari, and Schindler]{stitchvm}
Hyojun Go, Hyungjin Chung, Prune Truong, Goutam Bhat, Li Mi, Zhaochong An,
  Zixiang Zhao, Dominik Narnhofer, Serge Belongie, Federico Tombari, and Konrad
  Schindler.
\newblock {Stitched Value Model for Diffusion Alignment}.
\newblock In \emph{Advances in Neural Information Processing Systems
  (NeurIPS)}, 2026.

\bibitem[Gretton et~al.(2012)Gretton, Borgwardt, Rasch, Sch{\"o}lkopf, and
  Smola]{gretton2012mmd}
Arthur Gretton, Karsten~M. Borgwardt, Malte~J. Rasch, Bernhard Sch{\"o}lkopf,
  and Alexander Smola.
\newblock {A Kernel Two-Sample Test}.
\newblock \emph{Journal of Machine Learning Research}, 13\penalty0
  (25):\penalty0 723--773, 2012.

\bibitem[Ho and Salimans(2022)]{cfg}
Jonathan Ho and Tim Salimans.
\newblock {Classifier-Free Diffusion Guidance}.
\newblock \emph{arXiv preprint arXiv:2207.12598}, 2022.

\bibitem[Holderrieth et~al.(2026)Holderrieth, Chen, Eyring, Shah, Anantharaman,
  He, Akata, Jaakkola, Boffi, and Simchowitz]{diamondmaps}
Peter Holderrieth, Douglas Chen, Luca Eyring, Ishin Shah, Giri Anantharaman,
  Yutong He, Zeynep Akata, Tommi Jaakkola, Nicholas~Matthew Boffi, and Max
  Simchowitz.
\newblock {Diamond Maps: Efficient Reward Alignment via Stochastic Flow Maps}.
\newblock In \emph{Proceedings of the 43rd International Conference on Machine
  Learning (ICML)}, pages 43401--43435. PMLR, 2026.

\bibitem[Huang et~al.(2026)Huang, Lin, Shah, Nair, and Boffi]{fmrg}
Jerry~Y. Huang, Justin Lin, Sheel Shah, Kartik Nair, and Nicholas~M. Boffi.
\newblock {How to Guide Your Flow: Few-Step Alignment via Flow Map Reward
  Guidance}.
\newblock In \emph{Proceedings of the 43rd International Conference on Machine
  Learning (ICML)}, pages 47101--47152. PMLR, 2026.

\bibitem[Janner et~al.(2022)Janner, Du, Tenenbaum, and Levine]{diffuser}
Michael Janner, Yilun Du, Joshua Tenenbaum, and Sergey Levine.
\newblock {Planning with Diffusion for Flexible Behavior Synthesis}.
\newblock In \emph{Proceedings of the 39th International Conference on Machine
  Learning (ICML)}, pages 9902--9915. PMLR, 2022.

\bibitem[Kim et~al.(2026{\natexlab{a}})Kim, Jun, Choi, Kim, Jin, Park, Park,
  Kim, and Ye]{agf}
Jeongsol Kim, Youngjun Jun, Kyumin Choi, Youngmin Kim, Seonghyun Jin, Sunwoo
  Park, Jangho Park, Kwanyoung Kim, and Jong~Chul Ye.
\newblock {Adjoint Guidance Flow: Amortized Critic Guidance for {VLA}
  Policies}.
\newblock \emph{arXiv preprint arXiv:2609.34944}, 2026{\natexlab{a}}.

\bibitem[Kim and Ye(2021)]{noise2score}
Kwanyoung Kim and Jong~Chul Ye.
\newblock {{Noise2Score}: {Tweedie's} Approach to Self-Supervised Image
  Denoising without Clean Images}.
\newblock In \emph{Advances in Neural Information Processing Systems
  (NeurIPS)}, pages 864--874, 2021.

\bibitem[Kim et~al.(2026{\natexlab{b}})Kim, Shin, Na, Park, Kim, and
  Moon]{lidar}
Yeongmin Kim, Donghyeok Shin, Byeonghu Na, Minsang Park, Richard~Lee Kim, and
  Il-Chul Moon.
\newblock {Lookahead Sample Reward Guidance for Test-Time Scaling of Diffusion
  Models}.
\newblock In \emph{Proceedings of the 43rd International Conference on Machine
  Learning (ICML)}, pages 58969--58995. PMLR, 2026{\natexlab{b}}.

\bibitem[Kirstain et~al.(2023)Kirstain, Polyak, Singer, Matiana, Penna, and
  Levy]{pickscore}
Yuval Kirstain, Adam Polyak, Uriel Singer, Shahbuland Matiana, Joe Penna, and
  Omer Levy.
\newblock {Pick-a-Pic: An Open Dataset of User Preferences for Text-to-Image
  Generation}.
\newblock In \emph{Advances in Neural Information Processing Systems
  (NeurIPS)}, pages 36652--36663, 2023.

\bibitem[Li and Levine(2026)]{qam}
Qiyang Li and Sergey Levine.
\newblock {Q-Learning with Adjoint Matching}.
\newblock In \emph{International Conference on Learning Representations
  (ICLR)}, 2026.

\bibitem[Li et~al.(2025)Li, Zhao, Wang, Scalia, Eraslan, Nair, Biancalani, Ji,
  Regev, Levine, and Uehara]{svdd}
Xiner Li, Yulai Zhao, Chenyu Wang, Gabriele Scalia, G{\"o}kcen Eraslan, Surag
  Nair, Tommaso Biancalani, Shuiwang Ji, Aviv Regev, Sergey Levine, and
  Masatoshi Uehara.
\newblock {Derivative-Free Guidance in Continuous and Discrete Diffusion Models
  with Soft Value-based Decoding}.
\newblock In \emph{Advances in Neural Information Processing Systems
  (NeurIPS)}, pages 95507--95545, 2025.

\bibitem[Liang et~al.(2026)Liang, Zhang, and Yang]{ecg}
Zhengyang Liang, Qihang Zhang, and Ceyuan Yang.
\newblock {Explicit Critic Guidance for Aligning Diffusion Models}.
\newblock \emph{arXiv preprint arXiv:2605.27736}, 2026.

\bibitem[Lipman et~al.(2023)Lipman, Chen, Ben-Hamu, Nickel, and
  Le]{flowmatching}
Yaron Lipman, Ricky T.~Q. Chen, Heli Ben-Hamu, Maximilian Nickel, and Matt Le.
\newblock {Flow Matching for Generative Modeling}.
\newblock In \emph{International Conference on Learning Representations
  (ICLR)}, 2023.

\bibitem[Liu et~al.(2023{\natexlab{a}})Liu, Zhu, Gao, Feng, Liu, Zhu, and
  Stone]{libero}
Bo Liu, Yifeng Zhu, Chongkai Gao, Yihao Feng, Qiang Liu, Yuke Zhu, and Peter
  Stone.
\newblock {{LIBERO}: Benchmarking Knowledge Transfer for Lifelong Robot
  Learning}.
\newblock In \emph{Advances in Neural Information Processing Systems
  (NeurIPS)}, pages 44776--44791, 2023{\natexlab{a}}.

\bibitem[Liu et~al.(2025{\natexlab{a}})Liu, Liu, Liang, Li, Liu, Wang, Wan,
  Zhang, and Ouyang]{flowgrpo}
Jie Liu, Gongye Liu, Jiajun Liang, Yangguang Li, Jiaheng Liu, Xintao Wang,
  Pengfei Wan, Di Zhang, and Wanli Ouyang.
\newblock {{Flow-GRPO}: Training Flow Matching Models via Online {RL}}.
\newblock In \emph{Advances in Neural Information Processing Systems
  (NeurIPS)}, pages 40783--40818, 2025{\natexlab{a}}.

\bibitem[Liu et~al.(2023{\natexlab{b}})Liu, Gong, and Liu]{rectflow}
Xingchao Liu, Chengyue Gong, and Qiang Liu.
\newblock {Flow Straight and Fast: Learning to Generate and Transfer Data with
  Rectified Flow}.
\newblock In \emph{International Conference on Learning Representations
  (ICLR)}, 2023{\natexlab{b}}.

\bibitem[Liu et~al.(2025{\natexlab{b}})Liu, Xiao, Domingo-Enrich, Liu, and
  Zhang]{vggflow}
Zhen Liu, Tim~Z. Xiao, Carles Domingo-Enrich, Weiyang Liu, and Dinghuai Zhang.
\newblock {Value Gradient Guidance for Flow Matching Alignment}.
\newblock In \emph{Advances in Neural Information Processing Systems
  (NeurIPS)}, pages 11372--11404, 2025{\natexlab{b}}.

\bibitem[Lu et~al.(2023)Lu, Chen, Chen, Su, Li, and Zhu]{cep}
Cheng Lu, Huayu Chen, Jianfei Chen, Hang Su, Chongxuan Li, and Jun Zhu.
\newblock {Contrastive Energy Prediction for Exact Energy-Guided Diffusion
  Sampling in Offline Reinforcement Learning}.
\newblock In \emph{Proceedings of the 40th International Conference on Machine
  Learning (ICML)}, pages 22825--22855. PMLR, 2023.

\bibitem[Ma et~al.(2025)Ma, Tong, Jia, Hu, Su, Zhang, Yang, Li, Jaakkola, Jia,
  and Xie]{ma2025scaling}
Nanye Ma, Shangyuan Tong, Haolin Jia, Hexiang Hu, Yu-Chuan Su, Mingda Zhang,
  Xuan Yang, Yandong Li, Tommi Jaakkola, Xuhui Jia, and Saining Xie.
\newblock {Scaling Inference Time Compute for Diffusion Models}.
\newblock In \emph{Proceedings of the IEEE/CVF Conference on Computer Vision
  and Pattern Recognition (CVPR)}, pages 2523--2534, 2025.

\bibitem[Meng et~al.(2021)Meng, Song, Li, and Ermon]{meng2021highorder}
Chenlin Meng, Yang Song, Wenzhe Li, and Stefano Ermon.
\newblock {Estimating High Order Gradients of the Data Distribution by
  Denoising}.
\newblock In \emph{Advances in Neural Information Processing Systems
  (NeurIPS)}, pages 25359--25369, 2021.

\bibitem[Murray et~al.(2012)Murray, Marchesotti, and Perronnin]{ava}
Naila Murray, Luca Marchesotti, and Florent Perronnin.
\newblock {{AVA}: A Large-Scale Database for Aesthetic Visual Analysis}.
\newblock In \emph{Proceedings of the IEEE Conference on Computer Vision and
  Pattern Recognition (CVPR)}, pages 2408--2415, 2012.

\bibitem[Nakamoto et~al.(2024)Nakamoto, Mees, Kumar, and Levine]{vgps}
Mitsuhiko Nakamoto, Oier Mees, Aviral Kumar, and Sergey Levine.
\newblock {Steering Your Generalists: Improving Robotic Foundation Models via
  Value Guidance}.
\newblock In \emph{Proceedings of The 8th Conference on Robot Learning (CoRL)},
  pages 4996--5013. PMLR, 2024.

\bibitem[Nasiriany et~al.(2026)Nasiriany, Nasiriany, Maddukuri, and
  Zhu]{robocasa365}
Soroush Nasiriany, Sepehr Nasiriany, Abhiram Maddukuri, and Yuke Zhu.
\newblock {RoboCasa365: A Large-Scale Simulation Framework for Training and
  Benchmarking Generalist Robots}.
\newblock In \emph{International Conference on Learning Representations
  (ICLR)}, 2026.

\bibitem[Pan et~al.(2023)Pan, Yan, Liew, Feng, and Tan]{sag}
Jiachun Pan, Hanshu Yan, Jun~Hao Liew, Jiashi Feng, and Vincent Y.~F. Tan.
\newblock {Towards Accurate Guided Diffusion Sampling through Symplectic
  Adjoint Method}.
\newblock \emph{arXiv preprint arXiv:2312.12030}, 2023.

\bibitem[Patel et~al.(2025)Patel, Wen, Metaxas, and Yang]{flowchef}
Maitreya Patel, Song Wen, Dimitris~N. Metaxas, and Yezhou Yang.
\newblock {FlowChef: Steering of Rectified Flow Models for Controlled
  Generations}.
\newblock In \emph{Proceedings of the IEEE/CVF International Conference on
  Computer Vision (ICCV)}, pages 15308--15318, 2025.

\bibitem[Potaptchik et~al.(2026)Potaptchik, Saravanan, Mammadov, Prat, Albergo,
  and Teh]{mfm}
Peter Potaptchik, Adhi Saravanan, Abbas Mammadov, Alvaro Prat, Michael~S.
  Albergo, and Yee~Whye Teh.
\newblock {Meta Flow Maps Enable Scalable Reward Alignment}.
\newblock In \emph{Proceedings of the 43rd International Conference on Machine
  Learning (ICML)}, pages 99255--99295. PMLR, 2026.

\bibitem[Psenka et~al.(2024)Psenka, Escontrela, Abbeel, and Ma]{qsm}
Michael Psenka, Alejandro Escontrela, Pieter Abbeel, and Yi Ma.
\newblock {Learning a Diffusion Model Policy from Rewards via Q-Score
  Matching}.
\newblock In \emph{Proceedings of the 41st International Conference on Machine
  Learning (ICML)}, pages 41163--41182. PMLR, 2024.

\bibitem[Radford et~al.(2021)Radford, Kim, Hallacy, Ramesh, Goh, Agarwal,
  Sastry, Askell, Mishkin, Clark, Krueger, and Sutskever]{clip}
Alec Radford, Jong~Wook Kim, Chris Hallacy, Aditya Ramesh, Gabriel Goh,
  Sandhini Agarwal, Girish Sastry, Amanda Askell, Pamela Mishkin, Jack Clark,
  Gretchen Krueger, and Ilya Sutskever.
\newblock {Learning Transferable Visual Models From Natural Language
  Supervision}.
\newblock In \emph{Proceedings of the 38th International Conference on Machine
  Learning (ICML)}, pages 8748--8763. PMLR, 2021.

\bibitem[Ren et~al.(2025)Ren, Lidard, Ankile, Simeonov, Agrawal, Majumdar,
  Burchfiel, Dai, and Simchowitz]{dppo}
Allen~Z. Ren, Justin Lidard, Lars~L. Ankile, Anthony Simeonov, Pulkit Agrawal,
  Anirudha Majumdar, Benjamin Burchfiel, Hongkai Dai, and Max Simchowitz.
\newblock {Diffusion Policy Policy Optimization}.
\newblock In \emph{International Conference on Learning Representations
  (ICLR)}, 2025.

\bibitem[Ruan et~al.(2026)Ruan, Cao, Burger, Pesaranghader, Kamali, Kim,
  Vijaykumar, Aspuru-Guzik, Gilitschenski, and Rhinehart]{qpilots}
Yifan Ruan, Chenyang Cao, Andreas Burger, Ali Pesaranghader, Kaveh Kamali,
  Jaehong Kim, Nandita Vijaykumar, Al{\'a}n Aspuru-Guzik, Igor Gilitschenski,
  and Nicholas Rhinehart.
\newblock {{QPILOTS}: Efficient Test-Time Q-Steering for Flow Policies}.
\newblock \emph{arXiv preprint arXiv:2606.14801}, 2026.

\bibitem[Sabour et~al.(2026)Sabour, Albergo, Domingo-Enrich, Boffi, Fidler,
  Kreis, and Vanden-Eijnden]{fmtt}
Amirmojtaba Sabour, Michael~S. Albergo, Carles Domingo-Enrich, Nicholas~M.
  Boffi, Sanja Fidler, Karsten Kreis, and Eric Vanden-Eijnden.
\newblock {Test-time Scaling of Diffusions with Flow Maps}.
\newblock In \emph{Advances in Neural Information Processing Systems
  (NeurIPS)}, 2026.

\bibitem[Saharia et~al.(2022)Saharia, Chan, Saxena, Li, Whang, Denton,
  Ghasemipour, Gontijo~Lopes, Karagol~Ayan, Salimans, Ho, Fleet, and
  Norouzi]{imagen}
Chitwan Saharia, William Chan, Saurabh Saxena, Lala Li, Jay Whang, Emily~L.
  Denton, Seyed Kamyar~Seyed Ghasemipour, Raphael Gontijo~Lopes, Burcu
  Karagol~Ayan, Tim Salimans, Jonathan Ho, David~J. Fleet, and Mohammad
  Norouzi.
\newblock {Photorealistic Text-to-Image Diffusion Models with Deep Language
  Understanding}.
\newblock In \emph{Advances in Neural Information Processing Systems
  (NeurIPS)}, pages 36479--36494, 2022.

\bibitem[Schuhmann(2022)]{aesthetic}
Christoph Schuhmann.
\newblock {CLIP+MLP Aesthetic Score Predictor}.
\newblock
  \url{https://github.com/christophschuhmann/improved-aesthetic-predictor},
  2022.

\bibitem[Shen et~al.(2022)Shen, Hajiramezanali, Scalia, Tseng, Diamant,
  Biancalani, and Loukas]{mpcguidance}
Max~W. Shen, Ehsan Hajiramezanali, Gabriele Scalia, Alex Tseng, Nathaniel
  Diamant, Tommaso Biancalani, and Andreas Loukas.
\newblock {Conditional Diffusion with Less Explicit Guidance via Model
  Predictive Control}.
\newblock \emph{arXiv preprint arXiv:2210.12192}, 2022.

\bibitem[Shin et~al.(2026)Shin, Shin, Zhu, Guo, Chen, Lee, Choi, and Choi]{eam}
Jeongwoo Shin, Dongsoo Shin, Yuchen Zhu, Wei Guo, Yongxin Chen, Joonseok Lee,
  Jaewoong Choi, and Jaemoo Choi.
\newblock {Efficient Adjoint Matching for Fine-tuning Diffusion Models}.
\newblock In \emph{Advances in Neural Information Processing Systems
  (NeurIPS)}, 2026.

\bibitem[Shukor et~al.(2025)Shukor, Aubakirova, Capuano, Kooijmans, Palma,
  Zouitine, Aractingi, Pascal, Russi, Marafioti, Alibert, Cord, Wolf, and
  Cadene]{smolvla}
Mustafa Shukor, Dana Aubakirova, Francesco Capuano, Pepijn Kooijmans, Steven
  Palma, Adil Zouitine, Michel Aractingi, Caroline Pascal, Martino Russi,
  Andres Marafioti, Simon Alibert, Matthieu Cord, Thomas Wolf, and Remi Cadene.
\newblock {SmolVLA: A Vision-Language-Action Model for Affordable and Efficient
  Robotics}.
\newblock \emph{arXiv preprint arXiv:2506.01844}, 2025.

\bibitem[Singhal et~al.(2025)Singhal, Horvitz, Teehan, Ren, Yu, McKeown, and
  Ranganath]{fksteering}
Raghav Singhal, Zachary Horvitz, Ryan Teehan, Mengye Ren, Zhou Yu, Kathleen
  McKeown, and Rajesh Ranganath.
\newblock {A General Framework for Inference-time Scaling and Steering of
  Diffusion Models}.
\newblock In \emph{Proceedings of the 42nd International Conference on Machine
  Learning (ICML)}, pages 55810--55827. PMLR, 2025.

\bibitem[Song et~al.(2021{\natexlab{a}})Song, Meng, and Ermon]{ddim}
Jiaming Song, Chenlin Meng, and Stefano Ermon.
\newblock {Denoising Diffusion Implicit Models}.
\newblock In \emph{International Conference on Learning Representations
  (ICLR)}, 2021{\natexlab{a}}.

\bibitem[Song et~al.(2023)Song, Zhang, Yin, Mardani, Liu, Kautz, Chen, and
  Vahdat]{lgd}
Jiaming Song, Qinsheng Zhang, Hongxu Yin, Morteza Mardani, Ming-Yu Liu, Jan
  Kautz, Yongxin Chen, and Arash Vahdat.
\newblock {Loss-Guided Diffusion Models for Plug-and-Play Controllable
  Generation}.
\newblock In \emph{Proceedings of the 40th International Conference on Machine
  Learning (ICML)}, pages 32483--32498. PMLR, 2023.

\bibitem[Song et~al.(2021{\natexlab{b}})Song, Sohl-Dickstein, Kingma, Kumar,
  Ermon, and Poole]{scoresde}
Yang Song, Jascha Sohl-Dickstein, Diederik~P. Kingma, Abhishek Kumar, Stefano
  Ermon, and Ben Poole.
\newblock {Score-Based Generative Modeling through Stochastic Differential
  Equations}.
\newblock In \emph{International Conference on Learning Representations
  (ICLR)}, 2021{\natexlab{b}}.

\bibitem[{Stability AI}(2024)]{sd35}
{Stability AI}.
\newblock {Stable Diffusion 3.5}.
\newblock \url{https://github.com/Stability-AI/sd3.5}, 2024.

\bibitem[Tang et~al.(2025)Tang, Peng, Tang, Hong, Wang, and Chang]{dno}
Zhiwei Tang, Jiangweizhi Peng, Jiasheng Tang, Mingyi Hong, Fan Wang, and
  Tsung-Hui Chang.
\newblock {Inference-Time Alignment of Diffusion Models with Direct Noise
  Optimization}.
\newblock In \emph{Proceedings of the 42nd International Conference on Machine
  Learning (ICML)}, pages 58905--58930. PMLR, 2025.

\bibitem[Uehara et~al.(2024)Uehara, Zhao, Black, Hajiramezanali, Scalia,
  Diamant, Tseng, Biancalani, and Levine]{uehara2024}
Masatoshi Uehara, Yulai Zhao, Kevin Black, Ehsan Hajiramezanali, Gabriele
  Scalia, Nathaniel~Lee Diamant, Alex~M. Tseng, Tommaso Biancalani, and Sergey
  Levine.
\newblock {Fine-Tuning of Continuous-Time Diffusion Models as
  Entropy-Regularized Control}.
\newblock \emph{arXiv preprint arXiv:2402.15194}, 2024.

\bibitem[Wagenmaker et~al.(2025)Wagenmaker, Zhang, Nakamoto, Park, Yagoub,
  Nagabandi, Gupta, and Levine]{dsrl}
Andrew Wagenmaker, Yunchu Zhang, Mitsuhiko Nakamoto, Seohong Park, Waleed
  Yagoub, Anusha Nagabandi, Abhishek Gupta, and Sergey Levine.
\newblock {Steering Your Diffusion Policy with Latent Space Reinforcement
  Learning}.
\newblock In \emph{Proceedings of The 9th Conference on Robot Learning (CoRL)},
  pages 258--282. PMLR, 2025.

\bibitem[Wallace et~al.(2023)Wallace, Gokul, Ermon, and Naik]{doodl}
Bram Wallace, Akash Gokul, Stefano Ermon, and Nikhil Naik.
\newblock {End-to-End Diffusion Latent Optimization Improves Classifier
  Guidance}.
\newblock In \emph{Proceedings of the IEEE/CVF International Conference on
  Computer Vision (ICCV)}, pages 7280--7290, 2023.

\bibitem[Wallace et~al.(2024)Wallace, Dang, Rafailov, Zhou, Lou, Purushwalkam,
  Ermon, Xiong, Joty, and Naik]{diffusiondpo}
Bram Wallace, Meihua Dang, Rafael Rafailov, Linqi Zhou, Aaron Lou, Senthil
  Purushwalkam, Stefano Ermon, Caiming Xiong, Shafiq Joty, and Nikhil Naik.
\newblock {Diffusion Model Alignment Using Direct Preference Optimization}.
\newblock In \emph{Proceedings of the IEEE/CVF Conference on Computer Vision
  and Pattern Recognition (CVPR)}, pages 8228--8238, 2024.

\bibitem[Wang et~al.(2025)Wang, Cheng, Liao, Qu, and Liu]{ocflow}
Luran Wang, Chaoran Cheng, Yizhen Liao, Yanru Qu, and Ge Liu.
\newblock {Training Free Guided Flow-Matching with Optimal Control}.
\newblock In \emph{International Conference on Learning Representations
  (ICLR)}, 2025.

\bibitem[Wang et~al.(2023)Wang, Hunt, and Zhou]{diffusionql}
Zhendong Wang, Jonathan~J. Hunt, and Mingyuan Zhou.
\newblock {Diffusion Policies as an Expressive Policy Class for Offline
  Reinforcement Learning}.
\newblock In \emph{International Conference on Learning Representations
  (ICLR)}, 2023.

\bibitem[Wu et~al.(2023{\natexlab{a}})Wu, Trippe, Naesseth, Blei, and
  Cunningham]{tds}
Luhuan Wu, Brian~L. Trippe, Christian~A. Naesseth, David~M. Blei, and John~P.
  Cunningham.
\newblock {Practical and Asymptotically Exact Conditional Sampling in Diffusion
  Models}.
\newblock In \emph{Advances in Neural Information Processing Systems
  (NeurIPS)}, pages 31372--31403, 2023{\natexlab{a}}.

\bibitem[Wu et~al.(2023{\natexlab{b}})Wu, Hao, Sun, Chen, Zhu, Zhao, and
  Li]{hpsv2}
Xiaoshi Wu, Yiming Hao, Keqiang Sun, Yixiong Chen, Feng Zhu, Rui Zhao, and
  Hongsheng Li.
\newblock {Human Preference Score v2: A Solid Benchmark for Evaluating Human
  Preferences of Text-to-Image Synthesis}.
\newblock \emph{arXiv preprint arXiv:2306.09341}, 2023{\natexlab{b}}.

\bibitem[Xu et~al.(2023)Xu, Liu, Wu, Tong, Li, Ding, Tang, and
  Dong]{imagereward}
Jiazheng Xu, Xiao Liu, Yuchen Wu, Yuxuan Tong, Qinkai Li, Ming Ding, Jie Tang,
  and Yuxiao Dong.
\newblock {{ImageReward}: Learning and Evaluating Human Preferences for
  Text-to-Image Generation}.
\newblock In \emph{Advances in Neural Information Processing Systems
  (NeurIPS)}, pages 15903--15935, 2023.

\bibitem[Yang et~al.(2026)Yang, Jiang, Sheng, Bai, Yin, Ma, Liu, Sun, Lyu,
  Dong, and Tang]{gaf}
Liuhaichen Yang, Zhuang Jiang, Chenchao Sheng, Ningwei Bai, Qichen Yin, Hanbo
  Ma, Junkai Liu, Junkai Sun, Dongcheng Lyu, Yi Dong, and Zezhi Tang.
\newblock {Guided Action Flow: Value-Guided Sampling for Frozen
  Vision-Language-Action Policies}.
\newblock \emph{arXiv preprint arXiv:2607.02092}, 2026.

\bibitem[Ye et~al.(2024)Ye, Lin, Han, Xu, Liu, Liang, Ma, Zou, and Ermon]{tfg}
Haotian Ye, Haowei Lin, Jiaqi Han, Minkai Xu, Sheng Liu, Yitao Liang, Jianzhu
  Ma, James Zou, and Stefano Ermon.
\newblock {TFG: Unified Training-Free Guidance for Diffusion Models}.
\newblock In \emph{Advances in Neural Information Processing Systems
  (NeurIPS)}, pages 22370--22417, 2024.

\bibitem[Yu et~al.(2023)Yu, Wang, Zhao, Ghanem, and Zhang]{freedom}
Jiwen Yu, Yinhuai Wang, Chen Zhao, Bernard Ghanem, and Jian Zhang.
\newblock {{FreeDoM}: Training-Free Energy-Guided Conditional Diffusion Model}.
\newblock In \emph{Proceedings of the IEEE/CVF International Conference on
  Computer Vision (ICCV)}, pages 23174--23184, 2023.

\bibitem[Zheng et~al.(2026)Zheng, Chen, Ye, Wang, Zhang, Jiang, Su, Ermon, Zhu,
  and Liu]{diffusionnft}
Kaiwen Zheng, Huayu Chen, Haotian Ye, Haoxiang Wang, Qinsheng Zhang, Kai Jiang,
  Hang Su, Stefano Ermon, Jun Zhu, and Ming-Yu Liu.
\newblock {{DiffusionNFT}: Online Diffusion Reinforcement with Forward
  Process}.
\newblock In \emph{International Conference on Learning Representations
  (ICLR)}, 2026.

\bibitem[Zhou et~al.(2026)Zhou, Peng, Xu, Li, Springenberg, Frans, and
  Levine]{qgf}
Zhiyuan Zhou, Andy Peng, Charles Xu, Qiyang Li, Jost~Tobias Springenberg, Kevin
  Frans, and Sergey Levine.
\newblock {Test-Time Gradient Guidance of Flow Policies in Reinforcement
  Learning}.
\newblock In \emph{Advances in Neural Information Processing Systems
  (NeurIPS)}, 2026.

\bibitem[Ziakas et~al.(2026)Ziakas, Russo, and Bose]{fmq}
Christos Ziakas, Alessandra Russo, and Avishek~Joey Bose.
\newblock {Aligning Flow Map Policies with Optimal {$Q$}-Guidance}.
\newblock \emph{arXiv preprint arXiv:2605.12416}, 2026.

\end{thebibliography}
}

\clearpage
\ifpreprint\else\setcounter{page}{1}\fi
\appendix
\onecolumn
\raggedbottom %
\setcounter{topnumber}{4}\setcounter{totalnumber}{6}
\renewcommand{\topfraction}{0.95}\renewcommand{\textfraction}{0.05}
\renewcommand{\floatpagefraction}{0.75}
\startcontents[app]
\begin{center}
{\Large\textbf{\thetitle}\\[0.5em]Supplementary Material}
\end{center}
\vspace{0.6em}
{\centering\appcontents\par}
\vspace{1em}

\section{\old{Related Work}}
\label{app:related}

\old{\Cref{tab:paradigms} summarizes the state at which each estimator evaluates the reward or a value, and its cost per step and in training.}

\begin{table}[h]
\centering\footnotesize
\caption{\textbf{Evaluation state and cost of each estimator.} G: generator, D: decoder, R: reward, V: \old{value model}, H: head\old{, F: flow map}.}
\label{tab:paradigms}
\vspace{-0.5em}
\setlength{\tabcolsep}{2.2pt}
\old{\begin{tabular}{@{}lcccc@{}}
\toprule
 & Value & Backprop & & One for\\
Method & at & per step & Training & all rewards\\
\midrule
\multicolumn{5}{@{}l}{\textit{Training-free}}\\
DPS, FreeDoM~\cite{dps,freedom} & $\hat\vz_{0\mid t}$ & G, D, R & -- & --\\
\old{QGF, FlowChef~\cite{qgf,flowchef}} & $\hat\vz_{0\mid t}$ & \old{D, R ($D_t{=}I$)} & -- & --\\
\old{DOODL, D-Flow~\cite{doodl,dflow}} & $\vz_0$ & \old{G (all steps), D, R} & -- & --\\
\midrule
\multicolumn{5}{@{}l}{\textit{Learned}}\\
StitchVM~\cite{stitchvm} & $\vz_t$ & V & V & \xmark\\
AGF~\cite{agf} & $\vz_0$ & -- & \makecell{H, backprop\\through all steps} & \xmark\\
\old{Flow map~\cite{fmrg,diamondmaps}} & $\vz_0$ & \old{F, D, R} & \old{F, distillation} & \cmark\\
\midrule
\rowcolor{oursrow}\textbf{\method} (Ours) & $\vz_s$ & V, G (VJP) & -- & --\\
\rowcolor{oursrow}\textbf{\methodfast} (Ours) & $\vz_s$ & V & \makecell{H, backprop\\through one step} & \cmark\\
\bottomrule
\end{tabular}}
\vspace{-1em}
\end{table}

\paragraph{\old{Training-free guidance.}} \old{DPS~\cite{dps} differentiates the reward of the posterior mean, and FreeDoM~\cite{freedom} and universal guidance~\cite{universalguidance} build on the same one-step estimate with additional recurrence or optimization\old{, and LGD~\cite{lgd} reduces its bias with Monte Carlo samples around it}. TFG~\cite{tfg} unifies such training-free methods in one hyperparameter space and searches it per task, and its analysis reads the gradient through the posterior mean as preconditioned by the posterior covariance, as in \cref{eq:dcov}. \old{Dandapanthula and Boffi~\cite{tilting} trace the over-optimization of plug-in reward guidance to a finite-particle approximation of the Doob $h$-function, already for Gaussian and Gaussian mixture targets.} Our unification differs in kind: a one-parameter family indexed by the lookahead ratio, whose two ends are learned-value guidance and DPS, }with an error analysis that relates the ratio to the posterior mismatch behind the value-model error and to the bias of the step (\cref{sec:correction}).

\paragraph{\old{Lookahead guidance.}} \old{Several methods evaluate the guide after simulating sampler steps.} \old{At one extreme, DOODL~\cite{doodl} and D-Flow~\cite{dflow} optimize the initial noise by differentiating the reward of the final sample through every step, OC-Flow~\cite{ocflow} casts this as optimal control, and ReNO~\cite{reno} and DNO~\cite{dno} optimize the noise against preference rewards.} \old{SAG~\cite{sag} estimates the clean sample with a multi-step solver and backpropagates through it with a symplectic adjoint. MPC guidance~\cite{mpcguidance} simulates five to ten unconditional steps to a less noisy state, evaluates a noised classifier or a conditional diffusion model there, and backpropagates through all simulated steps, at a memory cost linear in their number. LiDAR~\cite{lidar} draws lookahead samples and derives guidance without backpropagation. \method shares with MPC guidance the evaluation at a less noisy state, but it uses a learned reward value instead of a classifier \old{or a conditional diffusion model}, takes one deterministic step so that the pullback reduces to one correction term, shows that this term needs the transposed Jacobian under CFG, and distills this term into a head. REG~\cite{reg} multiplies the gradient of existing guidance element-wise by a correction derived from the denoiser Jacobian, which amounts to a diagonal, hence symmetric, approximation of that Jacobian\old{. On} SD3.5, the symmetric part alone \old{loses part of the HPSv2 of the transpose (\cref{tab:ablation_full})}.} \old{QGF~\cite{qgf} and FlowChef~\cite{flowchef} replace this Jacobian with the identity, and QGF reports that including it hurts in offline reinforcement learning. The pullback $P_s=\tfrac{s}{t}I+\big(1-\tfrac{s}{t}\big)D_t$ of \method instead shrinks it toward the identity, and dropping the correction lowers HPSv2 on SD3.5 (\cref{tab:ablation}).}

\paragraph{\old{Flow-map lookahead.}} \old{A flow map jumps from $\vz_t$ toward the clean endpoint in one or a few evaluations, and recent methods evaluate the reward there. FMTT~\cite{fmtt} tilts trajectories with this lookahead and importance weights, and FMRG~\cite{fmrg} derives guidance from deterministic optimal control and pulls the reward gradient at the endpoint back through the flow map, the gradient that AGF regresses (\cref{app:prop-agf}). Diamond Maps~\cite{diamondmaps} and meta flow maps~\cite{mfm} instead average the \old{exponentiated} rewards of posterior samples drawn by a stochastic flow map, which estimates the \old{soft} value consistently but requires \old{learning} such a map. \method needs no flow map. It takes one step of the frozen generator to an intermediate state and evaluates a learned value there, so that the lookahead ratio trades the value-model error against the bias of the step.}

\paragraph{\old{Learned values and amortized guidance.}} \old{Classifier guidance~\cite{dhariwal2021} differentiates a classifier trained on noisy images, and StitchVM~\cite{stitchvm} stitches the denoiser and the reward model into a value model on noisy latents. \old{In offline reinforcement learning, CEP~\cite{cep} trains the intermediate energy with a contrastive objective that recovers the exact guidance given unlimited capacity and data.} SVDD~\cite{svdd} decodes with soft values without derivatives, FK steering~\cite{fksteering} \old{and twisted SMC~\cite{tds} resample} interacting particles with reward potentials on intermediate states, and actor-critic fine-tuning learns a critic on noisy states that can also steer sampling~\cite{ecg}. AGF~\cite{agf} amortizes guidance by regressing adjoint gradients along sampling paths onto a head. DEFT~\cite{deft} keeps the unconditional model frozen and fine-tunes a small network to learn the conditional $h$-transform. \methodfast likewise trains only a small head with the generator frozen, but its target is the known one-step correction, which depends on the reward only through $\vg_s$, so one head per generator serves all rewards (\cref{tab:share}).}

\paragraph{\old{Reward fine-tuning.}} \old{The generator can instead be fine-tuned on the reward, by backpropagation through sampling~\cite{draft,imagereward}, reinforcement learning~\cite{ddpo,dpok,flowgrpo,diffusionnft,vard}, \old{\old{direct preference optimization on human comparisons}~\cite{diffusiondpo},} stochastic optimal control~\cite{adjointmatching,eam}, or value-gradient matching~\cite{vggflow}. \method leaves the generator frozen and acts only at inference.}

\paragraph{\old{Critic guidance of robot policies.}} \old{Diffuser~\cite{diffuser} guides trajectory sampling with a return model trained on noisy trajectories, and V-GPS~\cite{vgps} re-ranks the actions of a frozen generalist policy with a learned value, which Q-BoN follows. QGF~\cite{qgf} differentiates a critic at the denoised action with the Jacobian of the policy replaced by the identity\old{, QPILOTS~\cite{qpilots} also evaluates the critic at a projected clean action, and Guided Action Flow~\cite{gaf} steers a frozen VLA policy with a small critic on a real robot}. AGF~\cite{agf} amortizes the costate into a guidance network trained per task. Training-based alternatives backpropagate the critic through the denoising chain~\cite{diffusionql}, match the policy score to the critic gradient~\cite{qsm}, fine-tune the policy~\cite{dppo,qam,fmq}, or learn a policy over the latent noise of a frozen policy~\cite{dsrl}. Both variants of \method keep the policy frozen.}

\section{\old{Proofs}}
\label{app:proofs}

\old{Throughout, $k_t(\vz\mid \vz_0)=\mathcal N\big(\vz;(1-t)\vz_0,t^2I\big)$ is the forward kernel, and moments are taken under the posterior $p(\cdot\mid \vz_t=\vz)$ unless stated otherwise. Restated results keep their numbers from the main text, and results stated only here are numbered by appendix section.}

\subsection{\old{Proof of \texorpdfstring{\Cref{lem:cov}}{Lemma 1}}}
\label{app:proof-lem}

\begin{restated}{\Cref{lem:cov}}
Let $\vz_t=(1-t)\vz_0+t\veps$ with $t\in(0,1)$, $\veps\sim\mathcal N(0,I)$ independent of $\vz_0$, and $\vz_0$ drawn from any distribution with finite second moment, and let $r$ have finite variance. Then the value $V_t(\vz)=\E[r(\vz_0)\mid\vz_t=\vz]$ and the Jacobian $D_t$ of the exact posterior mean satisfy
\begin{align}
\nabla V_t(\vz)&=\tfrac{1-t}{t^2}\,\Cov\big[\vz_0,\,r(\vz_0)\mid\vz_t=\vz\big],\tag{\ref*{eq:cov}}\\
D_t(\vz)&=\tfrac{1-t}{t^2}\,\Cov\big[\vz_0\mid\vz_t=\vz\big].\tag{\ref*{eq:dcov}}
\end{align}
\end{restated}

\proofhead \pfirst{(i) Posterior score.} \old{Since $p(\vz_0\mid\vz_t=\vz)\propto p(\vz_0)\,k_t(\vz\mid\vz_0)$,}
\begin{equation}
\nabla_\vz\log p(\vz_0\mid\vz_t=\vz)=\frac{(1-t)\vz_0-\vz}{t^2}-\nabla_\vz\log p_t(\vz).
\end{equation}
\old{The posterior expectation of this score is zero, so $\nabla_\vz\log p_t(\vz)=\big((1-t)\hat\vz_{0\mid t}-\vz\big)/t^2$. Therefore,}
\begin{equation}
\nabla_\vz\log p(\vz_0\mid\vz_t=\vz)=\tfrac{1-t}{t^2}\big(\vz_0-\hat\vz_{0\mid t}\big).
\end{equation}

\pstep{(ii) Value gradient.} \old{Differentiating $V_t(\vz)=\int r(\vz_0)\,p(\vz_0\mid\vz_t=\vz)\,d\vz_0$ under the integral and using (i),}
\begin{align}
\nabla V_t(\vz)&=\tfrac{1-t}{t^2}\,\E\big[r(\vz_0)\big(\vz_0-\hat\vz_{0\mid t}\big)\mid\vz_t=\vz\big]\nonumber\\
&=\tfrac{1-t}{t^2}\,\E\big[\big(r(\vz_0)-V_t(\vz)\big)\big(\vz_0-\hat\vz_{0\mid t}\big)\mid\vz_t=\vz\big],
\end{align}
\old{where the second line uses $\E[\vz_0-\hat\vz_{0\mid t}\mid\vz_t]=0$. Therefore, \cref{eq:cov} holds.}

\pstep{(iii) Jacobian of the posterior mean.} \old{Applying (ii) to each coordinate $r(\vz_0)=z_{0,i}$, which has finite variance by the moment assumption, gives the second-order Tweedie identity~\cite{meng2021highorder},}
\begin{equation}
\nabla_\vz\hat z_{0\mid t,i}=\tfrac{1-t}{t^2}\,\Cov\big[\vz_0,z_{0,i}\mid\vz_t=\vz\big].
\end{equation}
\old{Therefore, \cref{eq:dcov} holds, and $D_t$ is symmetric.}\qedmark

\subsection{\old{Errors of DPS, a Learned Value, and AGF}}
\label{app:errors}

\old{This subsection states and proves the three error expressions of \cref{sec:motivation}.}

\begin{appresult}[DPS error]{Proposition}\label[proposition]{app:prop-dps}
\old{For the exact denoiser of $p$, DPS satisfies}
\begin{equation}
\vg^{\mathrm{DPS}}_t-\nabla V_t=-\tfrac{1-t}{t^2}\,\Cov\big[\vz_0,\,r(\vz_0)-\bar r(\vz_0)\mid\vz_t\big],\tag{\ref*{eq:dpserr}}
\end{equation}
\old{where $\bar r$ is the first-order Taylor expansion of $r$ at $\hat\vz_{0\mid t}$.}
\end{appresult}

\proofhead \old{By \cref{eq:dps,eq:dcov},}
\begin{equation}
\vg^{\mathrm{DPS}}_t=\tfrac{1-t}{t^2}\Cov[\vz_0\mid\vz_t]\,\nabla r(\hat\vz_{0\mid t})=\tfrac{1-t}{t^2}\Cov\big[\vz_0,\,\bar r(\vz_0)\mid\vz_t\big],
\end{equation}
\old{since $\bar r(\vz_0)=r(\hat\vz_{0\mid t})+\nabla r(\hat\vz_{0\mid t})^{\!\top}(\vz_0-\hat\vz_{0\mid t})$ differs from $\nabla r(\hat\vz_{0\mid t})^{\!\top}\vz_0$ by a constant. Subtracting \cref{eq:cov}, the error is the posterior covariance of $\vz_0$ with the nonlinear part $r-\bar r$ of the reward. Therefore, \cref{eq:dpserr} holds.}\qedmark

\begin{appresult}[Learned-value error]{Proposition}\label[proposition]{app:prop-svm}
\old{A value model regressed onto the reward of \old{samples from $q$ given their forward-noised versions} converges to $V^q_t(\vz)=\E_q[r(\vz_0)\mid\vz_t=\vz]$, so that $\vg^{\mathrm{SVM}}_t=\nabla V^q_t$ satisfies}
\begin{equation}
\vg^{\mathrm{SVM}}_t-\nabla V_t=\tfrac{1-t}{t^2}\big(\Cov_q-\Cov_p\big)\big[\vz_0,\,r(\vz_0)\mid\vz_t\big].\tag{\ref*{eq:svmerr}}
\end{equation}
\end{appresult}

\proofhead \pfirst{(i) Regression target.} \old{Over all functions of $(\vz_t,t)$, the loss $\E_q\|V(\vz_t,t)-r(\vz_0)\|^2$ with $\vz_0\sim q$ and $\vz_t\sim k_t(\cdot\mid\vz_0)$ is minimized by the conditional mean under $q$. Therefore, $\vg^{\mathrm{SVM}}_t=\nabla V^q_t$.}

\pstep{(ii) Error.} \old{\Cref{lem:cov} holds for any distribution of $\vz_0$, so it applies under $q$ and under $p$,}
\begin{align}
\nabla V^q_t&=\tfrac{1-t}{t^2}\Cov_q\big[\vz_0,r(\vz_0)\mid\vz_t\big],\\
\nabla V_t&=\tfrac{1-t}{t^2}\Cov_p\big[\vz_0,r(\vz_0)\mid\vz_t\big].
\end{align}
\old{Therefore, subtracting the two, \cref{eq:svmerr} holds.}\qedmark

\begin{remark}
\old{StitchVM regresses reward-model features rather than the reward itself, so \cref{app:prop-svm} applies to rewards that are affine in these features.}
\end{remark}

\begin{appresult}[AGF gap]{Proposition}\label[proposition]{app:prop-agf}
\old{For a linear reward $r(\vz_0)=\bm w^{\!\top}\vz_0$, AGF with a head that matches its target at zero guidance, $\vg^{\mathrm{AGF}}_t=\big(\tfrac{\partial\vz_0}{\partial\vz_t}\big)^{\!\top}\nabla r(\vz_0)$ with $\vz_0=\Psi_t(\vz_t)$ the endpoint of the probability-flow ODE of $p$ from $t$ to $0$, satisfies}
\begin{equation}
\vg^{\mathrm{AGF}}_t-\nabla V_t=\Big(\frac{\partial\vz_0}{\partial\vz_t}-D_t\Big)^{\!\top}\bm w.\tag{\ref*{eq:agfgap}}
\end{equation}
\end{appresult}

\proofhead \pfirst{(i) Value gradient.} \old{For a linear reward, $\Cov[\vz_0,\bm w^{\!\top}\vz_0\mid\vz_t]=\Cov[\vz_0\mid\vz_t]\,\bm w$, so \cref{eq:cov,eq:dcov} give}
\begin{equation}
\nabla V_t=\tfrac{1-t}{t^2}\Cov[\vz_0\mid\vz_t]\,\bm w=D_t\bm w=D_t^{\!\top}\bm w,
\end{equation}
\old{where the last equality uses the symmetry of $D_t$. Therefore, $\nabla V_t=D_t^{\!\top}\bm w$.}

\pstep{(ii) Gap.} \old{Since $\nabla r=\bm w$ at every point, $\vg^{\mathrm{AGF}}_t=\big(\tfrac{\partial\vz_0}{\partial\vz_t}\big)^{\!\top}\bm w$. Therefore, subtracting (i), \cref{eq:agfgap} holds.}\qedmark

\begin{remark}[path form]
\old{Along the path $\vz_\tau$ from $\tau=t$ to $0$, the sensitivity obeys $\frac{d}{d\tau}\frac{\partial\vz_\tau}{\partial\vz_t}=J_\vv(\vz_\tau,\tau)\frac{\partial\vz_\tau}{\partial\vz_t}$ with $J_\vv=(I-D_\tau)/\tau$, so that}
\begin{equation}
\frac{\partial\vz_0}{\partial\vz_t}=\mathcal T\exp\Big(-\int_0^t\frac{I-D_\tau(\vz_\tau)}{\tau}\,d\tau\Big),
\end{equation}
\old{a time-ordered product of the Jacobians $D_\tau$ along this one path, each a scaled posterior covariance by \cref{eq:dcov}, whereas $\nabla V_t$ involves only $D_t$ at $\vz_t$.}
\end{remark}

\begin{appresult}[Gaussian data]{Corollary}\label[corollary]{app:cor-gauss}
\old{If moreover $p=\mathcal N(\bm\mu,\sigma^2 I)$, then $\vg^{\mathrm{AGF}}_t=\rho_t\nabla V_t$ with $\rho_t=\sqrt{v_t}/\big((1-t)\sigma\big)$ and $v_t=(1-t)^2\sigma^2+t^2$, so that $\rho_t\ge1$, $\rho_0=1$, and $\rho_t\to\infty$ as $t\to1$.}
\end{appresult}

\proofhead \pfirst{(i) Posterior mean.} \old{The noisy state is $\vz_t\sim\mathcal N(\vm_t,v_tI)$ with $\vm_t=(1-t)\bm\mu$, and}
\begin{equation}
\hat\vz_{0\mid t}=\bm\mu+\tfrac{(1-t)\sigma^2}{v_t}(\vz_t-\vm_t),\qquad D_t=\tfrac{(1-t)\sigma^2}{v_t}I.
\end{equation}
\old{Therefore, $\nabla V_t=\frac{(1-t)\sigma^2}{v_t}\bm w$ by (i) of \cref{app:prop-agf}.}

\pstep{(ii) Flow map.} \old{The optimal velocity of \cref{eq:cfm} is $\E[\veps-\vz_0\mid\vz_t]$, which for this $p$ is}
\begin{equation}
\vv(\vz,t)=-\bm\mu+\frac{t-(1-t)\sigma^2}{v_t}\,(\vz-\vm_t)=\dot\vm_t+\frac{\dot v_t}{2v_t}\,(\vz-\vm_t),
\end{equation}
\old{so the ODE keeps $(\vz-\vm_t)/\sqrt{v_t}$ constant, and}
\begin{equation}
\Psi_t(\vz)=\bm\mu+\tfrac{\sigma}{\sqrt{v_t}}(\vz-\vm_t),\qquad \tfrac{\partial\vz_0}{\partial\vz_t}=\tfrac{\sigma}{\sqrt{v_t}}I.
\end{equation}
\old{Therefore, $\vg^{\mathrm{AGF}}_t=\frac{\sigma}{\sqrt{v_t}}\bm w$.}

\pstep{(iii) Ratio.} \old{Dividing (ii) by (i),}
\begin{equation}
\vg^{\mathrm{AGF}}_t=\rho_t\nabla V_t,\qquad \rho_t=\frac{\sigma/\sqrt{v_t}}{(1-t)\sigma^2/v_t}=\frac{\sqrt{v_t}}{(1-t)\sigma}.
\end{equation}
\old{Since $v_t\ge(1-t)^2\sigma^2$ with equality at $t=0$, and $v_t\to1$ while $(1-t)\sigma\to0$ as $t\to1$, the ratio is at least one, equals one at $t=0$, and diverges as $t\to1$. Therefore, \cref{app:cor-gauss} holds.}\qedmark

\subsection{\old{Proof of \texorpdfstring{\Cref{prop:pullback}}{Proposition 1}}}
\label{app:proof-pullback}

\begin{restated}{\Cref{prop:pullback}}
Let $V_\psi$ and $\vv_\theta$ be differentiable. Then
\begin{equation}
\vg^{\mathrm{VaLoP}}_t=\underbrace{\tfrac{s}{t}\,\vg_s}_{\text{learned-value form}}+\underbrace{\big(1-\tfrac{s}{t}\big)\,D_t^{\!\top}\vg_s}_{\text{DPS form}}.\tag{\ref*{eq:decomp}}
\end{equation}
In particular, $\vg^{\mathrm{VaLoP}}_t=\vg^{\mathrm{SVM}}_t$ at $s=t$, and if $V_\psi(\cdot,0)=r$, then $\vg^{\mathrm{VaLoP}}_t=\vg^{\mathrm{DPS}}_t$ at $s=0$.
\end{restated}

\proofhead \pfirst{(i) Jacobian of the step.} \old{By \cref{eq:step}, $\Phi_{t\to s}(\vz)=\vz-(t-s)\vv_\theta(\vz,t)$, and $D_t=I-tJ_\vv$ since $\hat\vz_{0\mid t}=\vz-t\vv_\theta$. Therefore, the step has the Jacobian}
\begin{equation}
P_s:=\partial_\vz\Phi_{t\to s}(\vz_t)=I-(t-s)J_\vv=\tfrac{s}{t}I+\big(1-\tfrac{s}{t}\big)D_t .
\end{equation}

\pstep{(ii) Decomposition.} \old{Since $\vg^{\mathrm{VaLoP}}_t$ is the gradient of $V_\psi(\cdot,s)\circ\Phi_{t\to s}$ at $\vz_t$ and $\vz_s=\Phi_{t\to s}(\vz_t)$, the chain rule and (i) give}
\begin{align}
\vg^{\mathrm{VaLoP}}_t=P_s^{\!\top}\vg_s&=\tfrac{s}{t}\vg_s+\big(1-\tfrac{s}{t}\big)D_t^{\!\top}\vg_s\nonumber\\
&=\vg_s-(t-s)J_\vv^{\!\top}\vg_s .
\end{align}
\old{Therefore, \cref{eq:valop,eq:decomp} hold.}

\pstep{(iii) The two ends.} \old{At $s=t$, $\Phi_{t\to t}$ is the identity, $\vz_s=\vz_t$, and the second term of \cref{eq:decomp} vanishes. At $s=0$, $\vz_s=\hat\vz_{0\mid t}$, the first term vanishes, and $V_\psi(\cdot,0)=r$. Hence\old{,}}
\begin{align}
s=t:&\quad \vg^{\mathrm{VaLoP}}_t=\nabla V_\psi(\vz_t,t)=\vg^{\mathrm{SVM}}_t,\\
s=0:&\quad \vg^{\mathrm{VaLoP}}_t=D_t^{\!\top}\nabla r(\hat\vz_{0\mid t})=\vg^{\mathrm{DPS}}_t,
\end{align}
\old{by \cref{eq:svm,eq:dps}. Therefore, \method coincides with learned-value guidance at $s=t$ and with DPS at $s=0$.}\qedmark

\subsection{\old{Proof of \texorpdfstring{\Cref{prop:mismatch}}{Proposition 2}}}
\label{app:proof-mismatch}

\begin{restated}{\Cref{prop:mismatch}}
Let $\mathrm{KL}(p\,\|\,q)<\infty$, $|r|\le B$, and $q_s$ be the marginal of $\vz_s$ under $q$, and let $\Delta_s=\E_{\vz_s\sim p_s}\mathrm{KL}\big(p(\vz_0\mid\vz_s)\,\|\,q(\vz_0\mid\vz_s)\big)$ be the posterior mismatch at level $s$. Then
\begin{equation}
\Delta_s=\mathrm{KL}(p\,\|\,q)-\mathrm{KL}(p_s\,\|\,q_s),\tag{\ref*{eq:mismatch}}
\end{equation}
which is nondecreasing in $s$ with $\Delta_0=0$ and $\Delta_1=\mathrm{KL}(p\,\|\,q)$, and $\E_{\vz_s\sim p_s}\big|V^q_s(\vz_s)-V_s(\vz_s)\big|\le B\sqrt{2\Delta_s}$.
\end{restated}

\proofhead \pfirst{(i) Chain rule.} \old{Under $p$ and $q$, the pair $(\vz_0,\vz_s)$ has the same forward kernel $k_s(\vz_s\mid\vz_0)$, which adds no divergence. Factorizing the divergence of the joint distributions through $\vz_0$ and through $\vz_s$,}
\begin{align}
\mathrm{KL}\big(p(\vz_0,\vz_s)\,\|\,q(\vz_0,\vz_s)\big)&=\mathrm{KL}(p\,\|\,q)\nonumber\\
&=\mathrm{KL}(p_s\,\|\,q_s)+\Delta_s .
\end{align}
\old{Therefore, \cref{eq:mismatch} holds.}

\pstep{(ii) Monotonicity.} \old{Let $s\le t$ with $s<1$, $a=\frac{1-t}{1-s}$, and $c^2=t^2-a^2s^2$, which is nonnegative since $t(1-s)\ge s(1-t)$. For $\veps'\sim\mathcal N(0,I)$ independent of $\vz_s$,}
\begin{equation}
a\vz_s+c\,\veps'\;\big|\;\vz_0\;\sim\;\mathcal N\big((1-t)\vz_0,\,t^2I\big),
\end{equation}
\old{the law of $\vz_t$ given $\vz_0$. Hence\old{,} $p_t$ and $q_t$ arise from $p_s$ and $q_s$ through one Markov kernel, and the data processing inequality and (i) give}
\begin{equation}
\mathrm{KL}(p_t\,\|\,q_t)\le\mathrm{KL}(p_s\,\|\,q_s)\quad\Longrightarrow\quad\Delta_s\le\Delta_t .
\end{equation}
\old{At $s=0$, $p_0=p$ and $q_0=q$, and at $s=1$, $p_1=q_1=\mathcal N(0,I)$. Therefore, $\Delta_s$ is nondecreasing in $s$ with $\Delta_0=0$ and $\Delta_1=\mathrm{KL}(p\,\|\,q)$.}

\pstep{(iii) Value bound.} \old{At a fixed $\vz_s$, $|r|\le B$ and Pinsker's inequality give}
\begin{align}
\big|V^q_s(\vz_s)-V_s(\vz_s)\big|&=\big|\E_q[r(\vz_0)\mid\vz_s]-\E_p[r(\vz_0)\mid\vz_s]\big|\nonumber\\
&\le 2B\,\mathrm{TV}\big(p(\cdot\mid\vz_s),q(\cdot\mid\vz_s)\big)\nonumber\\
&\le B\sqrt{2\,\mathrm{KL}\big(p(\cdot\mid\vz_s)\,\|\,q(\cdot\mid\vz_s)\big)} .
\end{align}
\old{Averaging over $p_s$ and applying Jensen's inequality to the square root,}
\begin{equation}
\E_{\vz_s\sim p_s}\big|V^q_s(\vz_s)-V_s(\vz_s)\big|\le B\sqrt{2\Delta_s}.
\end{equation}
\old{Therefore, the value under $q$ deviates from that under $p$ by at most $B\sqrt{2\Delta_s}$ on average at level $s$.}\qedmark

\subsection{\old{Error Split and the Correction}}
\label{app:split}

\old{This subsection proves the statements of \cref{sec:correction} on \cref{eq:split} and on the correction.}

\begin{appresult}[ends of the split]{Lemma}\label[lemma]{app:lem-ends}
\old{The second term of \cref{eq:split} \old{is zero} at $s=t$ and \old{is} the DPS error $\vg^{\mathrm{DPS}}_t-\nabla V_t$ at $s=0$.}
\end{appresult}

\proofhead \old{At $s=t$, $P_t=I$ and $\vz_s=\vz_t$, so the second bracket of \cref{eq:split} is $\nabla V_t(\vz_t)-\nabla V_t(\vz_t)=0$. At $s=0$, $P_0=I-tJ_\vv=D_t$, $\vz_s=\hat\vz_{0\mid t}$, and $V_0=r$ by \cref{eq:values}, so}
\begin{align}
P_0^{\!\top}\nabla V_0(\hat\vz_{0\mid t})-\nabla V_t(\vz_t)&=D_t^{\!\top}\nabla r(\hat\vz_{0\mid t})-\nabla V_t(\vz_t)\nonumber\\
&=\vg^{\mathrm{DPS}}_t-\nabla V_t(\vz_t).
\end{align}
\old{Therefore, \cref{app:lem-ends} holds.}\qedmark

\begin{appresult}[correction for a linear reward]{Lemma}\label[lemma]{app:lem-linear}
\old{Let $r(\vz_0)=\bm w^{\!\top}\vz_0$, $V_\psi(\cdot,0)=r$, and the denoiser be exact. At $s=0$, \method is exact, $\vg^{\mathrm{VaLoP}}_t=\nabla V_t$, whereas the uncorrected $\vg_s$ has the error $(I-D_t)^{\!\top}\bm w$, which tends to $\bm w$ as $t\to1$.}
\end{appresult}

\proofhead \pfirst{(i) Exact value gradient.} \old{For the exact denoiser, $V_t(\vz_t)=\bm w^{\!\top}\hat\vz_{0\mid t}$. Therefore, $\nabla V_t=D_t^{\!\top}\bm w$.}

\pstep{(ii) With and without the correction.} \old{At $s=0$, \cref{prop:pullback} and $\nabla r=\bm w$ give}
\begin{align}
\vg^{\mathrm{VaLoP}}_t&=D_t^{\!\top}\bm w=\nabla V_t,\\
\vg_s-\nabla V_t&=\bm w-D_t^{\!\top}\bm w=(I-D_t)^{\!\top}\bm w .
\end{align}
\old{Therefore, the corrected guidance is exact at $s=0$, and the uncorrected one has the error $(I-D_t)^{\!\top}\bm w$.}

\pstep{(iii) High noise.} \old{By \cref{eq:dcov}, $D_t=\frac{1-t}{t^2}\Cov[\vz_0\mid\vz_t]$, which vanishes as $t\to1$, where the posterior approaches the prior and its covariance stays finite. Therefore, the error of $\vg_s$ tends to $\bm w$.}\qedmark

\paragraph{\old{Numerical check for intermediate ratios.}} \old{For $\vz_0\sim\mathcal N(\bm\mu,\Sigma)$, the exact denoiser has}
\begin{equation}
D_t=(1-t)\Sigma\big((1-t)^2\Sigma+t^2I\big)^{-1},
\end{equation}
\old{and with a linear reward and the sampler step of \cref{eq:step}, all three gradients are functions of $\Sigma$ applied to $\bm w$, namely $\nabla V_t=D_t\bm w$, the uncorrected $\vg_s=D_s\bm w$, and the corrected $P_s^{\!\top}\vg_s=P_sD_s\bm w$. Over $10^4$ draws of an eight-dimensional $\Sigma$ with eigenvalues log-uniform in $[0.01,4]$, a Gaussian $\bm w$, $t\in[0.05,0.99]$, and $s/t\in[0.05,0.95]$, the corrected gradient is at least as aligned with $\nabla V_t$ as the uncorrected one in every draw, with mean cosine $0.986$ against $0.932$.}

\section{\old{Scope and Limitations}}
\label{app:limits}

\paragraph{\old{Requirements.}} \old{\method requires a value model for the generator's latent space, such as that of StitchVM,} \old{or a critic of clean actions for a policy,} \old{and \methodfast requires training a} \old{head once per generator, or once per task for a policy.}

\paragraph{\old{First-order guidance.}} \old{Like the methods we compare with, \method guides with the gradient of the value $V_t$, the leading-order term of the soft value in $\lambda$, $\mathcal V_t=V_t+\frac{\lambda}{2}\Var[r(\vz_0)\mid\vz_t]+O(\lambda^2)$~\cite{stitchvm}, and tunes the guidance strength $\eta$ instead of deriving it from $\lambda$.}

\paragraph{Posterior mismatch.} \Cref{prop:mismatch} bounds the error of the value that a value model regresses, averaged over states of the generator, rather than the error of its gradient. Since \cref{eq:svmerr} scales the difference of the posterior covariances by $(1-s)/s^2$, the gradient error need not decrease monotonically as $s$ decreases. The lookahead state also follows $p_s$ only when $\Phi_{t\to s}$ is the exact flow map, which the sampler step approximates. \old{Where a distilled flow map of the generator is available~\cite{fmrg,diamondmaps}, it can replace the sampler step and remove this bias.}

\paragraph{\old{Real robots.}} \old{All robot experiments run in simulation, on LIBERO and RoboCasa. We will evaluate \method and \methodfast on a real robot in future work.}

\section{\old{Experimental Details}}
\label{app:details}

\subsection{\old{Two-Dimensional Mixtures}}
\label{app:details-toy}

\paragraph{\old{Generator.}} \old{The generator is a mixture of three Gaussians with masses $0.58$, $0.32$, and $0.10$. A flow, a 4-layer MLP of width 256 with inputs $(\vz_t,t,\sin\pi t,\cos\pi t,\log t,\sqrt{1-t})$, is trained on it with the flow-matching loss for 20k steps (batch 2048, Adam, cosine schedule).}

\paragraph{\old{Reward.}} \old{The reward is $r=\kappa\langle\vu,\ve\rangle$ with the embedding $\vu(\vz_0)=\mathrm{normalize}(\tanh(A\vz_0+\vb)+o)\in\mathbb R^8$, $\kappa=5$, and $\lambda=0.65$. The readout $\ve$ is the embedding direction of the rewarded component.}

\paragraph{\old{Value models and their data.}} \old{Value models are 3-layer MLPs (width 128, SiLU) with inputs $(\vz_t,t,\sqrt t,\log t,\text{prompt})$ that output the embedding, guide with $\kappa\nabla_{\vz}\langle V_\psi,\ve\rangle$, and are trained for 8000 steps on forward-noised data \old{in which the rewarded component is rarer and displaced, with mass $0.03$, mean $(1.1,-1.5)$, and covariance scaled by $0.8$. Training is not bitwise reproducible, and \cref{fig:motivation}} \old{uses} \old{the first of three value models trained with different seeds}\old{, except that the StitchVM curves of \cref{fig:motivation}b average all three}\old{.}}

\paragraph{\old{Sampler and methods.}} \old{Guidance acts on $t\le0.9$. Samplers use 100 steps and start from the exact reward-tilted marginal at $t=0.9$, so that only the guidance error is compared\old{. Each} step shifts $\hat\vz_{0\mid t}$ by the guidance and transitions with the Gaussian kernel of the memoryless forward process. AGF is trained for 4 rounds of 2000 trajectories with vector-Jacobian products through $\vv_\theta$ and uses the ODE sampler of its training. DPS backpropagates through the learned denoiser. \method uses $s=\max(0.6t,0.004)$, the value model of StitchVM, and one vector-Jacobian product through $\vv_\theta$. Strengths are swept over $\eta\in\lambda\times\{0.5,0.75,1,1.5,2,3\}$ for the exact value and DPS, over $\lambda\times\{0.5,\dots,6\}$ for StitchVM and \method, and over $\{0.25,\dots,8\}$ for AGF\old{. Extending} the DPS sweep to $16\lambda$ does not lower its MMD$^2$.}

\paragraph{\old{Metrics.}} \old{MMD$^2$~\cite{gretton2012mmd} uses an RBF kernel summed over bandwidths $0.2$, $0.5$, and $1.0$ with 3000 samples per set\old{. The} floor is the MMD$^2$ between two independent target draws, and negative unbiased estimates are printed as $0$. The off-manifold rate is the fraction of samples whose $\log p$ falls below the 1\% quantile of generator samples. Direction diagnostics use 1000 chains of the exact-guided sampler at $\eta\in\{\lambda,2\lambda\}$ with 4 states each.} \old{In \cref{fig:motivation}b, the \method curve compares $\nabla V_\psi(\vz_s,s)$ with the exact $\nabla V_s(\vz_s)$ at the lookahead state $\vz_s=\Phi_{t\to s}(\vz_t)$, $s=0.6t$, of the same states.} In \cref{fig:motivation}b, AGF is represented by the costate it regresses, computed along the deterministic path of the learned flow from $\vz_t$ to $t=0$ with \old{zero guidance}.

\subsection{\old{Text-to-Image Generation}}
\label{app:details-t2i}

\paragraph{\old{Protocol.}} \old{Models, prompts, and metrics are as in \cref{sec:setup}, and all methods share the seeds. \old{The SD3.5 models use 100 steps and a CFG scale of $4.5$, as in StitchVM, and FLUX.1-dev uses 28 steps and $\gamma=1$, as it is guidance-distilled.} Costs in \cref{tab:cost} are measured for one $1024\times1024$ image on one H100 with \old{these} sampling steps. \old{Our implementation compiles the generator transformer with torch.compile in its default mode, runs the generator, the VAE, and the text encoder in bf16 as the official code does, caches the text embeddings and moves the text encoder to the CPU after encoding, and decodes each image once. Specific to our methods, the sampler reuses the velocity computed for the guidance, gradient checkpointing on FLUX.1-dev acts only when backpropagating through the generator, as \method does, and the head of \methodfast runs in bf16, which lowers its memory but not its time}\old{. For \cref{tab:cost}, StitchVM runs in our implementation with the general optimizations above. With its official implementation, it takes 19.0 and 38.0\,s with 28.2 and 44.5\,GB on SD3.5-Medium and SD3.5-Large, and the same on FLUX.1-dev. Pixel-space DPS runs the official StitchVM implementation}, and on FLUX.1-dev \old{its} checkpoints are read on the CPU and freed after loading\old{. Both implementations use the default scaled dot-product attention of diffusers}. \old{AGF is not run on the image generators, as it trains a guidance network for each reward by backpropagating through all sampling steps.} \old{In \cref{fig:motivation}c, \method uses $\gamma=2$ and the bars show one standard deviation.} \old{For \cref{fig:motivation}c, we take 16 states in each of four noise bands from SD3.5-Medium trajectories at $512^2$ on 14 HPDv2 training prompts with the HPSv2 reward\old{, where pixel-space DPS backpropagates HPSv2 through the decoder and the denoiser}, and measure the HPSv2 gain along each unit direction by central differences of size $0.02\|\vz_t\|$, averaged over eight memoryless continuations of 40 steps with shared noise.}}

\paragraph{\old{Guidance strengths.}} \old{\Cref{tab:strengths} lists the strengths used in \cref{tab:main}.} \old{All strengths are chosen separately for each reward by a sweep.} \old{At $512^2$, the HPSv2 of StitchVM stops rising beyond $\eta=400$ while its held-out metrics fall, and \method exceeds it already at its lowest strength (\cref{tab:strength}).} \old{The lookahead ratio and the Jacobian scale are chosen on held-out HPDv2 prompts at $512^2$, and the \method row of \cref{tab:norm} uses $\gamma=2$ at every step.}

\begin{table}[!ht]
\centering\footnotesize
\caption{\old{\textbf{Guidance strength} (SD3.5-Medium, $512^2$, HPSv2 reward, HPDv2 test prompts). Bold: best guided result per column.}}
\label{tab:strength}
\setlength{\tabcolsep}{4pt}
\old{\begin{tabular}{@{}lcccccc@{}}
\toprule
Method & $\eta$ & HPSv2 & Aes. & CLIP & Pick & IR\\
\midrule
Unguided & -- & 0.302 & 5.967 & 29.02 & 22.58 & 1.126\\
\midrule
 & 200 & 0.402 & 6.102 & 28.97 & 23.69 & 1.449\\
 & 400 & 0.417 & 5.939 & 28.64 & 23.56 & 1.408\\
\multirow{-3}{*}{DPS + StitchVM} & 800 & 0.417 & 5.789 & 27.76 & 23.16 & 1.265\\
\midrule
\rowcolor{oursrow} & 150 & 0.428 & 6.127 & 29.40 & \textbf{24.27} & \textbf{1.526}\\
\rowcolor{oursrow} & 200 & 0.434 & \textbf{6.128} & \textbf{29.42} & 24.21 & 1.510\\
\rowcolor{oursrow}\multirow{-3}{*}{\method} & 400 & \textbf{0.437} & 6.070 & 29.06 & 24.01 & 1.438\\
\midrule
\rowcolor{oursrow}\methodfast & 200 & 0.425 & 6.118 & 29.36 & 24.12 & 1.509\\
\bottomrule
\end{tabular}}
\end{table}

\paragraph{\old{Sampling.}} \old{\Cref{alg:t2isample} gives the exact form of \cref{alg:valop} for text-to-image models, with $\gamma(t)=2$ for $t\geq0.8$ and $1$ elsewhere, and $J_\gamma=J_c$ on FLUX.1-dev, which has no CFG.}

\begin{algorithm}[!ht]
\caption{\old{Guided sampling of a text-to-image model}}
\label{alg:t2isample}
\begin{algorithmic}[1]
\Require $\vv_\theta$ with CFG scale $\omega$, value model $V_\psi$, ratio $\rho$, strength $\eta$, Jacobian scale $\gamma(t)$, head $\vh_\phi$ and fit $(a,b)$ (\methodfast)
\State $\vz_1\sim\mathcal N(0,I)$
\For{$t=1,\,1-\Delta t,\dots,\Delta t$}
  \State $\vv\gets\vv_u+\omega(\vv_c-\vv_u)$;\; $s\gets\rho t$;\; $\vz_s\gets\vz_t-(t-s)\,\vv$ \Comment{lookahead}
  \State $\vg_s\gets\nabla_{\vz_s}V_\psi(\vz_s,s)$ \Comment{value backward}
  \If{\method}
    \State $\vxi_t\gets-(t-s)\,J_{\gamma(t)}^{\!\top}\vg_s$,\; $m_t\gets\|\nabla V_\psi(\vz_t,t)\|$ \Comment{VJP, second backward}
  \Else
    \State $\sigma\gets\mathrm{rms}(\vg_s)$,\; $\vxi_t\gets\sigma\,\vh_\phi(\vz_t,t,\vg_s/\sigma)$,\; $m_t\gets e^{a(t)}\|\vg_s\|^{1+b(t)}$ \Comment{head, fit}
  \EndIf
  \State $\vy_t\gets\vg_s+\vxi_t$;\; $\vz_{t-\Delta t}\gets\vz_t-\Delta t\,\vv+\eta\,m_t\,\vy_t/\|\vy_t\|$
\EndFor
\State \Return $\vz_0$
\end{algorithmic}
\end{algorithm}

\begin{table}[t]
\centering\footnotesize
\caption{\old{\textbf{Guidance strengths} in \cref{tab:main} for the HPSv2 (H), aesthetic (A), and CLIP (C) rewards.}}
\label{tab:strengths}
\setlength{\tabcolsep}{2.4pt}
\old{\begin{tabular}{@{}l ccc c ccc c ccc@{}}
\toprule
 & \multicolumn{3}{c}{SD3.5-M} && \multicolumn{3}{c}{SD3.5-L} && \multicolumn{3}{c}{FLUX.1-dev}\\
\cmidrule(lr){2-4}\cmidrule(lr){6-8}\cmidrule(lr){10-12}
Method & H & A & C && H & A & C && H & A & C\\
\midrule
DPS & 100 & 0.3 & 0.6 && 50 & 0.3 & 0.3 && 200 & 0.3 & 0.6\\
StitchVM & 400 & 12 & 75 && 600 & 8 & 45 && 400 & 16 & 100\\
\rowcolor{oursrow}\method & 300 & 6 & 37.5 && 300 & 4 & 22.5 && 800 & 16 & 100\\
\rowcolor{oursrow}\methodfast & 450 & 6 & 37.5 && 450 & 8 & 45 && 800 & 16 & 100\\
\bottomrule
\end{tabular}}
\end{table}

\subsection{Correction Head of \texorpdfstring{\methodfast}{VaLoP-Fast}}
\label{app:head}

\old{The head is trained once per generator for text-to-image models (\cref{alg:head}) and once per task for robot policies (\cref{alg:robothead}).}

\subsubsection{\old{Text-to-Image Models}}
\label{app:head-t2i}

\begin{algorithm}[!ht]
\caption{\old{Training of \methodfast for a text-to-image model}}
\label{alg:head}
\begin{algorithmic}[1]
\Require frozen $\vv_\theta$, HPSv2 value $V_\psi$, training prompts $\mathcal C$, ratio $\rho$, Jacobian scale $\gamma$, head $\vh_\phi$
\Statex \textit{States, at $512^2$ and at $1024^2$}
\For{$c\in\mathcal C$ and eight steps $t\in[0.05,0.95]$ of \old{\cref{alg:t2isample}} with \method}
  \State $s\gets\rho t$,\; $\vz_s\gets\vz_t-(t-s)\,\vv_\theta(\vz_t,t,c)$ \Comment{lookahead}
  \State $\vg^{(0)}\gets\nabla_{\vz_s}V_\psi(\vz_s,s)$,\; $\vg^{(k)}\gets\|\vg^{(0)}\|\,\vu_k/\|\vu_k\|$,\; $k=1,\dots,4$ \Comment{$\vu_k$: random, pulled back}
  \State $\vxi^{(k)}\gets-(t-s)\,J_\gamma^{\!\top}\vg^{(k)}$,\; $k=0,\dots,4$ \Comment{one VJP each}
\EndFor
\Statex \textit{Head, trained at $512^2$ and fine-tuned at $1024^2$}
\Repeat
  \State draw $(\vz_t,t,c,k)$,\; $\sigma\gets\mathrm{rms}(\vg^{(k)})$
  \State $\mathcal L\gets\big\|\vh_\phi(\vz_t,t,\vg^{(k)}/\sigma)-\vxi^{(k)}/\sigma\big\|^2\,/\,\big\|\vxi^{(k)}/\sigma\big\|^2$
  \State $\phi\gets\phi-\alpha\,\nabla_\phi\mathcal L$ \Comment{AdamW}
\Until{the steps of \cref{tab:head}}
\State $\phi\gets\arg\max_{\phi}\cos\big(\vh_\phi,\vxi^{(0)}\big)$ over checkpoints, on held-out states
\Statex \textit{Magnitude, per reward}
\State $a(t),b(t)\gets\arg\min\sum\big(\log\|\nabla V_\psi(\vz_t,t)\|-a(t)-(1+b(t))\log\|\vg_s\|\big)^2$ \Comment{20 bins of $\log t$}
\State \Return $\vh_\phi$,\; $\hat m_t=e^{a(t)}\,\|\vg_s\|^{1+b(t)}$
\end{algorithmic}
\end{algorithm}

\paragraph{Target.} The head predicts the correction in units of the scale of $\vg_s$. With $\mathrm{rms}(\cdot)$ the root mean square over entries, it takes $\hat\vg_s=\vg_s/\mathrm{rms}(\vg_s)$ and regresses $\vxi_t/\mathrm{rms}(\vg_s)$, so that its prediction scales with $\vg_s$ by construction, and \methodfast guides with the direction $\vy_t=\vg_s+\mathrm{rms}(\vg_s)\,\vh_\phi(\vz_t,t,\hat\vg_s)$. On SD3.5, the targets use $\gamma=2$, and on FLUX.1-dev, which has no CFG, $J_\gamma=J_c$.

\paragraph{Architecture.} The head reuses the generator and runs in bf16 (\cref{tab:head}). The frozen generator processes $\vz_t$ through its first four transformer blocks, whose activations are shared with the sampling pass, and the head consists of trainable copies of the next four blocks, the output normalization, and the output projection, initialized from the generator. A patch embedding of $\hat\vg_s$ is added to the image tokens, and a linear map of the timestep embedding at $s$ is added to that at $t$. Both added maps and the output projection are initialized at zero, so the head outputs zero at initialization and \methodfast starts from the uncorrected direction $\vg_s$.

\begin{table}[t]
\centering\footnotesize
\caption{\textbf{Correction head}: parameters, distillation states (strength $\eta$), and training steps (batch).}
\label{tab:head}
\setlength{\tabcolsep}{3pt}
\begin{tabular}{@{}lccc@{}}
\toprule
 & SD3.5-Medium & SD3.5-Large & FLUX.1-dev\\
\midrule
Parameters & 0.41B & 0.87B & 1.39B\\
States at $512^2$ & 20{,}032 (200) & 10{,}240 (200) & 10{,}240 (400)\\
States at $1024^2$ & 7{,}680 (200) & 1{,}280 (300) & \old{2{,}560 (800)}\\
Steps at $512^2$ & 30k (16) & 15k (8) & 20k (8)\\
Steps at $1024^2$ & 8k (8) & 3k (4) & \old{3k (4)}\\
\bottomrule
\end{tabular}
\end{table}

\paragraph{Training data.} States are collected on-policy by running the guided sampler with the HPSv2 value on prompts from the HPDv2 training split, which is disjoint from every evaluation set, and recording eight steps per trajectory with $t\in[0.05,0.95]$. Each state stores $\vz_t$, $t$, the prompt, $\vg_s$, the exact correction, and $\nabla V_\psi(\vz_t,t)$ for the magnitude fit. Since the correction is linear in $\vg_s$, each state also yields pairs for other inputs. We pull four random directions in the output space of the value model's encoder back to $\vz_s$, scale them to the norm of $\vg_s$, and compute their corrections, which gives five input and target pairs per state.

\paragraph{Optimization.} The head minimizes the squared error of \cref{eq:amort} in the normalized units above, divided by the squared norm of the target, with one of the five pairs of a state drawn at random per example. We use AdamW without weight decay, a cosine schedule that decays to $5\%$ of the initial rate, and gradient clipping at norm $1$, training from scratch at $512^2$ with learning rate $10^{-4}$ and fine-tuning at $1024^2$ with $3\times10^{-5}$ (\cref{tab:head}). We hold out $5\%$ of the data and keep the checkpoint with the highest cosine between the predicted and the exact direction on held-out states, evaluated every 1000 steps. The $512^2$ head already aligns with the $1024^2$ correction (cosine $0.85$). \old{On one GPU, the head trains for 3.2, 4.2, and 6.1\,h at $512^2$ and is fine-tuned for 1.0, 0.25, and 1.8\,h at $1024^2$ on SD3.5-Medium, SD3.5-Large, and FLUX.1-dev, using an A6000, an H100, and an A6000.}

\paragraph{\old{Magnitude fit.}} \old{\method computes $m_t$ with a second value-model backward pass, and \methodfast replaces it with $\hat m_t$.} \old{The coefficients $a$ and $b$ of $\hat m_t$ (\cref{sec:practical}) are fitted by least squares in 20 bins of $t$ in log space on exact pairs $(\vg_s,\nabla V_\psi(\vz_t,t))$ recorded on guided states, separately for each reward and resolution. Gradient norms differ between value models in scale and in how they change with the noise level, and the pairs come from states guided by the same reward\old{. A} fit on states guided by another reward mispredicts the magnitude. Fitting takes about half a day per reward. The fitted $b$ is negative for all rewards and noise levels, so a larger lookahead gradient comes with a relatively smaller gradient at $\vz_t$. On held-out states at $512^2$, the fit explains $54\%$ of the variance of the log magnitude ($R^2$), against $31\%$ with $t$ alone, and a small network on head features does no better.}

\paragraph{Cost.} Per step, \methodfast performs one generator forward pass, shared with the sampler, one value-model backward pass at $\vz_s$, and one forward pass of the head, without a generator backward pass. \old{Against \old{StitchVM in the same implementation}, it takes $13$ to \old{$15\%$} less time on the three generators (\cref{tab:cost}).}

\subsubsection{\old{Robot Policies}}
\label{app:head-robot}

\old{\rev{In place of the squared error of \cref{eq:amort-robot},} the head predicts the unit direction and the log-norm of the guidance of \method, \rev{whose norm is matched to that of the QGF gradient $\vg^{\mathrm{QGF}}_t=\nabla_{\vz_t}Q\big(o,\vz_t-t\,\sg[\vv_\theta(\vz_t,t)]\big)$, the critic gradient at the denoised action with the policy Jacobian replaced by the identity~\cite{qgf} (\cref{sec:practical}),} so that, unlike for images, \methodfast needs no critic at inference (\cref{alg:robothead}). It is the FiLM MLP of AGF\old{, with three hidden layers of width 128 and LayerNorm,} on the frozen image encoder of the critic, with 0.75M (SmolVLA) and 0.90M ($\pi_{0.5}$) parameters, and takes the noisy action chunk and \old{a 64-dimensional sinusoidal embedding of} $t$. It is trained on 100 unguided episodes per task (300 on Goal), with states rolled out under the head being trained\old{ and taken at the 11 times $t\in\{1,0.9,\dots,0\}$}, using AdamW at \old{a constant} learning rate \old{of} $3\times10^{-4}$\old{ with weight decay $10^{-4}$, gradient clipping at norm 1, and batch size 128 (32 on RoboCasa)}, which takes 1.6 to 2.9\,h \old{per task} on one GPU (\cref{tab:robotcost}).}

\begin{algorithm}[!ht]
\caption{\old{Training of \methodfast for a robot policy}}
\label{alg:robothead}
\begin{algorithmic}[1]
\Require frozen $\vv_\theta$, critic $Q$, unguided observations $\mathcal O$, ratio $\rho$, head $\vh_\phi=(\hat\vu_\phi,\hat\ell_\phi)$
\Repeat
  \State $o\sim\mathcal O$,\; $\vz_1\sim\mathcal N(0,I)$
  \Statex \textit{Rollout with the current head}
  \For{$t=1,0.9,\dots,0.1$}
    \State $\vv\gets\vv_\theta(\vz_t,t,o)$;\; $\vz_{t-\Delta t}\gets\vz_t-\Delta t\,\vv+\Delta t\,\min\big(e^{\hat\ell_\phi},\|\vv\|\big)\,\hat\vu_\phi/\|\hat\vu_\phi\|$ \Comment{clipped at $\|\vv\|$}
  \EndFor
  \Statex \textit{Targets}
  \For{$t=1,0.9,\dots,0$, with $s=\rho t$}
    \State $\vz_s\gets\vz_t-(t-s)\,\vv_\theta(\vz_t,t)$ \Comment{lookahead}
    \State $\vg_s\gets\nabla_{\vz_s}Q\big(o,\vz_s-s\,\vv_\theta(\vz_s,s)\big)$ \Comment{critic at the denoised action}
    \State $\vy_t\gets\vg_s-(t-s)\,J^{\!\top}\vg_s$ \Comment{one VJP}
    \State $m_t\gets\big\|\nabla_{\vz_t}Q\big(o,\vz_t-t\,\mathrm{sg}[\vv_\theta(\vz_t,t)]\big)\big\|$ \Comment{QGF norm}
  \EndFor
  \State $\mathcal L\gets\sum_t\big\|\hat\vu_\phi(o,\vz_t,t)-\vy_t/\|\vy_t\|\big\|^2+\big(\hat\ell_\phi(o,\vz_t,t)-\log m_t\big)^2$
  \State $\phi\gets\phi-\alpha\,\nabla_\phi\mathcal L$ \Comment{AdamW}
\Until{the step budget}
\Statex \textit{Inference:} \cref{alg:robotsample}, without $Q$
\end{algorithmic}
\end{algorithm}

\subsection{\rev{Robot Policies}}
\label{app:details-robot}

\paragraph{\old{Policies and critics.}} \old{We use the public SmolVLA checkpoint trained on LIBERO and the $\pi_{0.5}$ checkpoint trained on RoboCasa, both frozen,} \old{with the action chunking of AGF~\cite{agf}: chunks of 50 actions, of which SmolVLA executes 10 and $\pi_{0.5}$ all 50, and the critic scores the executed part. Each task has a critic, an ensemble of ten Q-networks on clean action chunks, reduced to the mean minus half the standard deviation and trained by temporal differences on unguided rollouts of the base policy. On LIBERO, we use the critics of AGF, and on RoboCasa, we train them with the recipe of AGF on about 640 unguided episodes per task. \old{Each Q-network takes the 64-pixel camera images, with the robot state on RoboCasa, and the action chunk, and has three hidden layers of width 128. It is trained for 10k steps with Adam at learning rate $3\times10^{-4}$, with 500 warm-up steps and cosine decay, batch size 256, gradient clipping at norm 1, discount $0.99$, and a Polyak-averaged target network with rate $0.005$, which guidance uses.} AGF uses its \old{official} guidance heads on RoboCasa.}

\paragraph{\old{Baselines.}} \old{Q-BoN samples $N$ action chunks from the unguided policy and executes the one with the highest critic value, as in V-GPS~\cite{vgps}. QDPS evaluates the critic at the denoised action and backpropagates through the policy, as DPS does (QGF-Jacobian in~\cite{qgf}), whereas QGF~\cite{qgf} replaces the Jacobian of the denoised action with the identity and backpropagates only through the critic. AGF regresses the costate onto a guidance network trained per task. We omit QAM~\cite{qam}, which fine-tunes the policy.}

\paragraph{\old{\method and \methodfast.}} \old{\method reads the critic at the denoised lookahead action with $s=0.6t$, pulls its gradient back with one VJP, and rescales the result to the norm of the QGF gradient at $\vz_t$. \methodfast amortizes this guidance with the head of \cref{app:head-robot}, and \cref{alg:robotsample} gives both samplers.}

\begin{algorithm}[!ht]
\caption{\old{Guided sampling of a robot policy}}
\label{alg:robotsample}
\begin{algorithmic}[1]
\Require $\vv_\theta$, critic $Q$, observation $o$, ratio $\rho$, strength $\eta=w\,\Delta t$, head $\vh_\phi=(\hat\vu_\phi,\hat\ell_\phi)$ (\methodfast)
\State $\vz_1\sim\mathcal N(0,I)$
\For{$t=1,0.9,\dots,0.1$}
  \State $\vv\gets\vv_\theta(\vz_t,t,o)$;\; $s\gets\rho t$
  \If{\method}
    \State $\vz_s\gets\vz_t-(t-s)\,\vv$ \Comment{lookahead}
    \State $\vg_s\gets\nabla_{\vz_s}Q\big(o,\vz_s-s\,\vv_\theta(\vz_s,s)\big)$ \Comment{critic at the denoised action}
    \State $\vy_t\gets\vg_s-(t-s)\,J^{\!\top}\vg_s$ \Comment{one VJP}
    \State $m_t\gets\big\|\nabla_{\vz_t}Q\big(o,\vz_t-t\,\mathrm{sg}[\vv]\big)\big\|$ \Comment{QGF norm}
  \Else
    \State $\vy_t\gets\hat\vu_\phi(o,\vz_t,t)$,\; $m_t\gets e^{\hat\ell_\phi(o,\vz_t,t)}$ \Comment{head, no critic}
  \EndIf
  \State $\vz_{t-\Delta t}\gets\vz_t-\Delta t\,\vv+\eta\,\min(m_t,1)\,\vy_t/\|\vy_t\|$ \Comment{\cref{eq:update}, clipped at norm 1}
\EndFor
\State \Return $\vz_0$
\end{algorithmic}
\end{algorithm}

\paragraph{\old{Protocol.}} \old{Gradient guidance is clipped at norm $1$, which rarely binds in practice, and the guidance of AGF at the norm of the policy velocity. For every method, one strength per suite is selected by the mean success rate of the suite, from $w\in\{0.25,0.5,1,2,4\}$ on LIBERO and $w\in\{0.5,1,2,4\}$ on RoboCasa (\cref{tab:libero_sweep,tab:robocasa_sweep}). We report 50 episodes per task, with time limits of 300, 220, and 520 steps on Goal, Spatial, and LIBERO-10 and the task horizons of RoboCasa, 450 to 1050 steps.} \old{\old{Shared episode seeds fix the initial action noise.}}

\begin{table}[!ht]
\centering\footnotesize
\caption{\old{\textbf{LIBERO over the strength grid.} Success rate (\%) per suite, with the base policy at 76.4, 64.4, and 43.4. Bold: highest, reported in \cref{tab:robot}.}}
\label{tab:libero_sweep}
\setlength{\tabcolsep}{3.2pt}
\old{\begin{tabular}{@{}l ccccc c ccccc c ccccc@{}}
\toprule
 & \multicolumn{5}{c}{Goal} && \multicolumn{5}{c}{Spatial} && \multicolumn{5}{c}{LIBERO-10}\\
\cmidrule(lr){2-6}\cmidrule(lr){8-12}\cmidrule(lr){14-18}
$w$ & 0.25 & 0.5 & 1 & 2 & 4 && 0.25 & 0.5 & 1 & 2 & 4 && 0.25 & 0.5 & 1 & 2 & 4\\
\midrule
QGF & 76.4 & 76.2 & 78.6 & \textbf{79.0} & 78.6 && 67.6 & 65.0 & 68.2 & 68.8 & \textbf{69.4} && 46.4 & 44.4 & 44.2 & 45.8 & \textbf{48.6}\\
QDPS & 77.8 & 79.4 & 78.8 & 74.8 & \textbf{80.2} && 65.8 & 65.0 & 66.6 & \textbf{68.2} & 68.0 && 43.6 & 46.0 & 41.6 & \textbf{47.6} & 47.4\\
AGF & 77.4 & \textbf{80.0} & 77.6 & 77.6 & \textbf{80.0} && 64.6 & 64.8 & 65.6 & \textbf{69.2} & 68.0 && 46.0 & 46.0 & \textbf{48.0} & 44.8 & 45.2\\
\rowcolor{oursrow}\method & 79.2 & 78.0 & 80.4 & 79.0 & \textbf{80.6} && 66.6 & 65.0 & 65.8 & \textbf{69.4} & 68.2 && 45.4 & 47.4 & 45.2 & \textbf{49.0} & 45.2\\
\rowcolor{oursrow}\methodfast & 78.0 & 77.0 & 79.0 & 76.0 & \textbf{81.8} && 67.6 & 67.8 & 66.6 & 68.6 & \textbf{69.8} && 44.8 & \textbf{49.4} & 44.8 & 44.4 & 45.8\\
\bottomrule
\end{tabular}}
\end{table}

\begin{table}[!ht]
\centering\footnotesize
\caption{\old{\textbf{RoboCasa over the strength grid.} Success rate (\%) over the 18 atomic tasks, with the base policy at 44.6. Bold: highest, reported in \cref{tab:robot}.}}
\label{tab:robocasa_sweep}
\old{\begin{tabular}{@{}lcccc@{}}
\toprule
$w$ & 0.5 & 1 & 2 & 4\\
\midrule
QGF & 44.8 & 45.3 & 44.4 & \textbf{47.1}\\
QDPS & 44.0 & 45.0 & 45.2 & \textbf{45.7}\\
AGF & 44.3 & 44.9 & \textbf{45.9} & 42.7\\
\rowcolor{oursrow}\method & 44.4 & 43.4 & 46.3 & \textbf{47.4}\\
\rowcolor{oursrow}\methodfast & 44.7 & 46.8 & 45.8 & \textbf{47.4}\\
\bottomrule
\end{tabular}}
\end{table}

\paragraph{\old{Cost.}} \old{\Cref{tab:robotcost} lists the costs on both policies, measured on one H100 at batch size one for one action chunk of ten flow steps, as medians over 30 stored observations after three warm-up runs. Memory is the peak GPU memory, including the policy of 605M (SmolVLA) or 4.14B ($\pi_{0.5}$) parameters, with its increase over Base in parentheses. \methodfast matches the inference time of AGF with a head of similar size on both policies, and on RoboCasa it adds almost no memory, since its head also predicts the magnitude and the critic is not loaded at inference. On SmolVLA, its head trains in less than half the time of AGF per task, at a higher peak memory, since its targets differentiate the policy once more. \method needs no training.}

\begin{table}[!ht]
\centering\footnotesize
\caption{\old{\textbf{Cost of critic guidance.} Inference per action chunk, with the ratio of the time to Base and the added memory in parentheses, and parameters of the guidance module\old{. Training is} per task. The critic, trained once per task and shared by all methods (about 1\,h on SmolVLA and 1.5\,h on $\pi_{0.5}$), is not counted.}}
\label{tab:robotcost}
\setlength{\tabcolsep}{3.5pt}
\old{\begin{tabular}{@{}l cc c cc c c c ccc c cc@{}}
\toprule
 & \multicolumn{5}{c}{Inference, time (ms) and memory (MB)} && && \multicolumn{6}{c}{Training per task}\\
\cmidrule(lr){2-6}\cmidrule(lr){10-15}
 & \multicolumn{2}{c}{SmolVLA} && \multicolumn{2}{c}{$\pi_{0.5}$} && Param. && \multicolumn{3}{c}{SmolVLA} && \multicolumn{2}{c}{$\pi_{0.5}$}\\
\cmidrule(lr){2-3}\cmidrule(lr){5-6}\cmidrule(lr){10-12}\cmidrule(lr){14-15}
Method & Time\old{$\downarrow$} & Mem.\old{$\downarrow$} && Time\old{$\downarrow$} & Mem.\old{$\downarrow$} && && s/step\old{$\downarrow$} & h\old{$\downarrow$} & GB\old{$\downarrow$} && h\old{$\downarrow$} & GB\old{$\downarrow$}\\
\midrule
Base & 448 & 1,338 && 228 & 9,169 && -- && -- & -- & -- && -- & --\\
Q-BoN ($N=8$) & 463 (1.03) & 1,435 (+97) && 250 (1.09) & 9,638 (+469) && \old{5.3M / 6.0M} && -- & -- & -- && -- & --\\
QGF & 587 (1.31) & 1,370 (+32) && 303 (1.33) & 9,249 (+80) && \old{5.3M / 6.0M} && -- & -- & -- && -- & --\\
QDPS & 1,045 (2.33) & 1,425 (+87) && 633 (2.77) & 9,249 (+80) && \old{5.3M / 6.0M} && -- & -- & -- && -- & --\\
AGF & 460 (1.03) & 1,370 (+32) && 240 (1.05) & 9,219 (+49) && 0.75M / 0.90M && 5.06 & 7.0 & 15.1 && \multicolumn{2}{c}{\old{official checkpoint}}\\
\rowcolor{oursrow}\textbf{\method} (Ours) & 2,048 (4.57) & 1,498 (+160) && 1,139 (4.99) & 9,319 (+150) && \old{5.3M / 6.0M} && -- & -- & -- && -- & --\\
\rowcolor{oursrow}\textbf{\methodfast} (Ours) & 460 (1.03) & 1,370 (+32) && 240 (1.05) & 9,173 (+4) && 0.75M / 0.90M && 3.01 & 2.1--2.9 & 24.6 && 1.6 & 26.4\\
\bottomrule
\end{tabular}}
\end{table}

\FloatBarrier
\clearpage
\section{\old{Additional Results}}
\label{app:extra}

\subsection{\old{Statistical Analysis}}
\label{app:stats}

\paragraph{\old{Seeds.}} \old{\Cref{tab:seeds} lists the standard deviations of \cref{tab:main}. On DrawBench, the target reward of either variant exceeds that of StitchVM and pixel-space DPS by more than one standard deviation in every setting.}

\begin{table}[H]
\centering\footnotesize
\caption{\old{\textbf{Means and standard deviations} of \cref{tab:main} over three DrawBench seeds and four GenEval seeds (three for DPS).}}
\label{tab:seeds}
\setlength{\tabcolsep}{4pt}
\old{\begin{tabular}{@{}l cccccc@{}}
\toprule
Method & HPSv2\old{$\uparrow$} & Aes.\old{$\uparrow$} & CLIP\old{$\uparrow$} & Pick\old{$\uparrow$} & IR\old{$\uparrow$} & GenE.\old{$\uparrow$}\\
\midrule
\multicolumn{7}{@{}l}{\textbf{SD3.5-Medium}}\\
Unguided & 0.286{\scriptsize$\pm$0.001} & 5.347{\scriptsize$\pm$0.024} & 28.80{\scriptsize$\pm$0.09} & 22.65{\scriptsize$\pm$0.06} & 0.987{\scriptsize$\pm$0.032} & 0.694{\scriptsize$\pm$0.017}\\
\multicolumn{7}{@{}l}{\textit{HPSv2 reward}}\\
DPS & 0.361{\scriptsize$\pm$0.000} & 5.426{\scriptsize$\pm$0.024} & 28.99{\scriptsize$\pm$0.18} & 23.54{\scriptsize$\pm$0.04} & 1.130{\scriptsize$\pm$0.028} & 0.711{\scriptsize$\pm$0.010}\\
DPS + StitchVM & 0.366{\scriptsize$\pm$0.001} & 5.541{\scriptsize$\pm$0.002} & 28.77{\scriptsize$\pm$0.16} & 23.28{\scriptsize$\pm$0.03} & 1.302{\scriptsize$\pm$0.048} & 0.647{\scriptsize$\pm$0.012}\\
\rowcolor{oursrow}\textbf{\method} (Ours) & 0.382{\scriptsize$\pm$0.001} & 5.654{\scriptsize$\pm$0.014} & \textbf{29.13}{\scriptsize$\pm$0.07} & \textbf{23.77}{\scriptsize$\pm$0.03} & 1.332{\scriptsize$\pm$0.015} & \textbf{0.726}{\scriptsize$\pm$0.011}\\
\rowcolor{oursrow}\textbf{\methodfast} (Ours) & \textbf{0.387}{\scriptsize$\pm$0.000} & \textbf{5.671}{\scriptsize$\pm$0.023} & 28.92{\scriptsize$\pm$0.06} & 23.72{\scriptsize$\pm$0.01} & \textbf{1.333}{\scriptsize$\pm$0.024} & 0.702{\scriptsize$\pm$0.017}\\
\multicolumn{7}{@{}l}{\textit{Aesthetic reward}}\\
DPS & 0.285{\scriptsize$\pm$0.001} & 5.877{\scriptsize$\pm$0.008} & \textbf{28.30}{\scriptsize$\pm$0.09} & 22.63{\scriptsize$\pm$0.05} & 0.944{\scriptsize$\pm$0.047} & 0.662{\scriptsize$\pm$0.012}\\
DPS + StitchVM & 0.279{\scriptsize$\pm$0.001} & 5.968{\scriptsize$\pm$0.031} & 26.95{\scriptsize$\pm$0.28} & 22.29{\scriptsize$\pm$0.03} & 0.801{\scriptsize$\pm$0.007} & 0.644{\scriptsize$\pm$0.022}\\
\rowcolor{oursrow}\textbf{\method} (Ours) & 0.288{\scriptsize$\pm$0.001} & \textbf{6.174}{\scriptsize$\pm$0.010} & 27.22{\scriptsize$\pm$0.27} & 22.56{\scriptsize$\pm$0.01} & 0.905{\scriptsize$\pm$0.007} & \textbf{0.672}{\scriptsize$\pm$0.016}\\
\rowcolor{oursrow}\textbf{\methodfast} (Ours) & \textbf{0.293}{\scriptsize$\pm$0.001} & 6.035{\scriptsize$\pm$0.033} & 27.62{\scriptsize$\pm$0.17} & \textbf{22.69}{\scriptsize$\pm$0.01} & \textbf{0.974}{\scriptsize$\pm$0.017} & \textbf{0.672}{\scriptsize$\pm$0.007}\\
\multicolumn{7}{@{}l}{\textit{CLIP reward}}\\
DPS & 0.254{\scriptsize$\pm$0.003} & 4.955{\scriptsize$\pm$0.034} & 34.27{\scriptsize$\pm$0.07} & 21.70{\scriptsize$\pm$0.06} & 0.622{\scriptsize$\pm$0.072} & 0.633{\scriptsize$\pm$0.023}\\
DPS + StitchVM & 0.248{\scriptsize$\pm$0.002} & 4.877{\scriptsize$\pm$0.032} & 34.42{\scriptsize$\pm$0.09} & 21.66{\scriptsize$\pm$0.07} & 0.688{\scriptsize$\pm$0.065} & 0.625{\scriptsize$\pm$0.006}\\
\rowcolor{oursrow}\textbf{\method} (Ours) & 0.266{\scriptsize$\pm$0.001} & 4.999{\scriptsize$\pm$0.036} & \textbf{35.98}{\scriptsize$\pm$0.15} & 22.08{\scriptsize$\pm$0.06} & 0.942{\scriptsize$\pm$0.006} & \textbf{0.725}{\scriptsize$\pm$0.006}\\
\rowcolor{oursrow}\textbf{\methodfast} (Ours) & \textbf{0.275}{\scriptsize$\pm$0.001} & \textbf{5.085}{\scriptsize$\pm$0.012} & 35.03{\scriptsize$\pm$0.13} & \textbf{22.26}{\scriptsize$\pm$0.02} & \textbf{1.022}{\scriptsize$\pm$0.031} & 0.722{\scriptsize$\pm$0.010}\\
\midrule
\multicolumn{7}{@{}l}{\textbf{SD3.5-Large}}\\
Unguided & 0.294{\scriptsize$\pm$0.001} & 5.461{\scriptsize$\pm$0.029} & 28.61{\scriptsize$\pm$0.15} & 22.87{\scriptsize$\pm$0.04} & 1.065{\scriptsize$\pm$0.019} & 0.721{\scriptsize$\pm$0.016}\\
\multicolumn{7}{@{}l}{\textit{HPSv2 reward}}\\
DPS & 0.349{\scriptsize$\pm$0.000} & 5.519{\scriptsize$\pm$0.026} & 28.82{\scriptsize$\pm$0.16} & 23.55{\scriptsize$\pm$0.05} & 1.111{\scriptsize$\pm$0.034} & 0.715{\scriptsize$\pm$0.017}\\
DPS + StitchVM & 0.360{\scriptsize$\pm$0.001} & 5.544{\scriptsize$\pm$0.010} & 28.65{\scriptsize$\pm$0.05} & 23.20{\scriptsize$\pm$0.03} & 1.248{\scriptsize$\pm$0.019} & 0.662{\scriptsize$\pm$0.014}\\
\rowcolor{oursrow}\textbf{\method} (Ours) & \textbf{0.371}{\scriptsize$\pm$0.004} & 5.633{\scriptsize$\pm$0.019} & 29.24{\scriptsize$\pm$0.09} & \textbf{23.78}{\scriptsize$\pm$0.08} & 1.290{\scriptsize$\pm$0.028} & \textbf{0.755}{\scriptsize$\pm$0.028}\\
\rowcolor{oursrow}\textbf{\methodfast} (Ours) & 0.368{\scriptsize$\pm$0.001} & \textbf{5.657}{\scriptsize$\pm$0.021} & \textbf{29.40}{\scriptsize$\pm$0.01} & 23.66{\scriptsize$\pm$0.03} & \textbf{1.323}{\scriptsize$\pm$0.019} & 0.722{\scriptsize$\pm$0.002}\\
\multicolumn{7}{@{}l}{\textit{Aesthetic reward}}\\
DPS & 0.295{\scriptsize$\pm$0.001} & 5.835{\scriptsize$\pm$0.017} & \textbf{28.51}{\scriptsize$\pm$0.04} & 22.87{\scriptsize$\pm$0.02} & \textbf{1.067}{\scriptsize$\pm$0.009} & 0.708{\scriptsize$\pm$0.006}\\
DPS + StitchVM & 0.287{\scriptsize$\pm$0.001} & 5.807{\scriptsize$\pm$0.024} & 28.05{\scriptsize$\pm$0.01} & 22.64{\scriptsize$\pm$0.02} & 0.992{\scriptsize$\pm$0.022} & 0.691{\scriptsize$\pm$0.008}\\
\rowcolor{oursrow}\textbf{\method} (Ours) & \textbf{0.297}{\scriptsize$\pm$0.002} & 5.989{\scriptsize$\pm$0.019} & 28.15{\scriptsize$\pm$0.08} & \textbf{22.88}{\scriptsize$\pm$0.02} & 1.058{\scriptsize$\pm$0.020} & \textbf{0.712}{\scriptsize$\pm$0.016}\\
\rowcolor{oursrow}\textbf{\methodfast} (Ours) & 0.296{\scriptsize$\pm$0.000} & \textbf{6.144}{\scriptsize$\pm$0.027} & 28.06{\scriptsize$\pm$0.18} & 22.80{\scriptsize$\pm$0.03} & 1.066{\scriptsize$\pm$0.017} & 0.706{\scriptsize$\pm$0.011}\\
\multicolumn{7}{@{}l}{\textit{CLIP reward}}\\
DPS & \textbf{0.286}{\scriptsize$\pm$0.001} & \textbf{5.321}{\scriptsize$\pm$0.022} & 32.13{\scriptsize$\pm$0.03} & \textbf{22.60}{\scriptsize$\pm$0.01} & 0.982{\scriptsize$\pm$0.061} & 0.713{\scriptsize$\pm$0.013}\\
DPS + StitchVM & 0.275{\scriptsize$\pm$0.001} & 5.183{\scriptsize$\pm$0.023} & 32.43{\scriptsize$\pm$0.08} & 22.40{\scriptsize$\pm$0.03} & 0.983{\scriptsize$\pm$0.030} & 0.724{\scriptsize$\pm$0.007}\\
\rowcolor{oursrow}\textbf{\method} (Ours) & 0.284{\scriptsize$\pm$0.002} & 5.281{\scriptsize$\pm$0.032} & \textbf{33.69}{\scriptsize$\pm$0.06} & \textbf{22.60}{\scriptsize$\pm$0.02} & 1.087{\scriptsize$\pm$0.022} & \textbf{0.759}{\scriptsize$\pm$0.012}\\
\rowcolor{oursrow}\textbf{\methodfast} (Ours) & 0.282{\scriptsize$\pm$0.000} & 5.222{\scriptsize$\pm$0.007} & 33.65{\scriptsize$\pm$0.10} & 22.44{\scriptsize$\pm$0.04} & \textbf{1.095}{\scriptsize$\pm$0.007} & 0.735{\scriptsize$\pm$0.015}\\
\midrule
\multicolumn{7}{@{}l}{\textbf{FLUX.1-dev}}\\
Unguided & 0.298{\scriptsize$\pm$0.001} & 5.729{\scriptsize$\pm$0.029} & 27.71{\scriptsize$\pm$0.26} & 22.79{\scriptsize$\pm$0.04} & 0.995{\scriptsize$\pm$0.064} & 0.660{\scriptsize$\pm$0.006}\\
\multicolumn{7}{@{}l}{\textit{HPSv2 reward}}\\
DPS & 0.320{\scriptsize$\pm$0.005} & 5.440{\scriptsize$\pm$0.033} & 27.09{\scriptsize$\pm$0.41} & 22.78{\scriptsize$\pm$0.16} & 0.922{\scriptsize$\pm$0.127} & 0.633{\scriptsize$\pm$0.013}\\
DPS + StitchVM & 0.314{\scriptsize$\pm$0.001} & 5.690{\scriptsize$\pm$0.006} & 27.82{\scriptsize$\pm$0.09} & 22.79{\scriptsize$\pm$0.06} & 1.080{\scriptsize$\pm$0.020} & 0.652{\scriptsize$\pm$0.009}\\
\rowcolor{oursrow}\textbf{\method} (Ours) & 0.328{\scriptsize$\pm$0.001} & 5.590{\scriptsize$\pm$0.023} & \textbf{28.36}{\scriptsize$\pm$0.31} & \textbf{23.12}{\scriptsize$\pm$0.02} & 1.190{\scriptsize$\pm$0.030} & \textbf{0.701}{\scriptsize$\pm$0.011}\\
\rowcolor{oursrow}\textbf{\methodfast} (Ours) & \textbf{0.333}{\scriptsize$\pm$0.001} & \textbf{5.782}{\scriptsize$\pm$0.028} & 28.28{\scriptsize$\pm$0.09} & 23.11{\scriptsize$\pm$0.02} & \textbf{1.198}{\scriptsize$\pm$0.039} & 0.686{\scriptsize$\pm$0.008}\\
\multicolumn{7}{@{}l}{\textit{Aesthetic reward}}\\
DPS & 0.297{\scriptsize$\pm$0.002} & 5.815{\scriptsize$\pm$0.026} & \textbf{27.61}{\scriptsize$\pm$0.20} & 22.78{\scriptsize$\pm$0.06} & 0.991{\scriptsize$\pm$0.046} & 0.641{\scriptsize$\pm$0.014}\\
DPS + StitchVM & 0.297{\scriptsize$\pm$0.001} & 5.893{\scriptsize$\pm$0.019} & 27.37{\scriptsize$\pm$0.23} & 22.74{\scriptsize$\pm$0.07} & 0.967{\scriptsize$\pm$0.077} & 0.651{\scriptsize$\pm$0.012}\\
\rowcolor{oursrow}\textbf{\method} (Ours) & 0.300{\scriptsize$\pm$0.001} & \textbf{5.996}{\scriptsize$\pm$0.010} & 27.29{\scriptsize$\pm$0.10} & 22.84{\scriptsize$\pm$0.04} & 0.968{\scriptsize$\pm$0.052} & \textbf{0.666}{\scriptsize$\pm$0.007}\\
\rowcolor{oursrow}\textbf{\methodfast} (Ours) & \textbf{0.301}{\scriptsize$\pm$0.001} & 5.929{\scriptsize$\pm$0.016} & 27.46{\scriptsize$\pm$0.21} & \textbf{22.85}{\scriptsize$\pm$0.07} & \textbf{1.007}{\scriptsize$\pm$0.059} & 0.659{\scriptsize$\pm$0.016}\\
\multicolumn{7}{@{}l}{\textit{CLIP reward}}\\
DPS & 0.279{\scriptsize$\pm$0.006} & 5.410{\scriptsize$\pm$0.069} & 29.28{\scriptsize$\pm$0.23} & 22.25{\scriptsize$\pm$0.12} & 0.819{\scriptsize$\pm$0.130} & 0.637{\scriptsize$\pm$0.015}\\
DPS + StitchVM & 0.286{\scriptsize$\pm$0.003} & 5.499{\scriptsize$\pm$0.029} & 29.55{\scriptsize$\pm$0.09} & 22.46{\scriptsize$\pm$0.08} & 0.936{\scriptsize$\pm$0.038} & 0.648{\scriptsize$\pm$0.025}\\
\rowcolor{oursrow}\textbf{\method} (Ours) & 0.291{\scriptsize$\pm$0.001} & 5.514{\scriptsize$\pm$0.027} & \textbf{30.79}{\scriptsize$\pm$0.02} & 22.65{\scriptsize$\pm$0.05} & \textbf{1.097}{\scriptsize$\pm$0.027} & 0.691{\scriptsize$\pm$0.024}\\
\rowcolor{oursrow}\textbf{\methodfast} (Ours) & \textbf{0.295}{\scriptsize$\pm$0.001} & \textbf{5.545}{\scriptsize$\pm$0.039} & 30.37{\scriptsize$\pm$0.18} & \textbf{22.67}{\scriptsize$\pm$0.02} & 1.082{\scriptsize$\pm$0.044} & \textbf{0.695}{\scriptsize$\pm$0.012}\\
\bottomrule
\end{tabular}}
\end{table}

\paragraph{\old{Paired $t$-test.}} \old{\Cref{tab:ablation_full} lists every setting of the ablation in \cref{sec:ablation} with its paired $t$-statistic against \method, and its last rows come from separate runs with the same seeds.}

\begin{table}[H]
\centering\footnotesize
\caption{\old{\textbf{Ablation at $1024^2$} (SD3.5-Medium, DrawBench, HPSv2 reward, $\eta=300$, \old{one seed}). Each row changes one setting of \method\old{, and parentheses give the} paired $t$-statistic against \method over the 200 images.} \old{Bold: $|t|\geq2$ on the target reward.}}
\label{tab:ablation_full}
\setlength{\tabcolsep}{4pt}
\old{\begin{tabular}{@{}llccccc@{}}
\toprule
Setting & Value & HPSv2\old{$\uparrow$} & Aes.\old{$\uparrow$} & CLIP\old{$\uparrow$} & Pick\old{$\uparrow$} & IR\old{$\uparrow$}\\
\midrule
Unguided & & 0.287 & 5.370 & 28.71 & 22.69 & 0.995\\
\rowcolor{oursrow}\method & $s/t=0.6$, $\beta=1$, $\gamma$ schedule & 0.382 & 5.665 & 29.06 & 23.79 & 1.339\\
\midrule
Ratio $s/t$ & $1$ (StitchVM) & \old{\textbf{0.358} ($\mathbf{-13.1}$)} & 5.505 ($-4.8$) & 28.63 ($-1.9$) & 23.28 ($-9.6$) & 1.322 ($-0.5$)\\
 & $0.8$ & \old{\textbf{0.385} ($\mathbf{+2.1}$)} & 5.623 ($-1.7$) & 29.36 ($+1.6$) & 23.79 ($-0.1$) & 1.386 ($+1.8$)\\
 & $0.4$ & \old{\textbf{0.369} ($\mathbf{-7.8}$)} & 5.599 ($-2.3$) & 29.23 ($+0.8$) & 23.60 ($-4.1$) & 1.267 ($-2.2$)\\
 & $0.2$ & \old{\textbf{0.359} ($\mathbf{-11.9}$)} & 5.535 ($-3.7$) & 29.27 ($+0.8$) & 23.41 ($-6.4$) & 1.217 ($-2.9$)\\
 & $0$ (DPS form) & \old{\textbf{0.352} ($\mathbf{-15.3}$)} & 5.574 ($-2.4$) & 28.84 ($-0.9$) & 23.30 ($-8.8$) & 1.193 ($-4.4$)\\
\midrule
Correction & none ($\vg_s$ only) & \old{\textbf{0.365} ($\mathbf{-9.3}$)} & 5.614 ($-1.7$) & 28.23 ($-3.4$) & 23.52 ($-5.8$) & 1.324 ($-0.5$)\\
$S-\beta A$ & $\beta=-1$ ($J$) & \old{\textbf{0.368} ($\mathbf{-10.0}$)} & 5.598 ($-2.2$) & 28.25 ($-3.8$) & 23.59 ($-4.8$) & 1.318 ($-0.9$)\\
 & $\beta=0$ ($S$) & \old{\textbf{0.379} ($\mathbf{-2.0}$)} & 5.625 ($-1.6$) & 29.42 ($+1.8$) & 23.76 ($-0.7$) & 1.334 ($-0.2$)\\
 & $\beta=0.5$ & 0.381 ($-0.3$) & 5.636 ($-1.1$) & 29.34 ($+1.4$) & 23.80 ($0.0$) & 1.346 ($+0.2$)\\
 & $\beta=1.5$ & 0.380 ($-0.8$) & 5.674 ($+0.5$) & 29.21 ($+1.0$) & 23.72 ($-1.7$) & 1.332 ($-0.3$)\\
\midrule
Jacobian scale $\gamma$ & $1$ at every step & 0.382 ($+0.2$) & 5.671 ($+0.3$) & 28.85 ($-1.1$) & 23.67 ($-2.6$) & 1.341 ($+0.1$)\\
 & $2$ at every step & 0.381 ($-0.1$) & 5.615 ($-1.7$) & 29.03 ($-0.1$) & 23.75 ($-0.8$) & 1.366 ($+0.9$)\\
 & $\omega=4.5$ at every step & \old{\textbf{0.368} ($\mathbf{-7.0}$)} & 5.605 ($-2.2$) & 28.83 ($-1.0$) & 23.64 ($-2.7$) & 1.299 ($-1.4$)\\
\midrule
Other runs & StitchVM, $\eta=400$ & 0.366 & 5.539 & 28.78 & 23.31 & 1.355\\
\rowcolor{oursrow} & \methodfast, $\eta=300$ & 0.378 & 5.660 & 28.99 & 23.70 & 1.342\\
\rowcolor{oursrow} & \methodfast, $\eta=450$ & 0.387 & 5.666 & 28.98 & 23.72 & 1.336\\
\bottomrule
\end{tabular}}
\end{table}

\paragraph{\old{Robot policies.}} \old{On LIBERO, all methods share the 1,500 episode seeds, so we compare each variant with the base policy by an exact McNemar test on the paired outcomes. \method succeeds in 240 episodes where the base policy fails and fails in 166 where it succeeds ($p=3\times10^{-4}$), and \methodfast in 243 against 159 ($p=3\times10^{-5}$), with $p<0.05$ on each suite.} \old{Against the two variants of \cref{tab:robot_abl}, \method succeeds more often in \old{all six} paired comparisons, with a significant difference only against $\vg_s$ alone on LIBERO-Goal ($p=0.03$).}

\FloatBarrier

\subsection{\old{Text-to-Image Ablations}}
\label{app:abl-t2i}

\paragraph{\old{Ablation at $1024^2$.}} \old{With $S$ and $A$ the symmetric and antisymmetric parts of $J_\gamma$, the corrections $S-\beta A$ range from the transpose ($\beta=1$) to the Jacobian itself ($\beta=-1$)\old{ in \cref{tab:ablation_full}}: the symmetric part alone ($\beta=0$) loses a little HPSv2, and $\beta$ between $0.5$ and $1.5$ performs as the transpose.} \old{The head of \methodfast recovers most of the gain of the correction over $\vg_s$ alone (last rows\old{ of \cref{tab:ablation_full}}), one head trained with the HPSv2 value serves the other rewards on par with per-reward heads (\cref{tab:share}), and the fitted magnitude preserves HPSv2 without the second value-model backward pass (\cref{tab:norm}).}

\paragraph{\rev{Jacobian scale.}} \rev{To choose $\gamma$, we compare the sampling scale $\gamma=\omega$ with $\gamma=1$ and $\gamma=2$ at every step (\cref{tab:ablation_full}). At $\gamma=\omega$, HPSv2 drops nearly to that of $\vg_s$ alone, and PickScore and the aesthetic score also decrease, whereas $\gamma=1$ and $\gamma=2$ match the schedule in HPSv2, with $\gamma=1$ lowering PickScore. The schedule thus matches $\gamma=2$ at every step with only the conditional backward pass for $t<0.8$.}

\paragraph{\old{Role of CFG.}} \old{Over eight states per noise level, CFG roughly doubles the asymmetry of the conditional velocity, and $\gamma=2$ lies in between (\cref{tab:asym}). Along $\vg_s$, the asymmetry grows with the noise level and exceeds $1$ at $t=0.9$ for $\gamma=2$ and $\gamma=\omega$, so the transpose and the choice of $\gamma$ matter most at high noise, where we use $\gamma=2$.} \old{On FLUX.1-dev, which has no CFG,} \old{the lookahead without the correction, at twice the strength of StitchVM, raises HPSv2 and every non-target metric above StitchVM, and the correction adds little HPSv2 there and lowers the aesthetic score, while the head raises HPSv2, the aesthetic score, and ImageReward (\cref{tab:flux}).}

\begin{table}[H]
\centering\footnotesize
\begin{minipage}[t]{0.45\linewidth}\centering
\caption{\old{\textbf{Asymmetry of the velocity Jacobian}, $\|J\vu-J^{\!\top}\vu\|/\|J\vu\|$ (SD3.5-Medium, $1024^2$). $J_c=J_{\gamma=1}$.}}
\label{tab:asym}
\vspace{-0.5em}
\setlength{\tabcolsep}{3pt}
\old{\begin{tabular}{@{}ccccccc@{}}
\toprule
 & \multicolumn{3}{c}{Random $\vu$} & \multicolumn{3}{c}{$\vu=\vg_s$}\\
\cmidrule(lr){2-4}\cmidrule(lr){5-7}
$t$ & $J_\omega$ & $J_{\gamma=2}$ & $J_c$ & $J_\omega$ & $J_{\gamma=2}$ & $J_c$\\
\midrule
0.3 & 0.871 & 0.590 & 0.367 & 0.914 & 0.681 & 0.497\\
0.6 & 0.931 & 0.708 & 0.449 & 0.989 & 0.824 & 0.649\\
0.9 & 0.934 & 0.700 & 0.433 & 1.194 & 1.031 & 0.786\\
\bottomrule
\end{tabular}}
\end{minipage}\hfill
\begin{minipage}[t]{0.52\linewidth}\centering
\caption{\old{\textbf{Lookahead and correction on FLUX.1-dev} ($1024^2$, DrawBench, HPSv2 reward, same seeds).}}
\label{tab:flux}
\vspace{-0.5em}
\setlength{\tabcolsep}{2.6pt}
\old{\begin{tabular}{@{}lcccccc@{}}
\toprule
Method & $\eta$ & HPSv2\old{$\uparrow$} & Aes.\old{$\uparrow$} & CLIP\old{$\uparrow$} & Pick\old{$\uparrow$} & IR\old{$\uparrow$}\\
\midrule
StitchVM & 400 & 0.314 & 5.688 & 27.72 & 22.79 & 1.092\\
Lookahead, $\vg_s$ only & 800 & 0.327 & 5.720 & 28.28 & 23.08 & 1.226\\
\rowcolor{oursrow}\method & 800 & 0.329 & 5.575 & \textbf{28.50} & \textbf{23.14} & 1.223\\
\rowcolor{oursrow}\methodfast & 800 & \textbf{0.332} & \textbf{5.768} & 28.25 & \textbf{23.14} & \textbf{1.266}\\
\bottomrule
\end{tabular}}
\end{minipage}
\end{table}

\paragraph{\old{One head for all rewards.}} \old{In \cref{tab:share}, the head trained with the HPSv2 value, used for all rewards in \cref{tab:main}, reaches a target reward similar to that of per-reward heads, with slightly better non-target metrics.} \old{Under both rewards, either head exceeds StitchVM on every metric, so one head per generator serves every reward.}

\begin{table}[H]
\centering\footnotesize
\caption{\old{\textbf{One head for all rewards} (SD3.5-Medium, $1024^2$). Head trained with the HPSv2 value or per reward.}}
\label{tab:share}
\setlength{\tabcolsep}{2.4pt}
\old{\begin{tabular}{@{}llccccc@{}}
\toprule
Method & Head & HPSv2\old{$\uparrow$} & Aes.\old{$\uparrow$} & CLIP\old{$\uparrow$} & Pick\old{$\uparrow$} & IR\old{$\uparrow$}\\
\midrule
\multicolumn{7}{@{}l}{\textit{CLIP reward}}\\
StitchVM, $\eta=75$ & -- & 0.249 & 4.849 & 34.49 & 21.70 & 0.762\\
\rowcolor{oursrow}\methodfast, $\eta=37.5$ & own & 0.274 & 5.092 & 34.95 & 22.24 & 0.987\\
\rowcolor{oursrow}\methodfast, $\eta=37.5$ & HPSv2 & 0.277 & 5.138 & 34.99 & 22.34 & 1.011\\
\midrule
\multicolumn{7}{@{}l}{\textit{Aesthetic reward}}\\
StitchVM, $\eta=12$ & -- & 0.279 & 5.970 & 27.09 & 22.26 & 0.799\\
\rowcolor{oursrow}\methodfast, $\eta=6$ & own & 0.293 & 6.064 & 27.49 & 22.70 & 0.985\\
\rowcolor{oursrow}\methodfast, $\eta=6$ & HPSv2 & 0.294 & 6.022 & 27.84 & 22.70 & 1.016\\
\bottomrule
\end{tabular}}
\end{table}

\paragraph{\old{Fitted magnitude.}} \old{Replacing the computed magnitude of \methodfast with the fit $\hat m_t$ of \cref{sec:practical} keeps HPSv2, raises ImageReward, and removes the second value-model backward pass (\cref{tab:norm}).} \old{With the fit, \methodfast still exceeds StitchVM on every metric.}

\begin{table}[H]
\centering\footnotesize
\caption{\old{\textbf{Guidance magnitude of \methodfast} (SD3.5-Medium, $512^2$, HPDv2 test prompts, HPSv2 reward). Backward passes through the generator / value model.}}
\label{tab:norm}
\setlength{\tabcolsep}{5pt}
\old{\begin{tabular}{@{}lccccccc@{}}
\toprule
Method & $\eta$ & Bwd. & HPSv2\old{$\uparrow$} & Aes.\old{$\uparrow$} & CLIP\old{$\uparrow$} & Pick\old{$\uparrow$} & IR\old{$\uparrow$}\\
\midrule
DPS + StitchVM & 400 & 0 / 1 & 0.417 & 5.939 & 28.64 & 23.56 & 1.408\\
\rowcolor{oursrow}\methodfast, computed & 200 & 0 / 2 & 0.426 & 6.095 & 29.31 & 24.05 & 1.455\\
\rowcolor{oursrow}\methodfast, fitted & 200 & 0 / 1 & 0.425 & 6.118 & 29.36 & 24.12 & 1.509\\
\midrule
\rowcolor{oursrow}\method & 200 & 1 / 2 & 0.434 & 6.128 & 29.42 & 24.21 & 1.510\\
\bottomrule
\end{tabular}}
\end{table}

\FloatBarrier

\subsection{\old{Pick-a-Pic}}
\label{app:pap}

\old{On 499 Pick-a-Pic prompts~\cite{pickscore} with one seed, both variants raise the target reward above StitchVM and pixel-space DPS in every setting, and \methodfast exceeds StitchVM on all but three held-out entries (\cref{tab:pap}).}

\begin{table}[H]
\centering\footnotesize
\caption{\old{\textbf{Reward guidance on Pick-a-Pic} ($1024^2$, 499 prompts, one seed). Columns and bold as in \cref{tab:main}.}}
\label{tab:pap}
\setlength{\tabcolsep}{2.4pt}
\old{\begin{tabular}{@{}l ccccc c ccccc c ccccc@{}}
\toprule
 & \multicolumn{5}{c}{\textbf{SD3.5-Medium}} && \multicolumn{5}{c}{\textbf{SD3.5-Large}} && \multicolumn{5}{c}{\textbf{FLUX.1-dev}}\\
\cmidrule(lr){2-6}\cmidrule(lr){8-12}\cmidrule(lr){14-18}
Method & HPSv2\old{$\scriptstyle\uparrow$} & Aes.\old{$\scriptstyle\uparrow$} & CLIP\old{$\scriptstyle\uparrow$} & Pick\old{$\scriptstyle\uparrow$} & IR\old{$\scriptstyle\uparrow$} && HPSv2\old{$\scriptstyle\uparrow$} & Aes.\old{$\scriptstyle\uparrow$} & CLIP\old{$\scriptstyle\uparrow$} & Pick\old{$\scriptstyle\uparrow$} & IR\old{$\scriptstyle\uparrow$} && HPSv2\old{$\scriptstyle\uparrow$} & Aes.\old{$\scriptstyle\uparrow$} & CLIP\old{$\scriptstyle\uparrow$} & Pick\old{$\scriptstyle\uparrow$} & IR\old{$\scriptstyle\uparrow$}\\
\cmidrule(r){1-1}\cmidrule(lr){2-6}\cmidrule(lr){8-12}\cmidrule(lr){14-18}
Unguided & 0.302 & 5.840 & 27.59 & 22.17 & 1.087 && 0.308 & 5.940 & 27.77 & 22.31 & 1.130 && 0.304 & 6.158 & 26.70 & 22.23 & 1.031\\
\midrule
\textit{HPSv2 reward} & \cmark &  &  &  &  && \cmark &  &  &  &  && \cmark &  &  &  & \\
DPS & 0.384 & 5.955 & 28.25 & 23.13 & 1.284 && 0.368 & 6.006 & 28.14 & 23.10 & 1.266 && 0.334 & 5.930 & 26.63 & 22.17 & 0.953\\
DPS + StitchVM & 0.387 & 5.967 & 27.86 & 22.96 & 1.393 && 0.380 & 5.940 & 27.82 & 22.77 & 1.347 && 0.325 & 6.119 & 26.89 & 22.21 & 1.127\\
\rowcolor{oursrow}\textbf{\method} (Ours) & 0.407 & 6.011 & \textbf{28.42} & \textbf{23.34} & \textbf{1.471} && \textbf{0.393} & \textbf{6.065} & 28.52 & \textbf{23.33} & \textbf{1.437} && 0.346 & 6.102 & \textbf{27.44} & 22.60 & 1.268\\
\rowcolor{oursrow}\textbf{\methodfast} (Ours) & \textbf{0.408} & \textbf{6.014} & 28.22 & 23.20 & 1.426 && 0.385 & 6.050 & \textbf{28.62} & 23.03 & 1.388 && \textbf{0.348} & \textbf{6.261} & 27.42 & \textbf{22.63} & \textbf{1.319}\\
\midrule
\textit{Aesthetic reward} &  & \cmark &  &  &  &&  & \cmark &  &  &  &&  & \cmark &  &  & \\
DPS & 0.303 & 6.379 & \textbf{27.51} & 22.21 & 1.058 && 0.309 & 6.355 & \textbf{27.93} & 22.35 & 1.123 && 0.304 & 6.258 & 26.58 & 22.20 & 1.007\\
DPS + StitchVM & 0.291 & 6.458 & 26.57 & 21.79 & 0.831 && 0.300 & 6.314 & 27.54 & 22.11 & 1.049 && 0.306 & 6.350 & 26.54 & 22.23 & \textbf{1.061}\\
\rowcolor{oursrow}\textbf{\method} (Ours) & 0.305 & \textbf{6.877} & 26.28 & 22.14 & 1.005 && \textbf{0.311} & 6.606 & 27.58 & \textbf{22.42} & 1.132 && 0.308 & \textbf{6.497} & 26.46 & 22.27 & 1.017\\
\rowcolor{oursrow}\textbf{\methodfast} (Ours) & \textbf{0.311} & 6.651 & 27.10 & \textbf{22.27} & \textbf{1.142} && \textbf{0.311} & \textbf{6.727} & 27.56 & 22.32 & \textbf{1.157} && \textbf{0.309} & 6.409 & \textbf{26.61} & \textbf{22.28} & 1.047\\
\midrule
\textit{CLIP reward} &  &  & \cmark &  &  &&  &  & \cmark &  &  &&  &  & \cmark &  & \\
DPS & 0.269 & 5.395 & 34.71 & 21.17 & 0.738 && \textbf{0.298} & \textbf{5.821} & 31.70 & 21.99 & 0.982 && 0.289 & 5.888 & 28.60 & 21.72 & 0.838\\
DPS + StitchVM & 0.258 & 5.370 & 34.73 & 21.11 & 0.686 && 0.285 & 5.714 & 31.84 & 21.80 & 0.962 && 0.296 & 5.998 & 29.44 & 22.00 & 1.019\\
\rowcolor{oursrow}\textbf{\method} (Ours) & 0.275 & 5.504 & \textbf{36.29} & 21.47 & 0.938 && 0.294 & 5.792 & \textbf{33.87} & \textbf{22.04} & \textbf{1.088} && 0.301 & 6.042 & \textbf{30.91} & 22.18 & 1.102\\
\rowcolor{oursrow}\textbf{\methodfast} (Ours) & \textbf{0.284} & \textbf{5.596} & 35.35 & \textbf{21.65} & \textbf{0.984} && 0.290 & 5.703 & 33.83 & 21.78 & 1.067 && \textbf{0.305} & \textbf{6.046} & 30.56 & \textbf{22.19} & \textbf{1.136}\\
\bottomrule
\end{tabular}}
\end{table}

\FloatBarrier

\subsection{\old{Robot Ablations}}
\label{app:abl-robot}

\paragraph{\old{Lookahead ratio and value models.}} \old{On LIBERO-Goal, $s=0.6t$ gives the highest success rate averaged over strengths (\cref{tab:robot_ratio}). The three value models of \cref{tab:robot_v} are regressed onto the critic, as StitchVM regresses onto the reward: an MLP on forward-noised action chunks $t\veps+(1-t)\va_0$, a head stitched to the features of the policy's action expert, as in StitchVM, and an MLP on the states of the policy's own sampling path. Both guidance forms take the norm of $\nabla V_\psi(\vz_t,t)$, and the strength is the best of $w\in\{0.25,0.5,1,2,4\}$.}

\begin{table}[!ht]
\centering\footnotesize
\caption{\old{\textbf{Learned value on SmolVLA} (LIBERO-Goal, success \%, base 76.4), read at $\vz_t$ (StitchVM) or with the lookahead (\method).}}
\label{tab:robot_v}
\vspace{-0.5em}
\old{\begin{tabular}{@{}lcc@{}}
\toprule
Value model & StitchVM & \cellcolor{oursrow}\method\\
\midrule
MLP, forward-noised actions & 77.8 & \cellcolor{oursrow}\textbf{80.6}\\
Stitched to the policy & 77.4 & \cellcolor{oursrow}\textbf{79.6}\\
MLP, policy sampling path & 80.2 & \cellcolor{oursrow}\textbf{80.4}\\
\bottomrule
\end{tabular}}
\end{table}

\begin{table}[!ht]
\centering\footnotesize
\caption{\old{\textbf{Lookahead ratio} of \method on LIBERO-Goal. Success rate (\%).} \old{Bold: best.}}
\label{tab:robot_ratio}
\old{\begin{tabular}{@{}lcccc@{}}
\toprule
$s/t$ & $w=1$ & $w=2$ & $w=4$ & \old{Avg.}\\
\midrule
0.4 & 77.0 & \old{\textbf{79.2}} & \textbf{80.6} & \old{78.9}\\
\rowcolor{oursrow}0.6 & \textbf{80.4} & 79.0 & \textbf{80.6} & \old{\textbf{80.0}}\\
0.8 & 79.0 & 78.4 & 80.0 & \old{79.1}\\
\bottomrule
\end{tabular}}
\end{table}

\FloatBarrier

\subsection{\old{Qualitative Results}}
\label{app:qual}

\begin{figure*}[p]
\centering
\includegraphics[width=\linewidth]{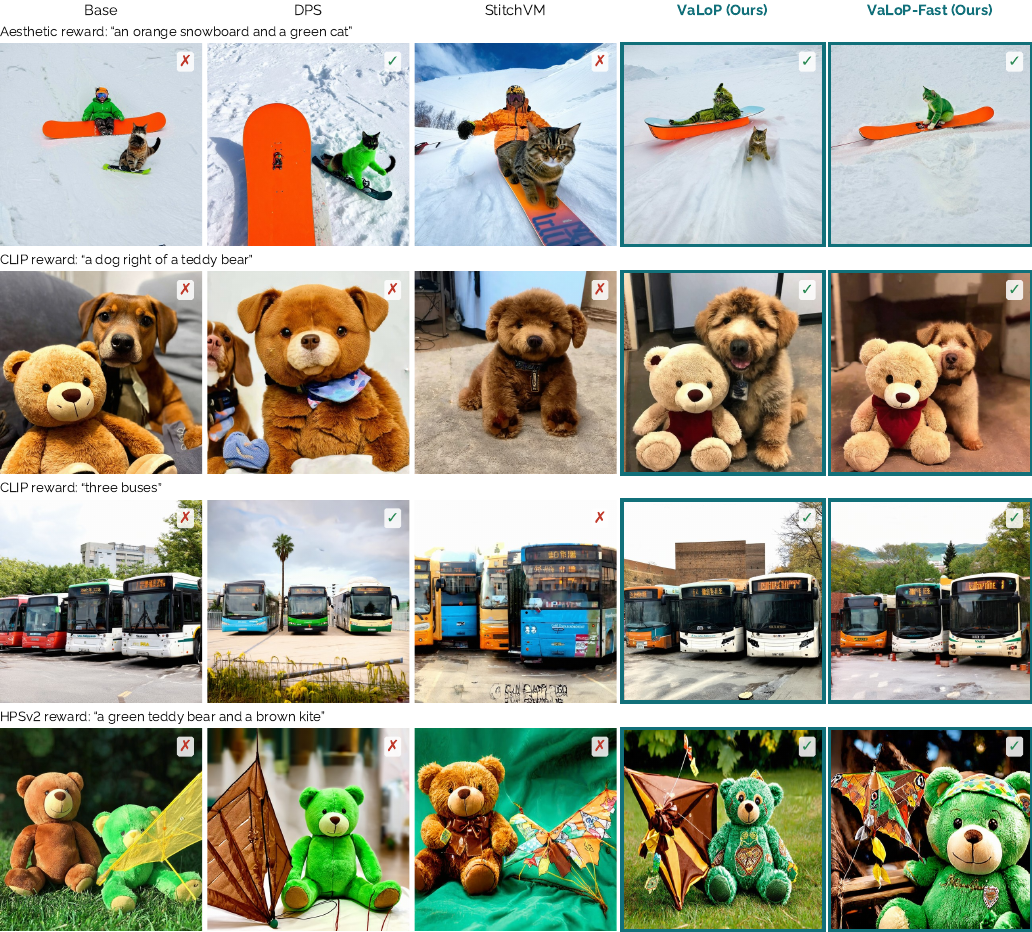}
\caption{\old{\textbf{Qualitative comparison on SD3.5-Medium} under the target reward of each row. \cmark/\xmark: GenEval correctness.}}
\label{fig:qual-m}
\end{figure*}

\begin{figure*}[p]
\centering
\includegraphics[width=\linewidth]{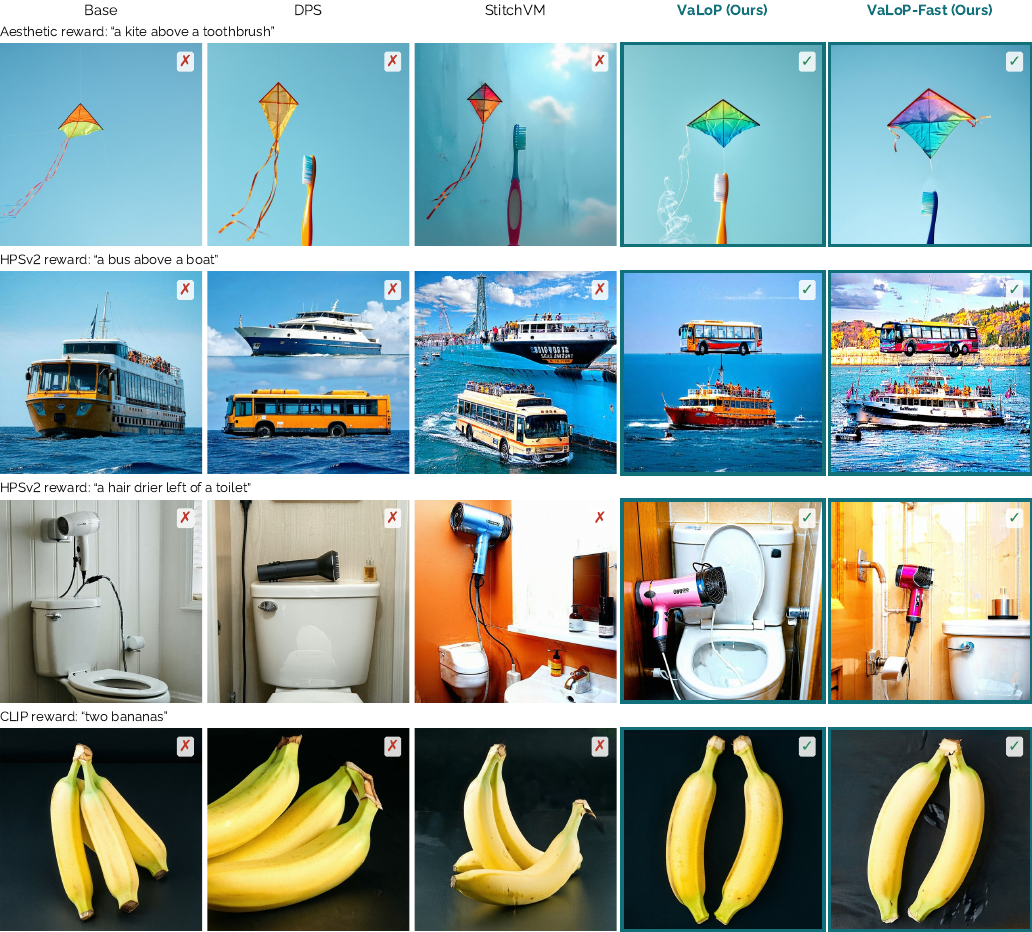}
\caption{\old{\textbf{Qualitative comparison on SD3.5-Large} under the target reward of each row. \cmark/\xmark: GenEval correctness.}}
\label{fig:qual-l}
\end{figure*}

\begin{figure*}[p]
\centering
\includegraphics[width=\linewidth]{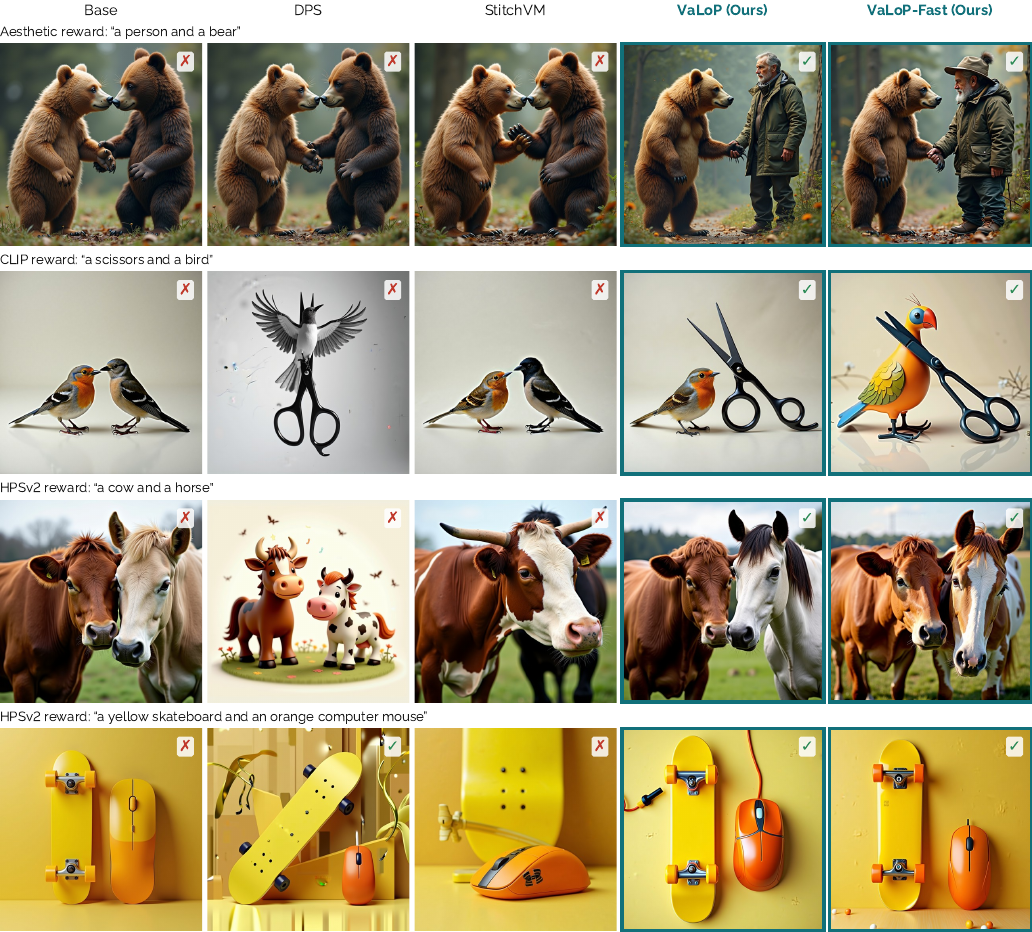}
\caption{\old{\textbf{Qualitative comparison on FLUX.1-dev} under the target reward of each row. \cmark/\xmark: GenEval correctness.}}
\label{fig:qual-f}
\end{figure*}

\paragraph{\old{Text-to-image generation.}} \old{\Cref{fig:qual-m,fig:qual-l,fig:qual-f} compare the methods on selected GenEval prompts for each model, on which the unguided model and StitchVM miss an object, its count, its attribute, or its relation. Both variants follow these prompts while guiding toward the target reward. The prompts are selected to show visible differences between the methods.}

\begin{figure*}[p]
\centering
\includegraphics[width=\linewidth]{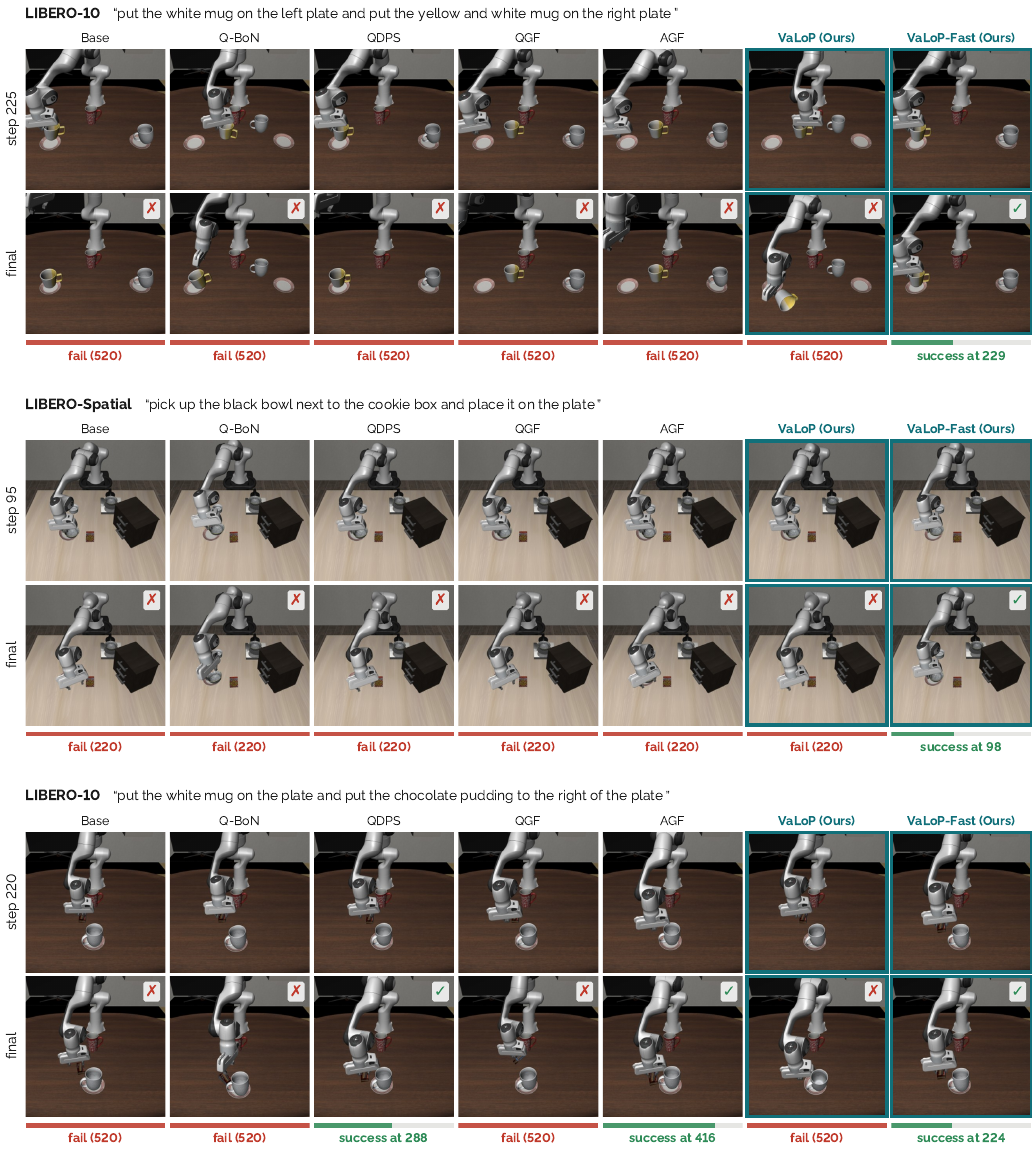}
\caption{\old{\textbf{Rollouts on LIBERO} (SmolVLA) from the same initial state, \old{just before} \methodfast succeeds and at the end of each episode, with bars for the steps to success or the time limit. Only \methodfast completes the first two tasks, and it is first on the third.}}
\label{fig:robot-libero1}
\end{figure*}

\begin{figure*}[p]
\centering
\includegraphics[width=\linewidth]{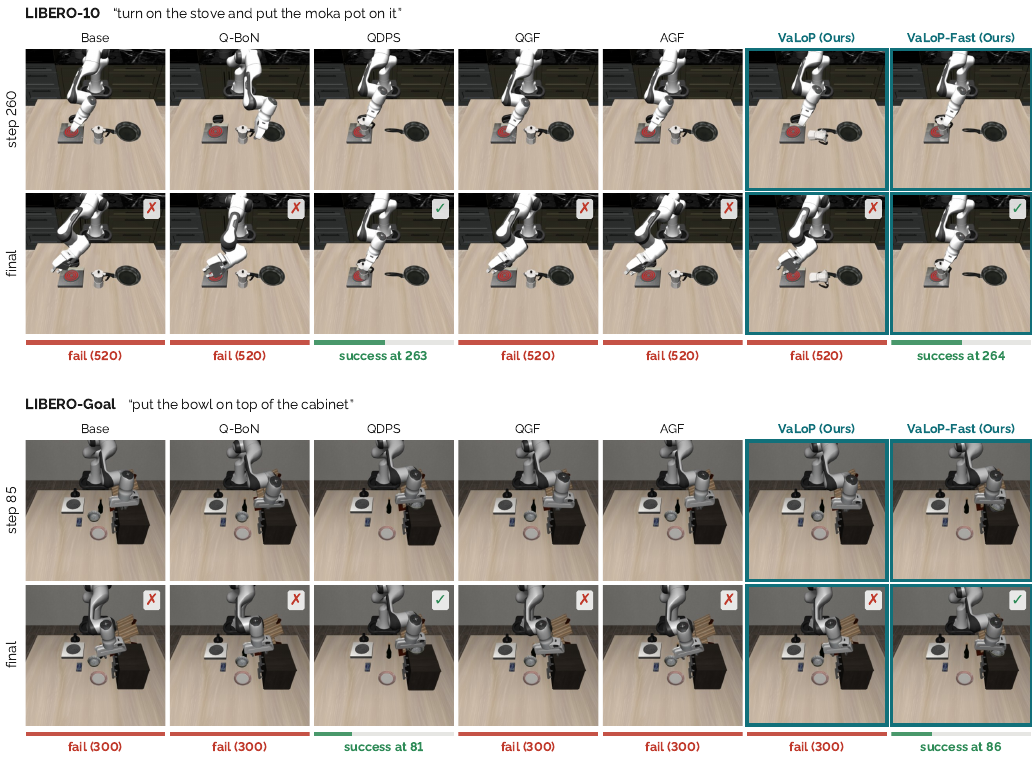}
\caption{\old{\textbf{Rollouts on LIBERO}, continued. QDPS also completes both tasks, one and five steps earlier than \methodfast.}}
\label{fig:robot-libero2}
\end{figure*}

\begin{figure*}[p]
\centering
\includegraphics[width=\linewidth]{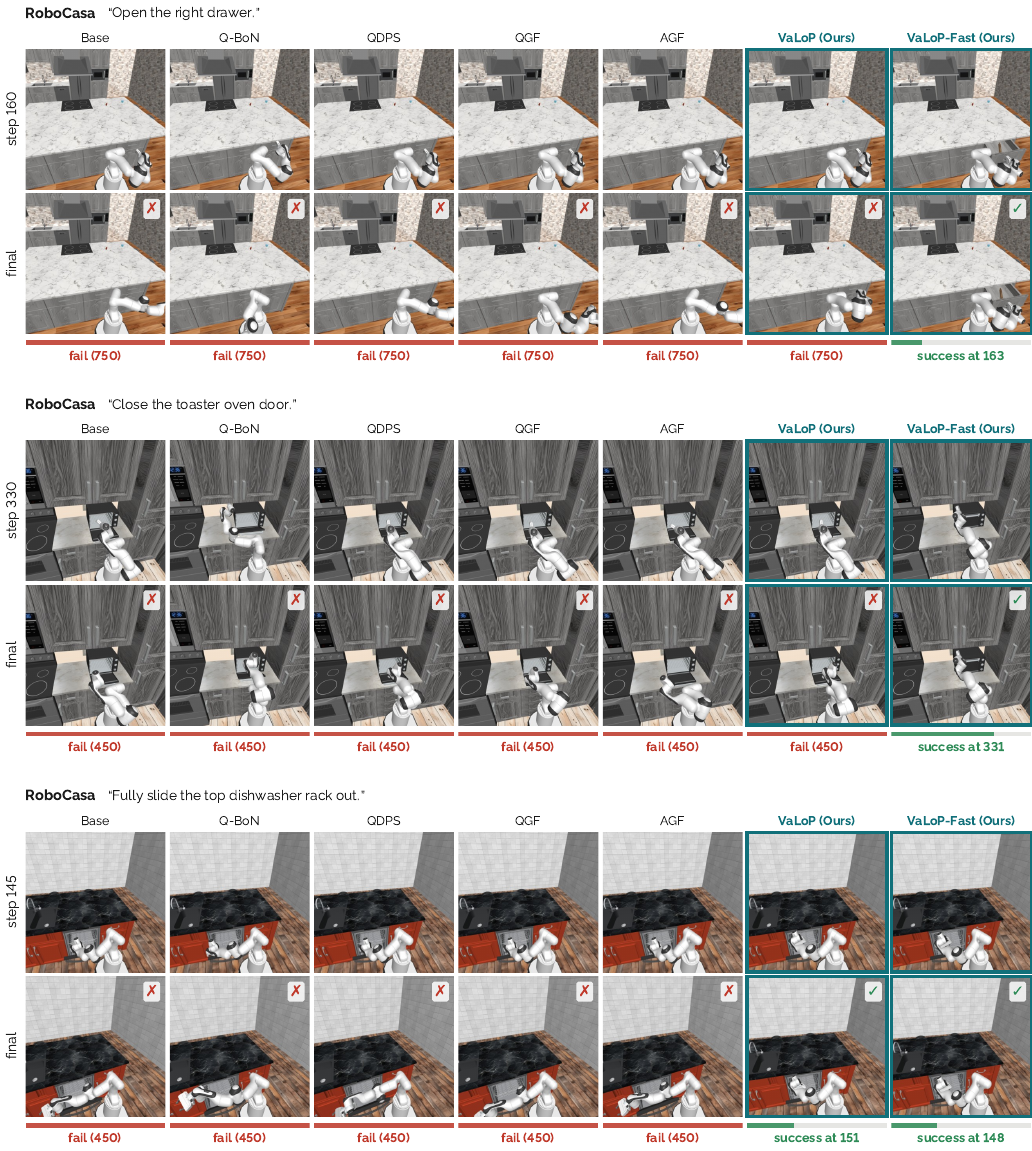}
\caption{\old{\textbf{Rollouts on RoboCasa} ($\pi_{0.5}$), shown as in \cref{fig:robot-libero1}. Only \methodfast opens the drawer and closes the toaster oven door, and only our two variants slide out the rack.}}
\label{fig:robot-robocasa1}
\end{figure*}

\begin{figure*}[p]
\centering
\includegraphics[width=\linewidth]{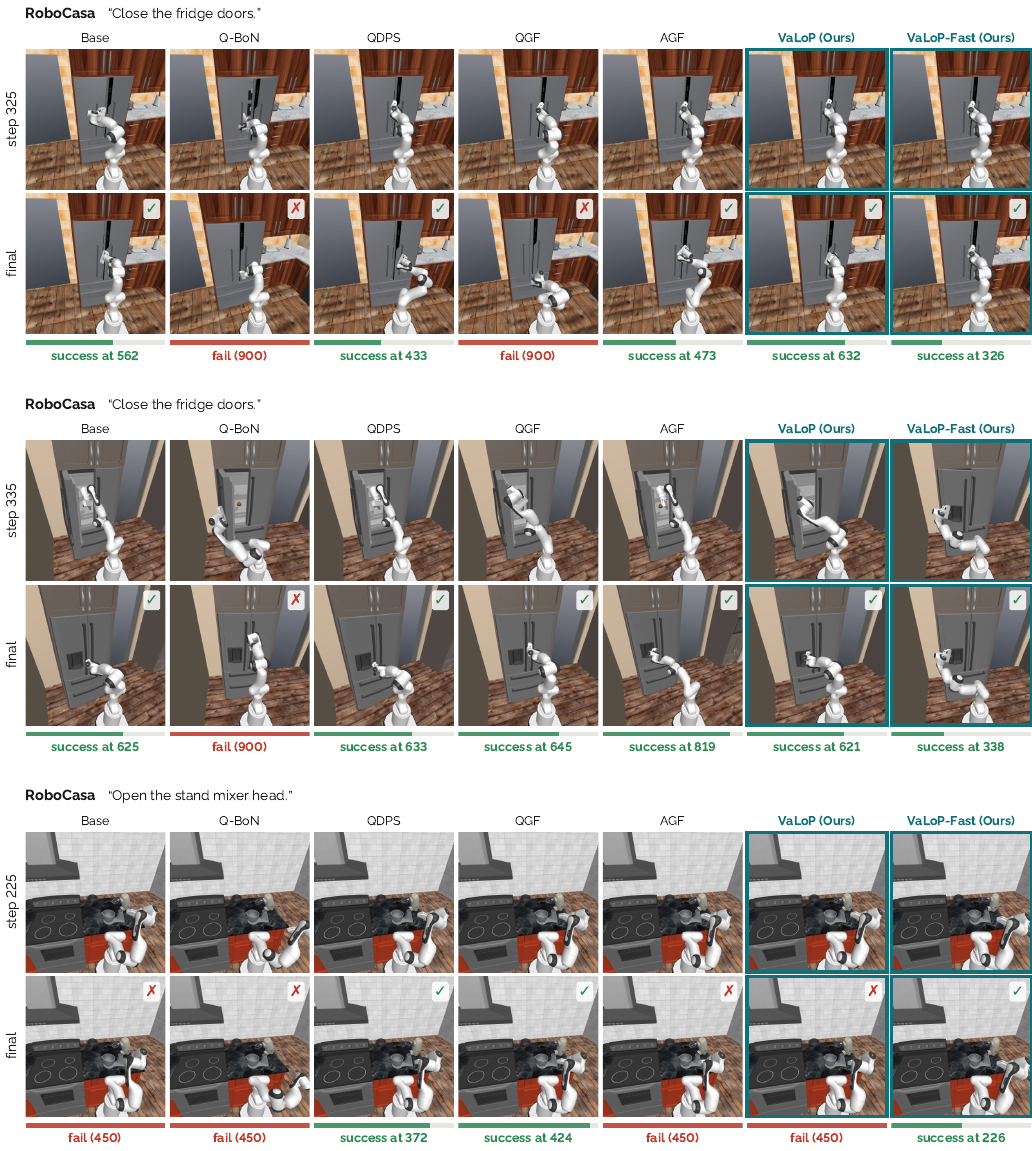}
\caption{\old{\textbf{Rollouts on RoboCasa}, continued. Several baselines also complete these tasks, but later than \methodfast.}}
\label{fig:robot-robocasa2}
\end{figure*}

\paragraph{\old{Robot rollouts.}} \old{\Cref{fig:robot-libero1,fig:robot-libero2,fig:robot-robocasa1,fig:robot-robocasa2} show eleven episodes from the same initial state for all methods, rendered from the recorded simulator states and selected to show visible differences between the methods. \methodfast succeeds in every episode and is the first to succeed in nine of them, while in the other two QDPS succeeds one and five steps earlier.}

\end{document}